%% file: main.tex
\documentclass{template}
\usepackage[utf8]{inputenc} % allow utf-8 input
\usepackage[T1]{fontenc}    % use 8-bit T1 fonts
\usepackage{booktabs}       % professional-quality tables
\usepackage{amsfonts}       % blackboard math symbols
\usepackage{nicefrac}       % compact symbols for 1/2, etc.
\usepackage{microtype}      % microtypography

\usepackage{mwe}
\usepackage{graphicx}
\usepackage{amssymb}
\usepackage{adjustbox}
\usepackage{bbding}
\usepackage{colortbl}
\usepackage{array}
\usepackage{multirow}
\usepackage{makecell}
\usepackage{bbm}
\usepackage{collcell,xfp}
\usepackage{pgf}
\usepackage{tikz}
\usepackage[most]{tcolorbox}
\usepackage{csquotes}
\usepackage[noorphans,vskip=1em,leftmargin=1em]{quoting}
\usepackage{pifont} % For custom symbols
\usepackage{enumitem} % For customized itemize settings
\usepackage{tcolorbox} % For creating a boxed environment
\usepackage{forest}
\usepackage{wrapfig}
\usepackage{caption}
\usepackage{subcaption}
\usepackage{longtable}
\usepackage{algpseudocode}
\usepackage{amsmath}
\usepackage[capitalize]{cleveref}
\usepackage[ruled,vlined,linesnumbered]{algorithm2e} 
\usepackage[percent]{overpic}
\usepackage{xspace}
\usepackage[normalem]{ulem}  % for \ul, \setul, \setulcolor
\usepackage{tocloft}  % TOC customization
\usepackage[numbers, sort]{natbib}  % citep
\usepackage{hyperref}       % hyperlinks
\usepackage{url}            % simple URL typesetting
\tcbuselibrary{breakable}
\usepackage[hang,flushmargin]{footmisc}
\tcbset{fontupper=\footnotesize\ttfamily}

\usepackage{fontawesome}
\newcommand{\huggingface}{\raisebox{-1.5pt}{\includegraphics[height=1.05em]{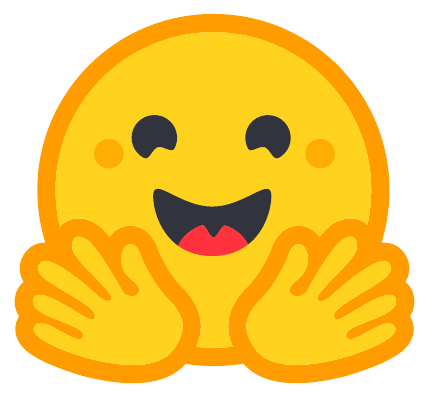}}\xspace}
\newcommand{\github}{\raisebox{-1.5pt}{\includegraphics[height=1.05em]{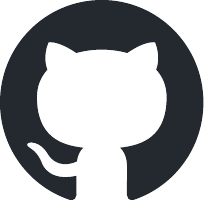}}\xspace}
\newcommand{\worldwideweb}{\raisebox{-1.5pt}{\includegraphics[height=1.05em]{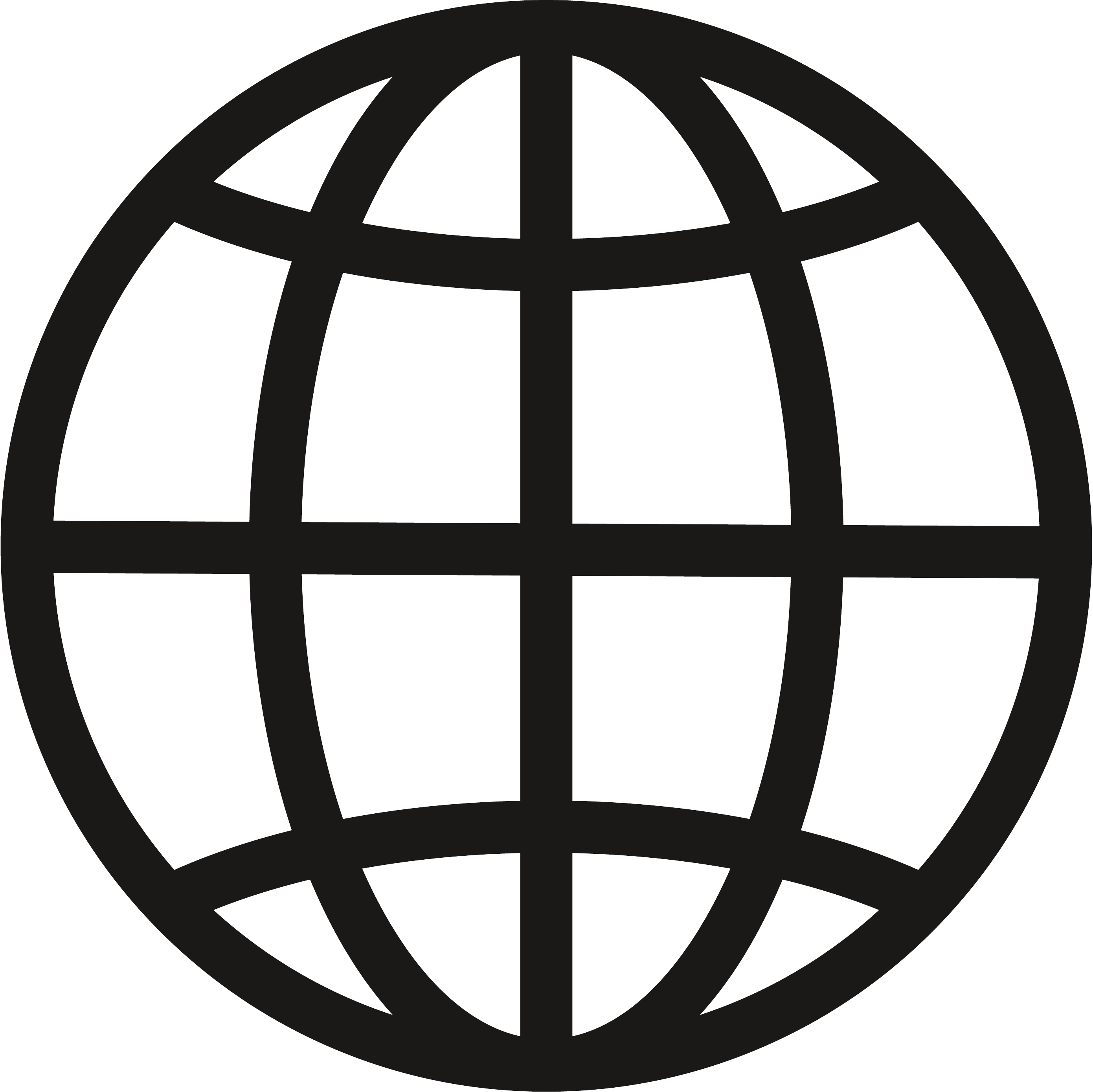}}\xspace}

\definecolor{scholarblue}{rgb}{0.21,0.49,0.74}
\definecolor{bluelink}{RGB}{0,113,188}
\definecolor{greenlink}{RGB}{0,188,113}
\hypersetup{
    colorlinks=true,%
    citecolor=scholarblue,%
    filecolor=red,%
    linkcolor=red!93!black,%
    urlcolor=bluelink
}

\usepackage{titlesec}
\titlespacing*{\paragraph}
    {0pt}     % left margin
    {0.25em}  % space before (vertical above)
    {1em}  % space after (vertical below, adjust as needed)

\definecolor{navyblue}{HTML}{0071BC}
\definecolor{blindcolor}{HTML}{AB2AC6}    % Blind – purple
\definecolor{chancecolor}{HTML}{F59E0B}   % Chance – amber
\definecolor{singlecolor}{HTML}{06B6D4}   % Single – cyan
\definecolor{multiplecolor}{HTML}{2563EB} % Multiple – blue
\definecolor{captioncolor}{HTML}{22C55E}  % Caption – green

 \newcommand{\culine}[2]{%
    \def\temp@uline{\bgroup\markoverwith
        {\textcolor{#1}{\rule[-0.5ex]{2pt}{1pt}}}\ULon}%
    \temp@uline{#2}%
}
 \newcommand{\cthickuline}[3][0.8pt]{%
    \def\temp@uline{\bgroup\markoverwith
        {\textcolor{#2}{\rule[-0.5ex]{2pt}{#1}}}\ULon}%
    \temp@uline{#3}%
}

\title{\center{\fontsize{13.6}{10}\selectfont SpatialScore: Towards Comprehensive Evaluation for Spatial Intelligence}}

\renewcommand\Affilfont{\normalfont\fontsize{11}{15}\selectfont\centering}
\author{
    Haoning~Wu\textsuperscript{1, 2}$^{*}$ \quad\ 
    Xiao~Huang\textsuperscript{1}$^{*}$ \quad\ 
    Yaohui~Chen\textsuperscript{1} \quad\ \\
    \vspace{-8pt}
    Ya~Zhang\textsuperscript{1, 2} \quad\ 
    Yanfeng~Wang\textsuperscript{1} \quad\
    Weidi~Xie\textsuperscript{1} \quad\
    \vspace{6pt}
    \textsuperscript{1} School of Artificial Intelligence, Shanghai Jiao Tong University \\ 
    \textsuperscript{2} Shanghai Artificial Intelligence Laboratory
}

\input{sections/0_abstract}

\renewcommand{\thefootnote}{\arabic{footnote}}  % Set back to default numbering

\definecolor{mygreen}{HTML}{B9FFBF}
\definecolor{mylightgreen}{HTML}{D9F0D9}

\begin{document}

\maketitle

\begingroup
    \renewcommand{\thefootnote}{\fnsymbol{footnote}}
    \footnotetext[1]{Haoning Wu led the project; Haoning Wu and Xiao Huang contributed equally.}
    % \footnotetext[2]{Corresponding Authors.}
\endgroup

\begin{center}
    \renewcommand{\arraystretch}{1.5}
    \begin{tabular}{rll}
        \worldwideweb{} & \textbf{Website} & \url{https://haoningwu3639.github.io/SpatialScore/}\\
        \github{} & \textbf{Code} & \url{https://github.com/haoningwu3639/SpatialScore/}\\
        \huggingface{} & \textbf{Checkpoints} & \href{https://huggingface.co/datasets/haoningwu/SpatialScore}{\nolinkurl{https://huggingface.co/haoningwu/SpatialScore}} \\
        \huggingface{} & \textbf{Benchmark} & \href{https://huggingface.co/datasets/haoningwu/SpatialScore}{\nolinkurl{https://huggingface.co/datasets/haoningwu/SpatialScore}} \\
        \huggingface{} & \textbf{Data} & \href{https://huggingface.co/datasets/haoningwu/SpatialScore}{\nolinkurl{https://huggingface.co/datasets/haoningwu/SpatialCorpus}} \\
    \end{tabular}
\end{center}

%%%%%%%%%%%%%%%%%%%%%%%%%%%%%%%%%%%%%%%%%%%%%%%%%%%%%%%%%%%%%%%%%%%%%%%%%%%%%%%%
% TABLE OF CONTENTS
%%%%%%%%%%%%%%%%%%%%%%%%%%%%%%%%%%%%%%%%%%%%%%%%%%%%%%%%%%%%%%%%%%%%%%%%%%%%%%%%

\newpage
{
    \hypersetup{linkcolor=black}
    \tableofcontents
}
\newpage

%%%%%%%%%%%%%%%%%%%%%%%%%%%%%%%%%%%%%%%%%%%%%%%%%%%%%%%%%%%%%%%%%%%%%%%%%%%%%%%%
% MAIN TEXT
%%%%%%%%%%%%%%%%%%%%%%%%%%%%%%%%%%%%%%%%%%%%%%%%%%%%%%%%%%%%%%%%%%%%%%%%%%%%%%%%

\input{sections/1_introduction}
\input{sections/2_benchmark}
\input{sections/3_method}
\input{sections/4_experiments}
\input{sections/5_relatedwork}
\input{sections/6_conclusion}
\input{sections/7_acknowledgement}
{
    \small
    \bibliographystyle{plain}
    \bibliography{main}
}

\input{sections/X_suppl}

\end{document}

%% file: sections/0_abstract.tex
\begin{abstract}
Existing evaluations of multimodal large language models~(MLLMs) on spatial intelligence are typically fragmented and limited in scope. 
In this work, we aim to conduct a holistic assessment of the spatial understanding capabilities of modern MLLMs and propose complementary data-driven and agent-based solutions.
Specifically, we make the following contributions:
(i) we introduce \textbf{SpatialScore}, 
to our knowledge, the most comprehensive and diverse benchmark for multimodal spatial intelligence to date. 
It covers multiple visual data types, input modalities, and question-answering formats, and contains approximately 5K manually verified samples spanning 30 distinct tasks;
(ii) using SpatialScore, we extensively evaluate 49 representative MLLMs, revealing persistent challenges and a substantial gap between current models and human-level spatial intelligence;
(iii) to advance model capabilities, we construct \textbf{SpatialCorpus}, a large-scale training resource with 331K multimodal QA samples that supports fine-tuning on spatial reasoning tasks and significantly improves the performance of existing models ({\em e.g.}, Qwen3-VL);
(iv) to complement this data-driven route with a training-free paradigm, we develop \textbf{SpatialAgent}, a multi-agent system equipped with 12 specialized spatial perception tools that supports both {\em Plan-Execute} and {\em ReAct} reasoning, enabling substantial gains in spatial reasoning without additional model training.
Extensive experiments and in-depth analyses demonstrate the effectiveness of our benchmark, corpus, and agent framework. 
We expect these resources to serve as a solid foundation for advancing MLLMs toward human-level spatial intelligence. 
All data, code, and models will be released to the research community.

\end{abstract}

%% file: sections/1_introduction.tex
\section{Introduction}
\label{sec:introduction}

Multimodal large language models~(MLLMs) have recently demonstrated strong performance across diverse domains and tasks~\cite{chow2025physbench, ghafarollahi2024sciagents, li2024agenthospital, rao2025unisoccer, rao2024matchtimeautomaticsoccergame, liu2025lamra, liu2026versavit}.
While modern MLLMs excel at general semantic question answering~\cite{li2024seed-bench, liu2024mmbench, yue2024mmmu}~({\em e.g.}, answering `who', `what', and `where' questions) and mathematical reasoning~\cite{lu2024mathvista, wang2024Math-Vision, zhang2024mathverse, zou2025dynamath}, progress on spatial intelligence~\cite{SpatialLLM, Spatial-MLLM, VLM-3R} remains fragmented~\cite{RoboSpatial, MIRAGE, SpatialEval, Space-10, cai2025HasGPT5, VLM4D, STI-Bench}. 
This gap is particularly concerning given that such human-like spatial reasoning ability is crucial for real-world applications like embodied AI and autonomous navigation.

\begin{figure*}[b]
    \centering
    \vspace{-0.3cm}
    \includegraphics[width=\textwidth]{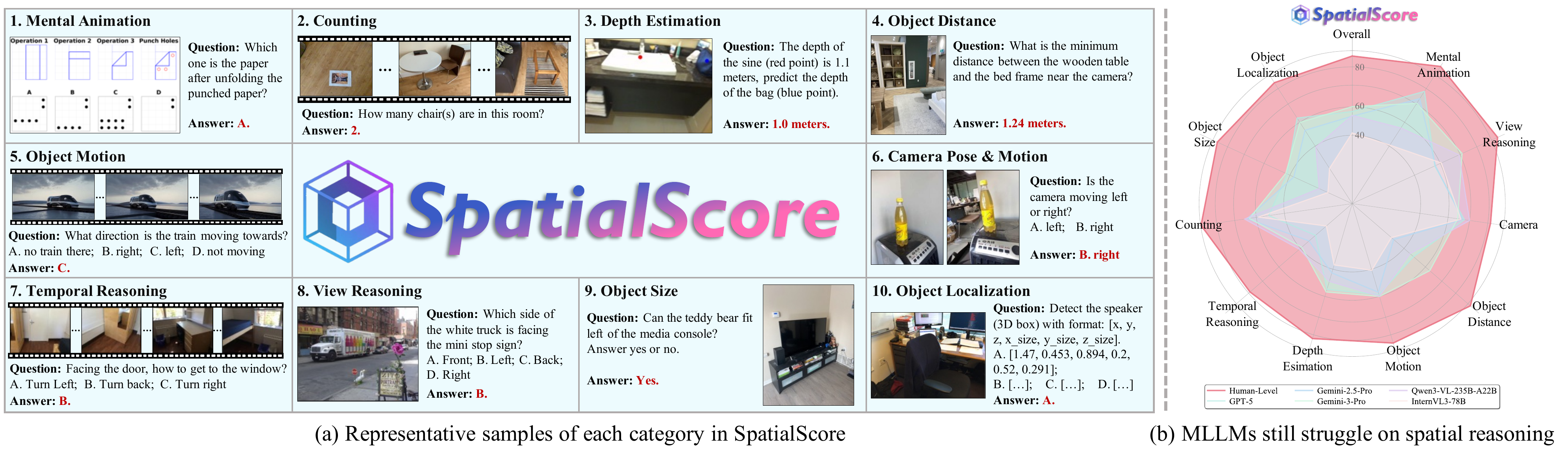} 
    \vspace{-0.6cm}
    \caption{
        \textbf{Overview.}
        (a) Representative examples from distinct categories in \textbf{SpatialScore}, which thoroughly assesses spatial intelligence capabilities via question-answering~(judgment, multi-choice, and open-ended QA);
        (b) Performance of state-of-the-art models compared to humans on SpatialScore.
    }
    \label{fig:teaser}
    % \vspace{-6pt}
\end{figure*}

Conventional computer vision has developed well-established tools~\cite{schoenberger2016sfm}~(often based on geometric optimization) and rigorous mathematical foundations~\cite{hartley2003MultipleViewGeometry} for spatial perception. 
Recent work~\cite{wang2025vggt, wang2024DUSt3R, DepthAnythingv2, meng2026scenegen} has revitalized these approaches with feed-forward neural networks, yet these advances largely remain within vision-only paradigms, lacking tight language integration and unified evaluation protocols. Building on recent progress in MLLMs and spatial perception, it is natural to treat the integration of semantic understanding and spatial perception as a next frontier. To this end, we seek to systematically investigate~(as illustrated in Fig.~\ref{fig:teaser}): {\em to what extent do existing MLLMs possess spatial intelligence, encompassing both spatial perception and spatial understanding?}

Several recent studies~\cite{OmniSpatial, MMSI-Bench, SPAR, Multi-SpatialMLLM} have begun to explore this direction. However, this line of work remains nascent and faces two key limitations: 
(i) \textbf{over-simplistic tasks}. 
Existing benchmarks~\cite{cai2024spatialbot, kamath2023whatsup, VSR, tong2024cambrian, wang2024spatial, SpatialReasoner, yang2019spatialsense} primarily focus on superficial spatial queries~({\em e.g.}, object presence or coarse position relations), while neglecting rigorous visual geometry perception~({\em e.g.}, camera pose or dynamics); 
(ii) \textbf{narrow evaluation scope}. 
Prior assessments~\cite{MindTheGap, SpatialViz-Bench, kamath2023whatsup, liao2024QSpatialBench, VSR, ma20243dsrbench, yang2019spatialsense} are typically fragmented, relying on naive questions ({\em e.g.}, Yes/No judgments), single-modality inputs ({\em e.g.}, static images), or isolated skills ({\em e.g.}, size estimation), and thus fail to provide a holistic measurement of spatial intelligence.

To tackle these challenges, we first repurpose widely used 3D datasets into question-answering formats and integrate them with spatially relevant samples from 23 existing datasets, constructing \textbf{SpatialScore}, a diverse and comprehensive benchmark for spatial understanding~(Fig.~\ref{fig:teaser}). 
SpatialScore contains approximately \textbf{5K} manually verified, high-quality samples spanning \textbf{30} distinct spatial reasoning tasks~({\em e.g.}, metric-based distance measurement, homography estimation), and covers diverse data types~(real-world, simulation, and AIGC), modalities~(images and videos), and question formats~(judgment, multi-choice, and open-ended QA). 
Extensive evaluations of 49 representative MLLMs on SpatialScore reveal persistent challenges in spatial intelligence and a substantial gap relative to human performance.

To enhance the spatial reasoning capabilities of MLLMs, we then explore two complementary pathways, progressing from a direct data-driven strategy to an agent-based solution. 
First, we leverage 2D simulators and existing 3D annotations~\cite{yeshwanth2023scannet++, xia2024WildRGB-D, brazil2023omni3d, zheng2023pointodyssey} to construct \textbf{SpatialCorpus}, a large-scale training resource containing 331K multimodal, spatially relevant QA samples. SpatialCorpus supports supervised fine-tuning of MLLMs ({\em e.g.}, Qwen3-VL~\cite{Qwen3-VL}) on spatial intelligence tasks, serving as a data-driven route to strengthen spatial reasoning. 
Second, we propose \textbf{SpatialAgent}, an agentic framework that orchestrates 12 specialized spatial perception tools ({\em e.g.}, depth estimator~\cite{DepthAnythingv2}, camera pose estimator~\cite{wang2025vggt}, motion estimator~\cite{teed2020raft}). 
SpatialAgent enables pretrained MLLMs to perform spatial reasoning under two paradigms:
(i) {\em Plan-Execute}: a hierarchical strategy that decomposes complex tasks into structured sub-tasks with sequential tool invocation;
(ii) {\em ReAct}: an interleaved reasoning-and-action scheme that iteratively refines decisions via context-aware tool interactions. 
Through dynamic tool orchestration, SpatialAgent improves the spatial understanding of off-the-shelf MLLMs in a training-free manner.

The remainder of this paper is organized as follows.
Sec.~\ref{sec:spatialscore} details the construction of the \textbf{SpatialScore} benchmark and presents a comprehensive evaluation of existing models.
Sec.~\ref{sec:Methodology} introduces our strategies for enhancing spatial intelligence, including the curation of the \textbf{SpatialCorpus} training resource and the design of the \textbf{SpatialAgent} multi-agent system.
Sec.~\ref{sec:Experiments} describes our experimental protocols and provides quantitative and qualitative evidence of the gains achieved by our approaches.
Sec.~\ref{sec:related_work} reviews related literature, and Sec.~\ref{sec:conclusion} summarizes our key insights and contributions.
To the best of our knowledge, this work establishes the most comprehensive and diverse spatial intelligence benchmark to date, and we hope it serves as a rigorous testbed to foster future advances in MLLMs.

%% file: sections/2_benchmark.tex
\begin{figure*}[t]
  \centering
  \includegraphics[width=\textwidth]{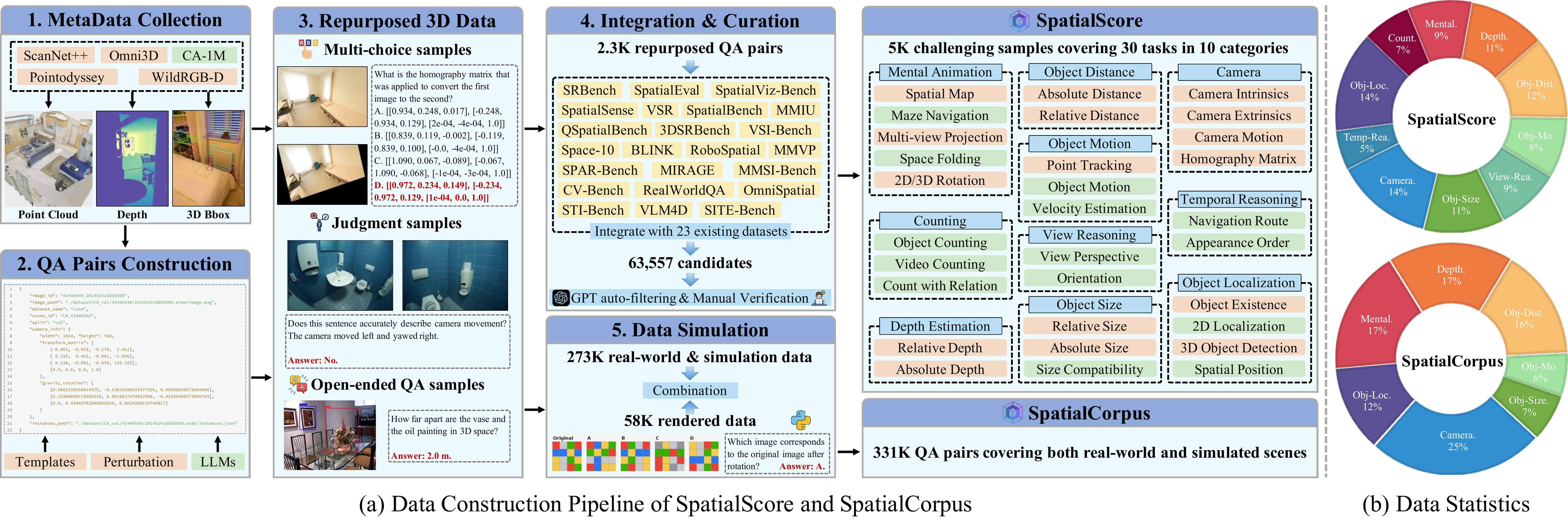} 
  \vspace{-0.6cm}
  \caption{
  \textbf{Dataset Construction and Statistics.}
      (a) Data construction pipeline for SpatialScore and SpatialCorpus.
      Here, the metadata, curation strategies, and tasks highlighted in \setlength{\fboxsep}{2pt}\colorbox[HTML]{D1E8CA}{green} are specific to the construction of SpatialScore, while those highlighted in \setlength{\fboxsep}{2pt}\colorbox[HTML]{F5D9C5}{orange} are applicable to both SpatialScore and SpatialCorpus.
      (b) Data distribution statistics across SpatialScore and SpatialCorpus.
  }
 \label{fig:dataset}
 % \vspace{-6pt}
\end{figure*}

\section{SpatialScore}
\label{sec:spatialscore}

This section first introduces the construction of \textbf{SpatialScore}, our proposed spatial intelligence benchmark~(Sec.~\ref{subsec:dataset_construction}). 
We then present a detailed statistical analysis and discussion of the collected data~(Sec.~\ref{subsec:data_statistics}). 
Finally, we report the performance of representative MLLMs on this holistic benchmark~(Sec.~\ref{subsec:comparison_on_spatialscore}).

\subsection{Dataset Construction}
\label{subsec:dataset_construction}
To enable a holistic evaluation of the spatial understanding capabilities of MLLMs, we construct \textbf{SpatialScore}, to our knowledge, the most comprehensive and diverse spatial intelligence benchmark to date. 
SpatialScore combines newly introduced question-answering data repurposed from existing 3D annotations with spatially related samples from 23 public datasets, as detailed below.

\vspace{2pt} 
\noindent \textbf{3D Data Repurposing.}
Given the scarcity of QA data on 3D visual geometry perception~({\em e.g.}, camera pose, point tracking, depth estimation) in existing MLLM benchmarks, we design a scalable and controllable pipeline that leverages precise 3D annotations ({\em e.g.}, depth, 3D bounding boxes) to generate high-quality QA pairs.
As illustrated in Fig.~\ref{fig:dataset}(a), we first randomly sample 500 scenes with accurate 3D annotations from multiple 3D detection and reconstruction datasets~(ScanNet++\cite{yeshwanth2023scannet++}, Omni3D~\cite{brazil2023omni3d}, WildRGB-D~\cite{xia2024WildRGB-D}, PointOdyssey~\cite{zheng2023pointodyssey}, and CA-1M~\cite{lazarow2024cubify}). 
We then construct open-ended QA pairs by combining predefined question templates with LLM-based rewriting~({\em e.g.}, prompting DeepSeek-v3~\cite{liu2024deepseekv3} to transform basic questions into more diverse formulations), thereby enriching the linguistic variety of questions.

To support convenient quantitative evaluation, we further convert a subset of these open-ended QA pairs into judgment ({\em i.e.}, Yes/No) and multi-choice formats, using three strategies to generate plausible and challenging distractors:
(i) randomly sampling same-category annotations ({\em e.g.}, depth or distance) within a suitable numeric range from the same or different scenes;
(ii) introducing small perturbations within justifiable margins of the ground-truth values ({\em e.g.}, homography matrices); and (iii) using DeepSeek-v3~\cite{liu2024deepseekv3} to synthesize confusing yet valid distractors. 
This process yields \textbf{2.3K} high-quality QA samples spanning judgment, multi-choice, and open-ended formats.

\input{tables/benchmark_comparison}

\vspace{2pt} 
\noindent \textbf{Data Integration \& Curation.}
We further integrate diverse spatial intelligence evaluation samples from existing datasets, including those focused on cognitive psychology~(SRBench~\cite{MindTheGap}, SpatialEval~\cite{SpatialEval}, SpatialViz-Bench~\cite{SpatialViz-Bench}); 
2D/3D spatial relations and distance reasoning~(SpatialSense~\cite{yang2019spatialsense}, VSR~\cite{VSR}, SpatialBench~\cite{cai2024spatialbot}, QSpatialBench~\cite{liao2024QSpatialBench}, 3DSRBench~\cite{ma20243dsrbench}, VSI-Bench~\cite{VSI-Bench}, Space-10~\cite{Space-10}, MIRAGE~\cite{MIRAGE}, RoboSpatial~\cite{RoboSpatial}, STI-Bench~\cite{STI-Bench}, VLM4D~\cite{VLM4D}, SITE-Bench~\cite{SITE}, SPAR-Bench~\cite{SPAR}, MMSI-Bench~\cite{MMSI-Bench}, OmniSpatial~\cite{OmniSpatial}); and spatially relevant subsets from general QA benchmarks (CV-Bench~\cite{tong2024cambrian}, MMVP~\cite{MMVP}, BLINK~\cite{fu2024blink}, MMIU~\cite{MMIU}, RealWorldQA~\cite{xai2023realworldQA}).

We combine these existing samples with our repurposed data, yielding \textbf{63,857} candidates. 
To ensure data quality and genuine visual dependency, we use a strong LLM~(GPT-OSS-120B~\cite{agarwal2025gpt-oss}) to filter out questions that can be answered without visual information, reducing the pool to \textbf{40,238} candidates.
Subsequently, five volunteers manually select 8,793 valid samples from these candidates based on the following criteria: 
(i) filtering out erroneous samples or those with low relevance to spatial reasoning;
(ii) eliminating duplicate samples;
and
(iii) ensuring category balance.
Through meticulous manual verification and reclassification by question type, we ultimately curate \textbf{5,025} high-quality, approximately balanced samples~(including 1,091 newly introduced ones), spanning \textbf{30} tasks that constitute our \textbf{SpatialScore} benchmark.
We further group these tasks into \textbf{10} intuitive categories based on their characteristics: mental animation, counting, depth estimation, object distance, object motion, camera pose \& motion, temporal reasoning, view reasoning, object size, and object localization. 

\input{tables/quantitative_results}

\subsection{Statistics \& Discussion}
\label{subsec:data_statistics}
We present the category-specific data distributions of \textbf{SpatialScore} and \textbf{SpatialCorpus}~(which will be further detailed in Sec.~\ref{subsec:supervised_fine-tuning}) in Fig.~\ref{fig:dataset}(b). 
Moreover, we provide comparisons with representative spatial intelligence benchmarks in Tab.~\ref{tab:benchmark_comparison} to demonstrate the completeness and diversity of our curated data. 
Concretely, our benchmark covers multiple data types~(real-world, simulated, and AI-generated content), diverse input modalities~(single images, multi-frame sequences, and videos), and question formats~(multi-choice, judgment, and open-ended QA).
More details are provided in Sec.~\ref{sec:additional_data_details} of the \textbf{Appendix}.

\subsection{Comparisons of Representative Models on SpatialScore}
\label{subsec:comparison_on_spatialscore}

To thoroughly assess spatial reasoning abilities, we conduct extensive experiments on our proposed SpatialScore across \textbf{49} representative MLLMs spanning diverse scales, including: general MLLMs, such as InternVL series~\cite{InternVL2.5, InternVL3, InternVL3.5}, Qwen series~\cite{Qwen2.5-VL, Qwen3-VL},
Kimi-VL~\cite{team2025kimivl}, LLaVA-1.5~\cite{liu2024llava1.5},
LLaVA-OneVision~\cite{li2025llavaonevision}, LLaMA-3.2V~\cite{grattafiori2024llama3}, and LLaMA-3.2V-CoT~\cite{xu2024llama-cot}, as well as models specifically fine-tuned for spatial understanding: SpaceQwen2.5VL~\cite{chen2024spatialvlm}, SpaceThinker~\cite{chen2024spatialvlm}, SpatialThinker~\cite{batra2025spatialthinker}, VST~\cite{yang2025VisualSpatialTuning}, Spatial-MLLM~\cite{Spatial-MLLM}, SpaceR~\cite{SpaceR}, SpatialReasoner~\cite{SpatialReasoner}, and SpaceLLaVA~\cite{chen2024spatialvlm}.
Moreover, proprietary models such as GPT-5~\cite{openai_gpt5_systemcard}, Gemini~\cite{comanici2025gemini, Gemini3}, and Claude-4.5-Sonnet~\cite{claude45} are also included.

Tab.~\ref{tab:quantitative_results} reports quantitative results of representative MLLMs on the proposed SpatialScore benchmark, from which we derive four key observations:
(i) \textbf{overall performance}. 
Gemini-3-Pro~\cite{Gemini3} achieves the highest overall score~(\textbf{60.12}), while Qwen3-VL-235B-A22B~\cite{Qwen3-VL} leads among open-source models~(\textbf{56.63}), substantially narrowing the gap with proprietary systems.
Despite these advances, there remains a large margin (\textbf{26.48}) between current state-of-the-art models and human-level spatial understanding (\textbf{86.60});
(ii) \textbf{model scale vs.~performance}. 
Larger models generally exhibit stronger performance, as observed in both the InternVL~\cite{InternVL2.5, InternVL3, InternVL3.5} and Qwen-VL~\cite{Qwen2.5-VL, Qwen3-VL} families. 
This trend suggests that stronger intrinsic reasoning ability can translate into improved spatial intelligence;
(iii) \textbf{limitations of existing fine-tuning}. 
Surprisingly, with the exception of VST~\cite{yang2025VisualSpatialTuning}, which demonstrates significant improvements in spatial understanding, most models fine-tuned on spatial-specific data ({\em e.g.}, SpaceThinker~\cite{chen2024spatialvlm}, SpatialThinker~\cite{batra2025spatialthinker}, and SpatialReasoner~\cite{SpatialReasoner}) deliver only marginal gains and sometimes even underperform their base models ({\em e.g.}, Qwen2.5-VL~\cite{Qwen2.5-VL}). 
This limited generalization underscores the diversity and difficulty of SpatialScore and indicates that current fine-tuning strategies and datasets for spatial understanding are still partial and insufficient;
(iv) \textbf{limitations of current models}. 
Although some models achieve near-human performance on certain fundamental tasks such as {\em mental animation} and {\em object localization}, they still struggle markedly with {\em view reasoning}, {\em camera pose}, {\em motion analysis}, and {\em real-world 3D perception}. 
This discrepancy exposes a pronounced deficiency in realistic 3D understanding in contemporary MLLMs.

%% file: tables/benchmark_comparison.tex
\begin{table}[t]
    \caption{
        \textbf{Comparison with Existing Spatial Intelligence Benchmarks.} 
        Here, Real and AIGC denote real-world samples and data generated by visual generative models~({\em e.g.}, Cosmos~\cite{agarwal2025cosmos}), respectively.
        The considered input modalities include single-image, multi-image sequence, and video.
    }
        \label{tab:benchmark_comparison}
        % \vspace{-6pt}
    \centering
    \small
    \setlength{\tabcolsep}{0.12cm} % 调整列间距
    \renewcommand{\arraystretch}{1.0} % 调整行高
    \resizebox{\textwidth}{!}{
        \begin{tabular}{l|c|ccc|ccc|ccc|c|c}
        \toprule
        \multirow{2}{*}{\textbf{Dataset}} & \multirow{2}{*}{\textbf{Publication}} & \multicolumn{3}{c|}{\textbf{Data Types}} & \multicolumn{3}{c|}{\textbf{Input Modalities}} & \multicolumn{3}{c|}{\textbf{QA Formats}} & \multirow{2}{*}{\textbf{\#Tasks}} & \multirow{2}{*}{\textbf{\#Samples}} \\
        \cline{3-5} \cline{6-8} \cline{9-11}
        & & \textbf{Real} & \textbf{Simulated} & \textbf{AIGC} & \textbf{Image} & \textbf{Sequence} & \textbf{Video} & \textbf{MCQ} & \textbf{Yes/No} & \textbf{Open} & & \\
        \midrule
        QSpatialBench~\cite{liao2024QSpatialBench} & EMNLP 2024 & \textcolor{green}{\Checkmark} & \textcolor{red}{\XSolidBrush} & \textcolor{red}{\XSolidBrush} & \textcolor{green}{\Checkmark} & \textcolor{red}{\XSolidBrush} & \textcolor{red}{\XSolidBrush} & \textcolor{red}{\XSolidBrush} & \textcolor{red}{\XSolidBrush} & \textcolor{green}{\Checkmark} & 2 & 271 \\
        SpatialEval~\cite{SpatialEval} & NeurIPS 2024 & \textcolor{green}{\Checkmark} & \textcolor{green}{\Checkmark} & \textcolor{red}{\XSolidBrush} & \textcolor{green}{\Checkmark} & \textcolor{red}{\XSolidBrush} & \textcolor{red}{\XSolidBrush} & \textcolor{green}{\Checkmark} & \textcolor{green}{\Checkmark} & \textcolor{red}{\XSolidBrush} & 4 & 4,635 \\
        VSI-Bench~\cite{VSI-Bench} & CVPR 2025 & \textcolor{green}{\Checkmark} & \textcolor{red}{\XSolidBrush} & \textcolor{red}{\XSolidBrush} & \textcolor{red}{\XSolidBrush} & \textcolor{red}{\XSolidBrush} & \textcolor{green}{\Checkmark} & \textcolor{green}{\Checkmark} & \textcolor{red}{\XSolidBrush} & \textcolor{green}{\Checkmark} & 8 & 5,156 \\
        RoboSpatial-Home~\cite{RoboSpatial} & CVPR 2025 & \textcolor{green}{\Checkmark} & \textcolor{red}{\XSolidBrush} & \textcolor{red}{\XSolidBrush} & \textcolor{green}{\Checkmark} & \textcolor{red}{\XSolidBrush} & \textcolor{red}{\XSolidBrush} & \textcolor{red}{\XSolidBrush} & \textcolor{green}{\Checkmark} & \textcolor{green}{\Checkmark} & 3 & 350 \\
        3DSRBench~\cite{ma20243dsrbench} & ICCV 2025 & \textcolor{green}{\Checkmark} & \textcolor{green}{\Checkmark} & \textcolor{red}{\XSolidBrush} & \textcolor{green}{\Checkmark} & \textcolor{red}{\XSolidBrush} & \textcolor{red}{\XSolidBrush} & \textcolor{green}{\Checkmark} & \textcolor{green}{\Checkmark} & \textcolor{red}{\XSolidBrush} & 4 & 2,772 \\
        STI-Bench~\cite{STI-Bench} & ICCV 2025 & \textcolor{green}{\Checkmark} & \textcolor{red}{\XSolidBrush} & \textcolor{red}{\XSolidBrush} & \textcolor{red}{\XSolidBrush} & \textcolor{red}{\XSolidBrush} & \textcolor{green}{\Checkmark} & \textcolor{green}{\Checkmark} & \textcolor{red}{\XSolidBrush} & \textcolor{red}{\XSolidBrush}   & 8 & 2,064 \\
        VLM4D~\cite{VLM4D} & ICCV 2025 & \textcolor{green}{\Checkmark} & \textcolor{red}{\XSolidBrush} & \textcolor{green}{\Checkmark} & \textcolor{red}{\XSolidBrush} & \textcolor{red}{\XSolidBrush} & \textcolor{green}{\Checkmark} & \textcolor{green}{\Checkmark} & \textcolor{green}{\Checkmark} & \textcolor{red}{\XSolidBrush} & 4 & 1,816 \\
        SITE-Bench~\cite{SITE} & ICCV 2025 & \textcolor{green}{\Checkmark} & \textcolor{green}{\Checkmark} & \textcolor{green}{\Checkmark} & \textcolor{green}{\Checkmark} & \textcolor{green}{\Checkmark} & \textcolor{green}{\Checkmark} & \textcolor{green}{\Checkmark} & \textcolor{red}{\XSolidBrush} & \textcolor{red}{\XSolidBrush} & 6 & 8,068 \\
        MIRAGE~\cite{MIRAGE} & NeurIPS 2025 & \textcolor{green}{\Checkmark} & \textcolor{red}{\XSolidBrush} & \textcolor{red}{\XSolidBrush} & \textcolor{green}{\Checkmark} & \textcolor{red}{\XSolidBrush} & \textcolor{red}{\XSolidBrush} & \textcolor{red}{\XSolidBrush} & \textcolor{green}{\Checkmark} & \textcolor{green}{\Checkmark} & 3 & 1,710 \\
        SPAR-Bench~\cite{SPAR} & NeurIPS 2025 & \textcolor{green}{\Checkmark} & \textcolor{green}{\Checkmark} & \textcolor{red}{\XSolidBrush} & \textcolor{green}{\Checkmark} & \textcolor{green}{\Checkmark} & \textcolor{green}{\Checkmark} & \textcolor{green}{\Checkmark} & \textcolor{green}{\Checkmark} & \textcolor{red}{\XSolidBrush} & 20 & 7,211 \\
        SpatialViz-Bench~\cite{SpatialViz-Bench} & ICLR 2026 & \textcolor{red}{\XSolidBrush} & \textcolor{green}{\Checkmark} & \textcolor{red}{\XSolidBrush} & \textcolor{green}{\Checkmark} & \textcolor{red}{\XSolidBrush} & \textcolor{red}{\XSolidBrush} & \textcolor{green}{\Checkmark} & \textcolor{red}{\XSolidBrush} & \textcolor{red}{\XSolidBrush} & 12 & 1,180 \\
        MMSI-Bench~\cite{MMSI-Bench} & ICLR 2026 & \textcolor{green}{\Checkmark} & \textcolor{green}{\Checkmark} & \textcolor{red}{\XSolidBrush} & \textcolor{red}{\XSolidBrush} & \textcolor{green}{\Checkmark} & \textcolor{green}{\Checkmark} & \textcolor{green}{\Checkmark} & \textcolor{red}{\XSolidBrush} & \textcolor{red}{\XSolidBrush} & 11 & 1,000 \\
        OmniSpatial~\cite{OmniSpatial} & ICLR 2026 & \textcolor{green}{\Checkmark} & \textcolor{green}{\Checkmark} & \textcolor{red}{\XSolidBrush} & \textcolor{green}{\Checkmark} & \textcolor{green}{\Checkmark} & \textcolor{green}{\Checkmark} & \textcolor{green}{\Checkmark} & \textcolor{green}{\Checkmark} & \textcolor{red}{\XSolidBrush} & 50 & 1,533 \\
        \midrule
        \rowcolor{cyan!10}
        \textbf{SpatialScore (Ours)} & & \textcolor{green}{\Checkmark} & \textcolor{green}{\Checkmark} & \textcolor{green}{\Checkmark} & \textcolor{green}{\Checkmark} & \textcolor{green}{\Checkmark} & \textcolor{green}{\Checkmark} & \textcolor{green}{\Checkmark} & \textcolor{green}{\Checkmark} & \textcolor{green}{\Checkmark} & \textbf{30} & \textbf{5,025} \\
        \bottomrule
        \end{tabular}
    }
    \vspace{-6pt}
\end{table}

%% file: tables/quantitative_results.tex
\begin{table}[ht!]
    \caption{
        \textbf{Results on SpatialScore.}
        Mental., Count., Depth., Obj-Dist., Obj-Mo., Camera., Temp-Rea., View-Rea., Obj-Size., Obj-Loc., refer to Mental Animation, Counting, Depth Estimation, Object Distance, Object Motion, Camera Pose \& Motion, Temporal Reasoning, View Reasoning, Object Size, and Object Localization, respectively.
        Best and second-best ones are \setlength{\fboxsep}{2pt}\colorbox{mygreen}{\textbf{bolded}} and \setlength{\fboxsep}{1.5pt}\colorbox{mylightgreen}{\underline{underlined}} in each group.
    }
    \label{tab:quantitative_results}
    \vspace{-2pt}
    \centering
    \setlength{\tabcolsep}{0.1cm}
    \renewcommand{\arraystretch}{1.0}
    \resizebox{\textwidth}{!}{
    \begin{tabular}{l|c|c|*{10}{c}}
    \toprule[1.5pt]
        \textbf{Methods} & \textbf{Overall} & \textbf{Rank} & \textbf{Mental.} & \textbf{Count.} & \textbf{Depth.} & \textbf{View-Rea.} & \textbf{Obj-Size.} &  \textbf{Obj-Loc.} & \textbf{Obj-Dist.} & \textbf{Obj-Mo.} & \textbf{Camera.} & \textbf{Temp-Rea.}   \\
        \midrule[0.8pt]
        \rowcolor{cyan!10}
        \multicolumn{13}c{\textbf{\textit{Baselines}}} \\
        \midrule
        Blind GPT-5~(Text-only)~\cite{cai2025HasGPT5}
        & 30.62 & - & 18.79 & 20.34 & 29.36 & 40.81 & 31.05 & 45.34 & 24.20 & 25.54 & 32.01 & 26.47 \\
        Chance-level~(Random) & 28.29 & - & 23.71 & 22.80 & 22.88 & 31.84 & 30.29 & 38.59 & 24.06 & 25.06 & 28.79 & 28.68 \\
        Human-level & \textbf{86.60} & - & \textbf{96.87} & \textbf{89.72} & \textbf{82.33} & \textbf{92.15} & \textbf{85.18} & \textbf{81.64} & \textbf{78.96} & \textbf{94.46} & \textbf{86.89} & \textbf{84.19} \\
        \midrule
        \rowcolor{cyan!10}
        \multicolumn{13}c{\textbf{\textit{Small-scale models}~(1B$\sim$4B)}} \\
        \midrule
        InternVL2.5-1B~\cite{InternVL2.5} & 31.66 & 13 & 30.20 & 38.42 & 22.26 & 32.29 & 32.29 & 51.08 & 25.12 & 26.02 & 27.76 & 25.74  \\
        InternVL3-1B~\cite{InternVL3} & 33.03 & 12 & 26.85 & 47.69 & 24.74 & 32.06 & 34.06 & 57.53 & 24.02 & 32.29 & 25.71 & 19.85 \\
        InternVL3.5-1B~\cite{InternVL3.5} & 33.55 & 11 & 31.10 & 40.27 & 33.47 & 30.72 & 30.95 & 53.08 & 26.72 & 24.82 & 29.43 & 29.41 \\
        Qwen3-VL-2B~\cite{Qwen3-VL} & 41.41 & 5 & \cellcolor{mylightgreen}\underline{35.35} & \cellcolor{mygreen}\textbf{52.74} & 34.64 & 35.65 & 49.69 & \cellcolor{mylightgreen}\underline{61.69} & 35.42 & 38.80 & 30.59 & 39.34 \\
        Qwen2.5-VL-3B~\cite{Qwen2.5-VL} & 36.92 & 8 & 32.21 & 46.07 & 33.07 & 34.53 & 38.48 & 56.53 & 29.73 & 37.59 & 30.59 & 24.26 \\
        SpaceQwen2.5VL-3B~\cite{chen2024spatialvlm} & 31.25 & 14 & 20.81 & 47.74 & 22.51 & 32.96 & 38.82 & 39.45 & 25.13 & 30.60 & 29.95 & 24.26 \\
        SpaceThinker-3B~\cite{chen2024spatialvlm} & 34.47 & 10 & 27.96 & 48.30 & 34.64 & 34.08 & 36.89 & 44.76 & 28.07 & 35.42 & 28.66 & 26.84 \\
        SpatialThinker-3B~\cite{batra2025spatialthinker} & 37.16 & 7 & 29.75 & 50.79 & 37.45 & 37.22 & 31.69 & 58.97 & 31.87 & 35.18 & 30.08 & 22.79 \\
        VST-3B-SFT~\cite{yang2025VisualSpatialTuning} & \cellcolor{mylightgreen}\underline{47.37} & \cellcolor{mylightgreen}\underline{2} & 25.73 & 48.45 & \cellcolor{mygreen}\textbf{53.11} & \cellcolor{mygreen}\textbf{45.74} & 53.02 & 61.41 & \cellcolor{mylightgreen}\underline{42.35} & \cellcolor{mylightgreen}\underline{60.48} & \cellcolor{mylightgreen}\underline{35.35} & \cellcolor{mylightgreen}\underline{50.74}  \\
        VST-3B-RL~\cite{yang2025VisualSpatialTuning} & \cellcolor{mygreen}\textbf{48.04} & \cellcolor{mygreen}\textbf{1} & 24.61 & \cellcolor{mylightgreen}\underline{52.38} & \cellcolor{mylightgreen}\underline{51.75} & \cellcolor{mylightgreen}\underline{43.72} & \cellcolor{mylightgreen}\underline{54.16} & \cellcolor{mygreen}\textbf{61.84} & \cellcolor{mygreen}\textbf{43.03} & \cellcolor{mygreen}\textbf{62.89} & \cellcolor{mygreen}\textbf{38.95} & 47.43  \\
        Spatial-MLLM-4B~\cite{Spatial-MLLM} & 42.59 & 3 & 33.56 & 47.61 & 34.80 & 43.27 & \cellcolor{mygreen}\textbf{56.16} & 57.68 & 40.49 & 36.39 & 28.41 & \cellcolor{mygreen}\textbf{53.31}  \\
        Qwen3-VL-4B~\cite{Qwen3-VL} & 42.52 & 4 & \cellcolor{mygreen}\textbf{37.81} & 48.22 & 37.68 & 34.75 & 48.20 & 59.40 & 33.40 & 53.01 & 34.06 & 38.24 \\
        InternVL2.5-4B~\cite{InternVL2.5} & 36.18 & 9 & 27.52 & 45.15 & 41.25 & 35.65 & 35.43 & 57.25 & 29.23 & 29.64 & 29.18 & 23.53 \\
        InternVL3.5-4B~\cite{InternVL3.5} & 37.81 & 6 & 34.23 & 40.40 & 40.97 & 34.75 & 29.59 & 60.11 & 31.15 & 38.07 & 30.46 & 34.19 \\
        \midrule
        \rowcolor{cyan!10}
        \multicolumn{13}c{\textbf{\textit{Middle-scale models}~(7B$\sim$14B)}} \\
        \midrule
        LLaVA-1.5-7B~\cite{liu2024llava1.5} & 25.32 & 19 & 20.58 & 21.64 & 29.34 & 22.42 & 22.25 & 47.20 & 14.43 & 24.58 & 27.63 & 2.21 \\
        LLaVA-OneVision-7B~\cite{li2025llavaonevision} & 36.57 & 15 & 25.50 & 45.97 & 38.92 & 36.32 & 42.31 & 56.67 & 29.01 & 29.40 & 32.13 & 16.18 \\
        Qwen2.5-VL-7B~\cite{Qwen2.5-VL} & 40.53 & 9 & 42.06 & 48.67 & 43.44 & 37.00 & 40.88 & 57.82 & 34.02 & 43.61 & 28.66 & 26.84 \\
        SpaceR-7B~\cite{SpaceR} & 43.36 & 6 & \cellcolor{mylightgreen}\underline{44.74} & 51.13 & 45.31 & 37.44 & 48.48 & 58.39 & 35.98 & 45.06 & 33.03 & 31.62 \\
        VST-7B-SFT~\cite{yang2025VisualSpatialTuning} & \cellcolor{mylightgreen}\underline{50.06} & \cellcolor{mylightgreen}\underline{2} & 39.37 & \cellcolor{mylightgreen}\underline{54.00} & \cellcolor{mylightgreen}\underline{53.67} & \cellcolor{mylightgreen}\underline{43.05} & \cellcolor{mylightgreen}\underline{52.91} & 63.27 & \cellcolor{mylightgreen}\underline{46.44} & \cellcolor{mylightgreen}\underline{62.65} & \cellcolor{mygreen}\textbf{39.72} & \cellcolor{mygreen}\textbf{45.96}  \\
        VST-7B-RL~\cite{yang2025VisualSpatialTuning} & \cellcolor{mygreen}\textbf{52.47} & \cellcolor{mygreen}\textbf{1} & 41.39 & \cellcolor{mygreen}\textbf{54.92} & \cellcolor{mygreen}\textbf{56.80} & \cellcolor{mygreen}\textbf{50.00} & \cellcolor{mygreen}\textbf{55.48} & \cellcolor{mygreen}\textbf{64.56} & \cellcolor{mygreen}\textbf{48.65} & \cellcolor{mygreen}\textbf{63.37} & \cellcolor{mygreen}\textbf{42.80} & \cellcolor{mylightgreen}\underline{45.59}  \\
        SpatialThinker-7B~\cite{batra2025spatialthinker} & 40.83 & 8 & 39.82 & 49.46 & 44.41 & 34.75 & 41.88 & 59.40 & 33.13 & 44.34 & 31.36 & 23.90 \\
        SpatialReasoner-8B~\cite{SpatialReasoner} & 36.31 & 16 & 36.02 & 41.21 & 36.12 & 36.32 & 34.63 & 53.23 & 28.00 & 33.73 & 31.75 & 26.10 \\
        InternVL2.5-8B~\cite{InternVL2.5} & 39.63 & 11 & 40.27 & 45.85 & 45.39 & 35.43 & 45.25 & 58.82 & 34.07 & 27.71 & 28.53 & 28.31 \\
        InternVL3-8B~\cite{InternVL3} & 41.57 & 7 & 37.81 & 51.95 & 39.46 & 38.79 & 43.05 & \cellcolor{mylightgreen}\underline{64.13} & 36.02 & 41.69 & 29.95 & 28.31 \\
        InternVL3.5-8B~\cite{InternVL3.5} & 37.75 & 12 & 38.48 & 42.64 & 41.25 & 31.39 & 33.29 & 60.83 & 32.27 & 36.39 & 32.39 & 13.60 \\
        Qwen3-VL-8B~\cite{Qwen3-VL} & 45.48 & 3 & 38.26 & 50.35 & 47.12 & 37.67 & 52.12 & 61.12 & 37.22 & 55.90 & 34.19 & 41.54 \\
        LLaMA-3.2V-11B~\cite{grattafiori2024llama3} & 35.13 & 17 & 40.72 & 47.81 & 38.96 & 36.55 & 25.98 & 51.94 & 25.33 & 33.98 & 29.05 & 17.28 \\
        LLaMA-3.2V-11B-CoT~\cite{xu2024llama-cot} & 37.51 & 13 & 28.64 & 39.71 & 40.51 & 35.87 & 47.98 & 54.52 & 31.84 & 31.77 & 29.18 & 25.74 \\
        LLaVA-1.5-13B~\cite{liu2024llava1.5} & 26.67 & 18 & 25.06 & 25.91 & 31.52 & 20.63 & 23.59 & 47.63 & 16.10 & 26.51 & 28.15 & 1.84 \\
        SpaceLLaVA-13B~\cite{chen2024spatialvlm} & 23.97 & 20 & 34.23 & 32.14 & 26.93 & 22.65 & 19.35 & 36.01 & 15.13 & 19.52 & 15.30 & 23.16 \\
        InternVL3-14B~\cite{InternVL3} & 44.89 & 4 & 44.52 & 53.05 & 42.92 & 37.89 & 48.86 & 62.84 & 40.86 & 43.37 & 35.35 & 35.29 \\
        InternVL3.5-14B~\cite{InternVL3.5} & 43.88 & 5 & \cellcolor{mygreen}\textbf{45.64} & 46.63 & 41.66 & 40.36 & 46.95 & \cellcolor{mygreen}\textbf{64.56} & 36.82 & 42.65 & 35.60 & 29.04 \\
        Kimi-VL-16B-A3B~\cite{team2025kimivl} & 40.10 & 10 & 29.08 & 53.32 & 38.25 & 41.03 & 45.17 & 59.11 & 35.85 & 43.61 & 27.38 & 25.74 \\
        Kimi-VL-A3B-Thinking~\cite{team2025kimivl} & 37.06 & 14 & 30.20 & 40.70 & 32.03 & 35.43 & 45.74 & 52.37 & 28.22 & 32.77 & 33.16 & 33.82 \\
        \midrule
        \rowcolor{cyan!10}
        \multicolumn{13}c{\textbf{\textit{Large-scale models}~(30B$\sim$78B)}} \\
        \midrule
        Qwen3-VL-30B-A3B~\cite{Qwen3-VL} & 50.71 & 3 & 46.31 & 58.86 & 48.49 & \cellcolor{mylightgreen}\underline{47.98} & \cellcolor{mylightgreen}\underline{56.52} & 66.43 & 40.73 & 59.52 & 39.20 & 45.59 \\
        Qwen2.5-VL-32B~\cite{Qwen2.5-VL} & 47.23 & 8 & 45.41 & 50.38 & 52.28 & 39.46 & 49.25 & 61.84 & 40.35 & 56.14 & 40.23 & 29.04 \\
        Qwen3-VL-32B~\cite{Qwen3-VL} & \cellcolor{mylightgreen}\underline{54.11} & \cellcolor{mylightgreen}\underline{2} & 43.40 & \cellcolor{mylightgreen}{\underline{61.16}} & 51.80 & 50.22 & 59.53 & \cellcolor{mylightgreen}\underline{69.58} & \cellcolor{mygreen}\textbf{50.94} & \cellcolor{mylightgreen}\underline{66.75} & 41.26 & \cellcolor{mylightgreen}\underline{47.79} \\
        InternVL2.5-38B~\cite{InternVL2.5} & 45.60 & 10 & 43.62 & 52.98 & 52.71 & 37.67 & 50.47 & 62.70 & 39.16 & 50.84 & 35.09 & 21.69 \\
        InternVL3-38B~\cite{InternVL3} & 48.95 & 5 & 45.64 & 58.06 & 49.10 & 42.15 & 50.61 & 67.58 & 41.04 & 57.83 & 39.46 & 33.82 \\
        InternVL3.5-38B~\cite{InternVL3.5} & 45.74 & 9 & 40.27 & 43.98 & 48.58 & 40.13 & 52.95 & 62.84 & 39.31 & 48.19 & 39.72 & 29.04 \\
        LLaVA-OneVision-72B~\cite{li2025llavaonevision} & 43.29 & 11 & 37.58 & 51.57 & 51.45 & 38.57 & 50.24 & 62.12 & 31.84 & 37.59 & 38.43 & 19.49 \\
        Qwen2.5-VL-72B~\cite{Qwen2.5-VL} & 48.42 & 7 & \cellcolor{mylightgreen}\underline{53.69} & 49.49 & \cellcolor{mygreen}\textbf{56.23} & 36.10 & 50.91 & 62.55 & 38.42 & 59.52 & \cellcolor{mygreen}\textbf{42.93} & 22.43 \\
        InternVL2.5-78B~\cite{InternVL2.5} & 48.71 & 6 & 51.45 & 54.15 & \cellcolor{mylightgreen}\underline{54.61} & 43.50 & 53.68 & 62.84 & 41.67 & 52.05 & 39.20 & 25.74 \\
        InternVL3-78B~\cite{InternVL3} & 50.67 & 4 & 50.34 & 59.19 & 48.74 & 45.74 & 50.32 & 65.71 & 42.50 & 60.48 & 41.52 & 43.75 \\
        Qwen3-VL-235B-A22B~\cite{Qwen3-VL} & \cellcolor{mygreen}\textbf{56.63} & \cellcolor{mygreen}\textbf{1} & \cellcolor{mygreen}\textbf{57.27} & \cellcolor{mygreen}\textbf{65.19} & 54.04 & \cellcolor{mygreen}\textbf{52.47} & \cellcolor{mygreen}\textbf{59.90} & \cellcolor{mygreen}\textbf{70.01} & \cellcolor{mylightgreen}\underline{50.40} & \cellcolor{mygreen}\textbf{69.40} & \cellcolor{mylightgreen}\underline{42.80} & \cellcolor{mygreen}\textbf{49.63} \\
        \rowcolor{cyan!10}
        \midrule
        \multicolumn{13}c{\textbf{\textit{Proprietary Models~(Commercial APIs)}}} \\
        \midrule
        Claude-4.5-Sonnet~\cite{claude45} & 45.68 & 4 & 51.01 & 49.67 & 47.33 & 39.91 & 54.08 & 53.66 & 38.68 & 46.99 & 36.12 & 41.18 \\
        Gemini-2.5-Pro~\cite{comanici2025gemini} & 56.37 & 3 & \cellcolor{mylightgreen}{73.29} & \cellcolor{mygreen}\textbf{64.04} & 51.15 & 46.41 & 58.24 & \cellcolor{mylightgreen}{66.52} & \cellcolor{mylightgreen}\underline{46.93} & \cellcolor{mylightgreen}\underline{57.84} & 48.78 & 56.25 \\
        Gemini-3-Pro~\cite{Gemini3} & \cellcolor{mygreen}\textbf{60.12} & \cellcolor{mygreen}\textbf{1} & 70.40 & \cellcolor{mylightgreen}\underline{62.99} & \cellcolor{mygreen}\textbf{58.15} & \cellcolor{mygreen}\textbf{62.44} & \cellcolor{mygreen}\textbf{60.02} & 64.20 & \cellcolor{mygreen}\textbf{49.50} & \cellcolor{mygreen}\textbf{71.12} & \cellcolor{mygreen}\textbf{51.37} & \cellcolor{mylightgreen}\underline{59.93} \\
        GPT-5~\cite{openai_gpt5_systemcard} & \cellcolor{mylightgreen}\underline{58.13} & \cellcolor{mylightgreen}\underline{2} & \cellcolor{mygreen}\textbf{78.08} & 57.59 & \cellcolor{mylightgreen}\underline{55.13} & \cellcolor{mylightgreen}\underline{54.04} & \cellcolor{mylightgreen}\underline{59.22} & \cellcolor{mygreen}\textbf{67.39} & 45.01 & 57.11 & \cellcolor{mylightgreen}\underline{50.90} & \cellcolor{mygreen}\textbf{62.13} \\
    \bottomrule[1.5pt]
    \end{tabular}
    }
    % \vspace{-6pt}
\end{table}

%% file: sections/3_method.tex
\section{Methodology}
\label{sec:Methodology}
This section investigates two complementary pathways for improving spatial understanding in MLLMs, from a direct data-driven approach to an agent-centric solution. Sec.~\ref{subsec:problem_formulation} first formalizes the problem scope.
Sec.~\ref{subsec:supervised_fine-tuning} then presents a data-driven solution based on supervised fine-tuning with our constructed \textbf{SpatialCorpus}.
Next, Sec.~\ref{subsec:SpatialAgent} describes our \textbf{SpatialAgent} multi-agent system, and Sec.~\ref{subsec:Toolbox} details the suite of spatial perception tools integrated into this framework.

\subsection{Problem Formulation}
\label{subsec:problem_formulation}

We adopt a question-answering paradigm to assess and improve spatial understanding. 
Given a textual question $\mathbf{q}$ and a visual input $\mathbf{v}$~(a single image, a multi-frame sequence, or a video), the basic MLLM-based QA process is formulated as:
\begin{align}
    \mathbf{r} = \Phi(\mathbf{q}, \mathbf{v})
\end{align}
where $\Phi(\cdot)$ denotes an MLLM and $\mathbf{r}$ represents the free-text response. 
In the data-driven pathway, we enhance the model's capabilities via supervised fine-tuning on our \textbf{SpatialCorpus}; the inference form remains the same, but $\hat{\Phi}$ denotes the fine-tuned model with updated parameters.

In parallel, an alternative pathway keeps the off-the-shelf pretrained MLLM and augments it at inference time with an agentic framework, \textbf{SpatialAgent}~($\mathcal{A}$). To be specific, our multi-agent system coordinates external spatial tools during reasoning, and the core process is expressed as:
\begin{align}
    \mathbf{r} = \mathcal{A}(\mathbf{q}, \mathbf{v}; \Phi; \mathcal{T})
\end{align}
where $\mathcal{T} = \{\mathbf{t}_1, \mathbf{t}_2, \dots, \mathbf{t}_n\}$ is a toolbox of specialized spatial perception tools (each $\mathbf{t}_i$ is a distinct tool), detailed in Sec.~\ref{subsec:Toolbox}. 
This formulation highlights the two complementary pathways: directly strengthening the backbone model through data, or keeping it fixed and augmenting it with tool-based reasoning.

\subsection{Supervised Fine-tuning}
\label{subsec:supervised_fine-tuning}
A fundamental but effective approach to enhancing spatial understanding is supervised fine-tuning~(SFT) on domain-specific data.
Formally, each training sample is represented as a multimodal triplet~($\mathbf{v}, \mathbf{q}, \mathbf{r}$), comprising a visual scene~($\mathbf{v}$) and a spatially relevant question~($\mathbf{q}$) as inputs, as well as a corresponding ground-truth answer~($\mathbf{r}$).
The training objective is to optimize the MLLM by minimizing the discrepancy between generated and reference responses.
This optimization process refines the model's reasoning capabilities across key dimensions such as \textit{position}, \textit{distance}, and \textit{camera transformation}, thereby obtaining a fine-tuned model~($\hat{\Phi}$) specialized for spatial intelligence.

To support this data-driven approach, we develop an automated data curation pipeline to build \textbf{SpatialCorpus}, which integrates both real-world and simulated environments, covering single-frame and multi-frame inputs with diverse question-answering formats, including multi-choice, judgment, and open-ended QA. 
Specifically, as illustrated in Fig.~\ref{fig:dataset}(a), we adopt the same data repurposing strategy introduced in Sec.~\ref{subsec:dataset_construction} to construct training data~(ensuring no overlap with the test distribution). 
Regarding data selection, we exclude CA-1M~\cite{lazarow2024cubify} from this phase as it only contains class-agnostic annotations. 
To mitigate the high cost of large-scale LLM rephrasing, we employ a diverse set of rule-based templates as a cost-effective alternative.
Moreover, to bolster {\em mental animation} capabilities, we incorporate synthetic data rendered via simulators, such as spatial maps and 2D/3D rotation sequences.

Ultimately, these efforts yield our \textbf{SpatialCorpus}, which contains about 331K QA pairs across 16 tasks in 7 categories, providing necessary training resources for supervised fine-tuning on spatial understanding tasks. 
Leveraging this constructed resource, we fine-tune the leading open-source model, Qwen3-VL~\cite{Qwen3-VL} using standard cross-entropy loss, achieving consistent improvements across diverse spatial reasoning tasks~(as evidenced in Sec.~\ref{subsec:quantitative_results}).

\begin{figure*}[t]
  \centering
  \includegraphics[width=\textwidth]{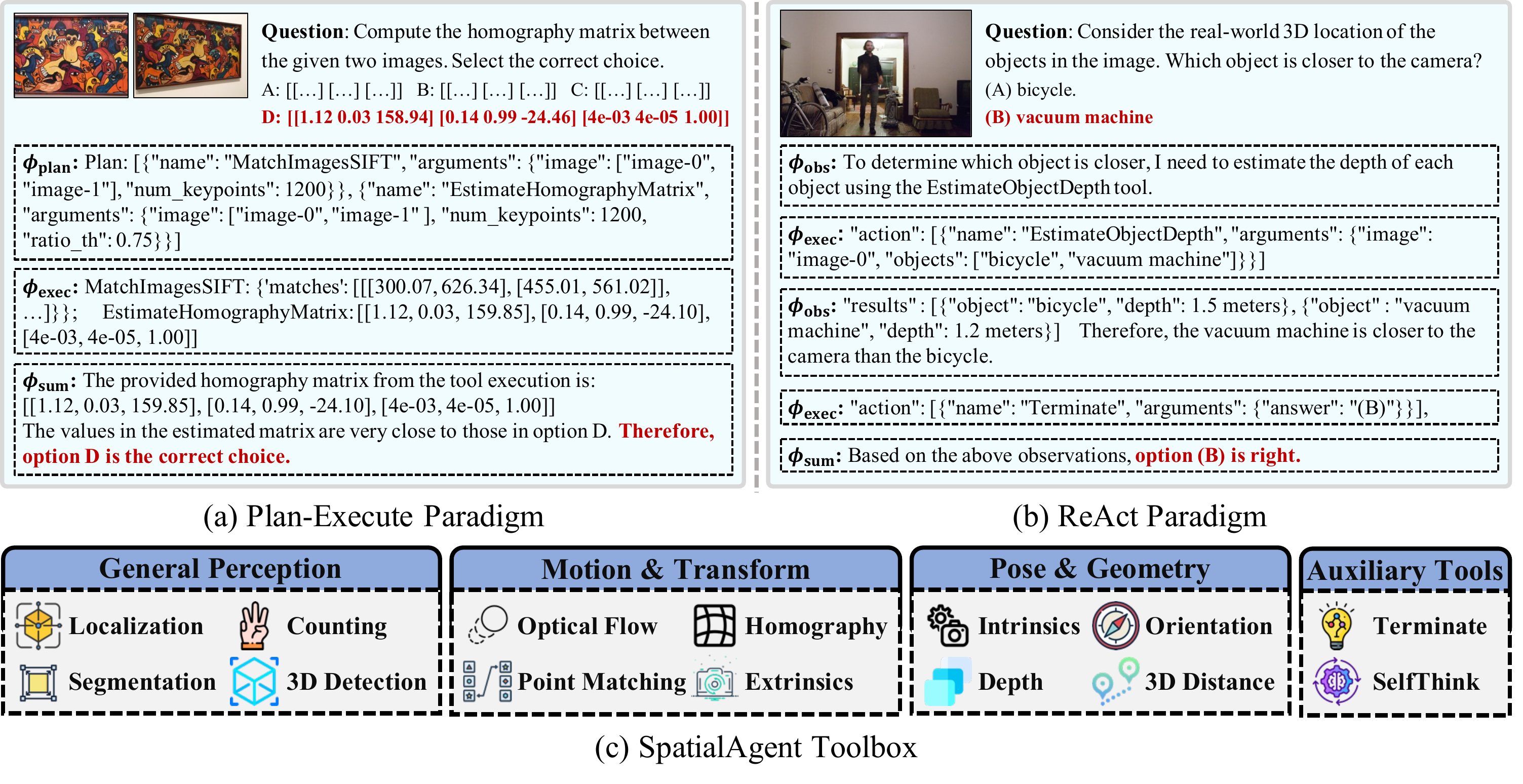}
  \vspace{-0.7cm}
  \caption{
      \textbf{Architecture and Workflow of SpatialAgent.}
      (a) Specialized spatial perception tools within SpatialAgent;
      (b) The {\em Plan-Execute} paradigm for task decomposition and stepwise execution;
      (c) The {\em ReAct} paradigm for iterative interaction and strategy refinement.
  }
 \label{fig:architecture}
 % \vspace{-6pt}
\end{figure*}

\subsection{SpatialAgent}
\label{subsec:SpatialAgent}
Given that supervised fine-tuning inevitably incurs high computational costs, overfitting, and potential catastrophic forgetting of general capabilities, we construct \textbf{SpatialAgent} as a more elegant, \textbf{training-free} alternative.
Specifically, we utilize meticulously designed prompts, guiding the agent core~($\Phi$) to fulfill distinct functional roles and execute two reasoning paradigms: {\em Plan-Execute}~({\em PE}) and {\em ReAct}, to improve spatial understanding abilities, as detailed below.

\vspace{2pt}
\noindent \textbf{Plan-Execute Paradigm.}
As depicted in Fig.~\ref{fig:architecture}(a), in this paradigm, our SpatialAgent comprises three components: {\em planner}, {\em executor}, and {\em summarizer}, expressed as $\mathcal{A}_{\mathrm{PE}} = \{\Phi_{\mathrm{plan}}, \Phi_{\mathrm{exec}}, \Phi_{\mathrm{sum}} \}$, which obtains the final response~($\mathbf{r}_{\mathrm{PE}}$) via a sequential feedforward process.
Given a question~($\mathbf{q}$) and visual input~($\mathbf{v}$), along with detailed specifications about the toolbox~($\mathcal{T}$), the planner~($\Phi_{\mathrm{plan}}$) first generates a plan for invoking tools~($\mathbf{p}$) of $k$ steps, each with a specific tool~($\mathbf{t}_i$) and its parameters~($\mathrm{args}_{i}$):
\begin{align}
    \mathbf{p} = \Phi_{\mathrm{plan}}(\mathbf{q}, \mathbf{v}; \mathcal{T}) = \{(\mathbf{t}_1, \mathrm{args}_1),  \dots, (\mathbf{t}_k, \mathrm{args}_k)\}
\end{align}
Then, the executor~($\Phi_{\mathrm{exec}}$) sequentially executes the plan~($\mathbf{p}$) and obtains the tool output set~($\mathcal{Y}$) consisting of results at each step~($\mathbf{y}_i$), denoted as:
\begin{align}
    \mathcal{Y} = \{\mathbf{y}_1, \mathbf{y}_2, \dots, \mathbf{y}_k\} = \Phi_{\mathrm{exec}}(\mathbf{p})
\end{align}
Finally, the summarizer~($\Phi_{\mathrm{sum}}$) produces the final response~($\mathbf{r}_{\mathrm{PE}}$) by reasoning according to the tool outputs~($\mathcal{Y}$) and original inputs~($\mathbf{q}$, $\mathbf{v}$), formulated as:
\begin{align}
    \mathbf{r}_{\mathrm{PE}} = \Phi_{\mathrm{sum}}(\mathcal{Y}, \mathbf{q}, \mathbf{v})
\end{align}
% 

% \vspace{2pt}
\noindent \textbf{ReAct Paradigm.}
As illustrated in Fig.~\ref{fig:architecture}(b), this paradigm adopts an interleaved reasoning process, with our SpatialAgent composed of {\em observer}, {\em executor}, and {\em summarizer}, denoted as $\mathcal{A}_{\mathrm{ReAct}} = \{\Phi_{\mathrm{obs}}, \Phi_{\mathrm{exec}}, \Phi_{\mathrm{sum}}\}$.
Here, we maintain a memory module~($\mathcal{M}$) that records all intermediate interactions between the observer~($\Phi_{\mathrm{obs}}$) and the executor~($\Phi_{\mathrm{exec}}$).
At step $i$, the memory state~($\mathcal{M}_{i}$) stores the complete history of observer decisions~($\mathbf{o}$) and execution results~($\mathbf{y}$), represented as:
\begin{align}
    \mathcal{M}_{i} = \{\mathbf{m}_1, \mathbf{m}_2, \dots \mathbf{m}_{i-1}\} = \{(\mathbf{o}_1, \mathbf{y}_1), (\mathbf{o}_2, \mathbf{y}_2), \dots, (\mathbf{o}_{i-1}, \mathbf{y}_{i-1}) \}, \,\,\,\, \mathrm{with} \,\, \mathcal{M}_1 = \varnothing 
\end{align}
At step $i$, the observer~($\Phi_{\mathrm{obs}}$) generates the next action~($\mathbf{o}_i$) based on the inputs~($\mathbf{q}$, $\mathbf{v}$) and the full interaction history~($\mathcal{M}_i$), while the executor~($\Phi_{\mathrm{exec}}$) processes accordingly, expressed as:
\begin{align}
    \mathbf{o}_i = \Phi_{\mathrm{obs}}(\mathcal{M}_i, \mathbf{q}, \mathbf{v}); \quad
    \mathbf{y}_i = \Phi_{\mathrm{exec}}(\mathbf{o}_i)
\end{align}
The iterative process continues until the observer~($\Phi_{\mathrm{obs}}$) outputs a {\em Terminate} action, triggering the summarization phase, where the summarizer~($\Phi_{\mathrm{sum}}$) generates the final response~($\mathbf{r}_{\mathrm{ReAct}}$) by consolidating all accumulated evidence in memory~($\mathcal{M}$) and the original inputs~($\mathbf{q}$, $\mathbf{v}$), denoted as:
\begin{align}
    \mathbf{r}_{\mathrm{ReAct}} = \Phi_{\mathrm{sum}}(\mathcal{M}, \mathbf{q}, \mathbf{v})
\end{align}
Each component within both paradigms is driven by carefully designed prompts~(as detailed in Sec.~\ref{subsec:spatialagent_development} of the \textbf{Appendix}), with distinct characteristics:
the {\em Plan-Execute} paradigm excels at efficient plan formulation and execution, though its predetermined path may sacrifice precision in complex scenarios. 
Conversely, the {\em ReAct} paradigm demonstrates better flexibility through dynamic planning that adapts to intermediate outputs, albeit at the cost of reduced efficiency due to its iterative nature.

\subsection{Toolbox}
\label{subsec:Toolbox}
As depicted in Fig.~\ref{fig:architecture}(c), SpatialAgent integrates a comprehensive toolbox~($\mathcal{T}$) with 12 specialized spatial perception tools, which are organized into four main categories: {\em general perception}, {\em motion} \& {\em transformation}, {\em pose} \& {\em geometry}, and {\em auxiliary tools}. 
Each tool is specified with clearly defined functionality, input/output formats, and usage examples. 
Notably, our toolbox exclusively employs \textbf{open-source} models, ensuring easy reproduction and continuous improvement as these underlying tools evolve.

\vspace{2pt}
\noindent \textbf{General Perception.}
To endow our framework with general perception abilities, 
we include a suite of open-set visual perception models. 
Concretely, Rex-Omni~\cite{Rex-Omni} is used for object counting and localizing objects with precise bounding boxes. 
These detections can serve as visual prompts for the segmentation tool SAM2~\cite{SAM2}, which segments instances to refine localization and quantify object proportions. 
Moreover, we employ DetAny3D~\cite{DetAny3D} as a 3D object detection tool to extract 3D bounding boxes.

\vspace{2pt}
\noindent \textbf{Motion \& Transformation.}
To understand dynamics within multiple frames and videos, we integrate RAFT~\cite{teed2020raft} for dense optical flow estimation.
This facilitates camera motion analysis and object-level tracking when combined with general perception modules. 
Furthermore, VGGT~\cite{wang2025vggt} can output camera extrinsics for each frame within sequences, and SIFT~\cite{lowe2004SIFT} is employed for robust feature matching and homography matrix estimation, supporting geometric transformation tasks.

\vspace{2pt}
\noindent \textbf{Pose \& Geometry.}
We utilize VGGT~\cite{wang2025vggt} to estimate camera parameters~(intrinsics and extrinsics) from single-frame or multi-frame inputs. 
And Depth-Anything-V2~\cite{DepthAnythingv2} is adopted to provide metric depth using domain-specific models~(indoor/outdoor), which interfaces seamlessly with general perception modules to yield depth for detected objects or regions.
Moreover, OrientAnything~\cite{orient_anything} can estimate 3D object orientations, facilitating fine-grained viewpoint relationship reasoning, and MapAnything~\cite{MapAnything} is employed to reconstruct 3D scenes and predict real-world distances between points.

\vspace{2pt}
\noindent \textbf{Auxiliary Tools.}
We also implement general-purpose utilities to support tool interaction and orchestration, where a dedicated {\em Terminate} action consolidates tool outputs and signals the completion of reasoning. 
Additionally, we employ targeted prompt engineering to enhance the step-by-step reasoning capabilities of open-source MLLMs~({\em e.g.}, Qwen3-VL~\cite{Qwen3-VL}) when serving as the agent core. 

%% file: sections/4_experiments.tex
\section{Experiments}
\label{sec:Experiments}
In this section, we elaborate on the experimental settings in Sec.~\ref{subsec:experimental_settings}, including evaluation metrics and implementation details, followed by comprehensive quantitative and qualitative assessments on our SpatialScore in Sec.~\ref{subsec:quantitative_results} and Sec.~\ref{subsec:qualitative_results}, respectively.

\subsection{Experimental Settings}
\label{subsec:experimental_settings}
\noindent \textbf{Evaluation Metrics.}
We employ accuracy as our primary measure, with tailored scoring protocols. 
For judgment and multi-choice questions~({\em e.g.}, relative depth), we directly compare model responses against ground truth. For open-ended questions involving numerical values~({\em e.g.}, metric-based size estimation), we adopt the Mean Relative Accuracy~(MRA) proposed in~\cite{VSI-Bench}.
Notably, for complex open-ended QAs, we compute final scores by averaging accuracy from two methods:
(i) employ carefully designed parsing functions to extract answers, 
and (ii) utilize an off-the-shelf LLM~(GPT-OSS-20B~\cite{agarwal2025gpt-oss}) to score the response~(with the prompt detailed in Sec.~\ref{subsec:spatialscore_evaluation} of the \textbf{Appendix}).

\vspace{2pt}
\noindent \textbf{Implementation Details.}
All experiments are conducted on 8$\times$ Nvidia A100 GPUs with {\em float16} precision, except for Qwen2.5-VL~\cite{Qwen2.5-VL} and Qwen3-VL~\cite{Qwen3-VL} series, which use {\em bfloat16} to prevent numerical overflow.
FlashAttention-2~\cite{dao2023flashattention2} is enabled for all supported models to optimize efficiency.
To ensure reproducibility, we fix the random seed and set the temperature to $0$ throughout all evaluations.
For video inputs, we follow~\cite{VSI-Bench} and uniformly sample 32 frames for open-source models. 
For proprietary models, we provide 16 frames to GPT‑5~\cite{openai_gpt5_systemcard} and Gemini~\cite{comanici2025gemini, Gemini3}, while Claude‑4.5‑Sonnet~\cite{claude45} supports only 8 input frames.
More details regarding the prompt templates and supervised fine-tuning~(SFT) configurations will be provided in Sec.~\ref{subsec:spatialscore_evaluation} and Sec.~\ref{subsec:supervised_fine-tuning_with_spatialcorpus} of the \textbf{Appendix}.

\input{tables/quantitative_results_sft_agent}

\subsection{Quantitative Results}
\label{subsec:quantitative_results}
As presented in Tab.~\ref{tab:quantitative_results_sft_agent}, we investigate two exploratory strategies for enhancing spatial understanding and summarize the following findings:
(i) Compared to direct evaluation using pre-trained models~(Zero-shot), incorporating randomly selected, pre-allocated examples as context for each category~({\em i.e.}, One/Two/Four-shot evaluation) yields slight performance improvements~({\em e.g.}, One-shot evaluation increases overall accuracy by 1.51 and 0.78 points on the 4B and 8B models, respectively);
(ii) to further enhance the spatial understanding capabilities of existing models, we first fine-tune Qwen3-VL~\cite{Qwen3-VL} series models on SpatialCorpus, constructed from 2D simulators and 3D annotations.
Leveraging the high quality and broad coverage, both 4B and 8B fine-tuned models achieve notable improvements of \textbf{+10.47} and \textbf{+9.23} in overall accuracy, respectively;
(iii) however, we observe that the gains from supervised fine-tuning largely concentrate on already optimized task types, such as {\em mental animation} and {\em camera analysis}, while tasks with limited data scalability~({\em e.g.}, {\em view reasoning}) may suffer from catastrophic forgetting. 
This suggests that, although effective, purely data-driven approaches remain inherently limited in achieving holistic improvements;
and 
(iv) these observations motivate the development of \textbf{SpatialAgent}, which enhances spatial intelligence capabilities by dynamically invoking spatial perception tools and reasoning through {\em Plan–Execute~(PE)} or {\em ReAct} paradigms. 
Notably, while SpatialAgent yields slightly smaller absolute improvements than supervised fine-tuning~(specifically, \textbf{+6.41} and \textbf{+7.27} in the {\em Plan-Execute} paradigm, and \textbf{+7.78} and \textbf{+8.33} in the {\em ReAct} paradigm, respectively). 
It requires no additional training and consistently enhances performance across nearly all tasks, demonstrating its practicality and robustness.
More analysis will be presented in Sec.~\ref{subsec:additional_quantitative_results} of the \textbf{Appendix}.

\subsection{Qualitative Results}
\label{subsec:qualitative_results}
\begin{figure*}[t]
  \centering
  \includegraphics[width=\textwidth]{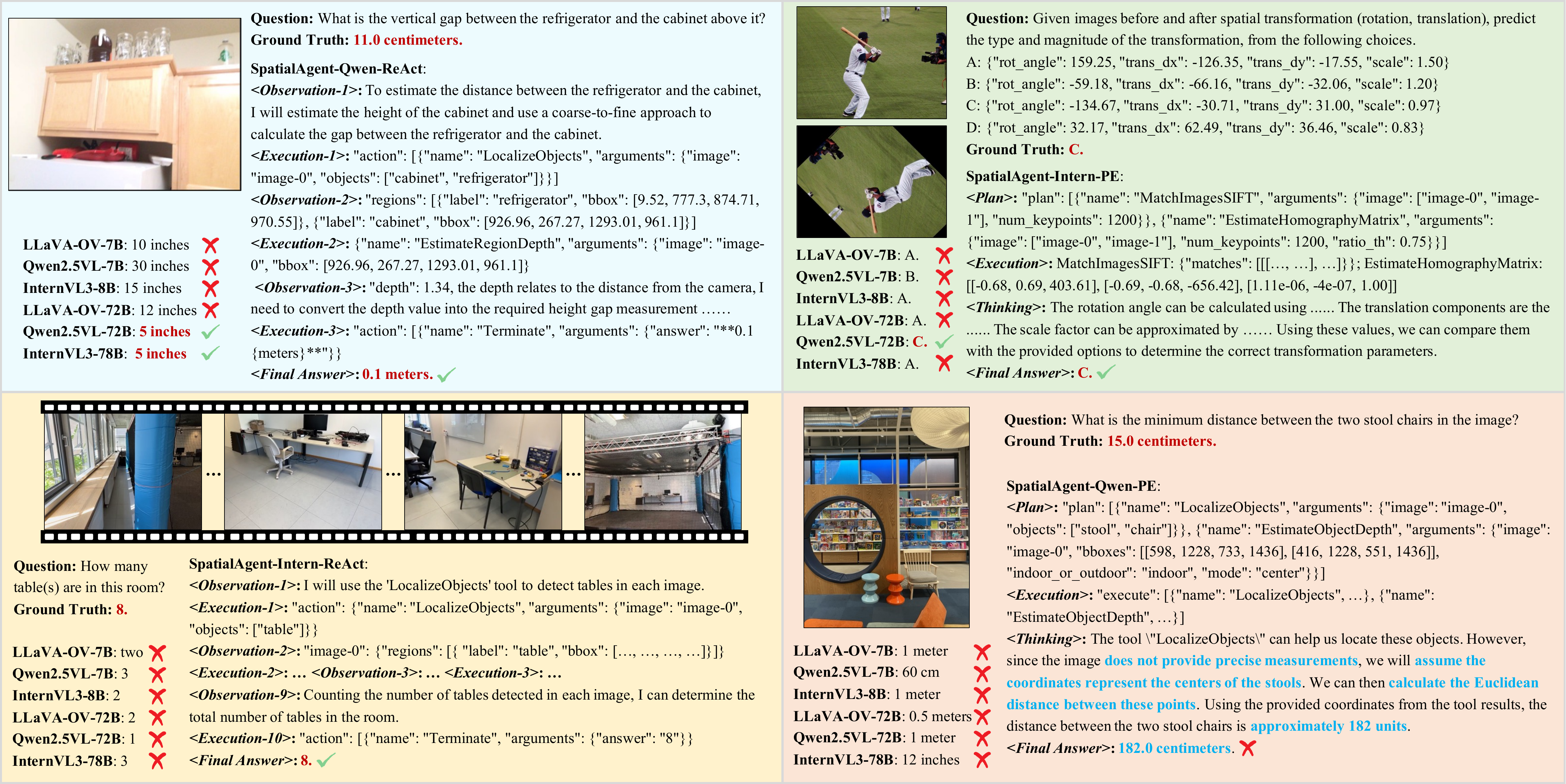} 
  \vspace{-0.5cm}
  \caption{
    \textbf{Qualitative Results.}
    We present the reasoning process of SpatialAgent against the direct responses of other models.
    While occasional errors occur due to tool execution or interpretation mistakes, these limitations are expected to diminish as MLLMs advance.
  }
 \label{fig:qualitative_results}
 \vspace{-6pt}
\end{figure*}

Fig.~\ref{fig:qualitative_results} presents representative case studies comparing SpatialAgent's reasoning trajectories, under both the {\em Plan-Execute} and {\em ReAct} paradigms, against various baselines. 
These results highlight SpatialAgent's structured and interpretable reasoning process, which systematically decomposes complex tasks and dynamically invokes appropriate tools for more accurate solutions.
While SpatialAgent exhibits strong performance across diverse spatial reasoning tasks, occasional failures still occur, typically due to suboptimal tool execution or misinterpretation of intermediate results~({\em e.g.}, confusing depth with object distance). 
Such limitations are expected to diminish as MLLMs continuously improve and as the toolbox design becomes more robust.
More qualitative comparisons will be provided in Sec.~\ref{subsec:additional_qualitative_results} of the \textbf{Appendix}.

%% file: tables/quantitative_results_sft_agent.tex
\begin{table*}[t]
    \caption{
        \textbf{Comparisons of our Data-driven and Agent-based Approaches on SpatialScore.}
        Qwen3-VL is adopted in two ways: (i) supervised fine-tuned on our SpatialCorpus; and (ii) as the agent core to conduct reasoning using the {\em Plan-Execute~(PE)} and {\em ReAct} paradigms in SpatialAgent.
    }
    \label{tab:quantitative_results_sft_agent}
    % \vspace{-6pt}
    \centering
    \setlength{\tabcolsep}{0.1cm}
    \renewcommand{\arraystretch}{1.0}
    \resizebox{\textwidth}{!}{
    \begin{tabular}{r|c|*{10}{c}}
    \toprule[1.5pt]
        \textbf{Methods} & \textbf{Overall} & \textbf{Mental.} & \textbf{Count.} & \textbf{Depth.} & \textbf{View-Rea.} & \textbf{Obj-Size.} &  \textbf{Obj-Loc.} & \textbf{Obj-Dist.} & \textbf{Obj-Mo.} & \textbf{Camera.} & \textbf{Temp-Rea.}   \\
        \midrule[0.8pt]
        \multicolumn{12}c{\textbf{Qwen3-VL-4B}} \\
        \midrule
        Blind~(text only) & 28.10 & 27.29 & 11.20 & 27.48 & 25.34 & 42.99 & 32.71 & 25.81 & 21.69 & 30.46 & 20.22 \\
        Zero-shot & 42.52 & 37.81 & 48.22 & 37.68 & 34.75 & 48.20 & 59.40 & 33.40 & 53.01 & 34.06 & 38.24 \\
        One-shot & 44.03 & 38.26 & 48.76 & 41.87 & 34.75 & 52.31 & 61.41 & 36.09 & 52.53 & 35.86 & 33.09 \\
        Two-shot & 45.66 & 40.27 & 51.32 & 43.25 & 34.75 & 51.88 & \underline{62.84} & 37.14 & 57.59 & 37.79 & 36.03 \\
        Four-shot & 46.59 & 43.85 & 51.78 & 45.16 & 35.65 & \underline{53.70} & \textbf{64.56} & 38.19 & 56.39 & 35.99 & 41.91 \\
        \rowcolor{cyan!10}
        w/ SpatialCorpus~(Ours) & \textbf{52.99} & \textbf{65.55} & 52.24 & \textbf{58.49} & 33.86 & 43.92 & 58.97 & \textbf{47.17} & \textbf{64.10} & \underline{55.78} & \textbf{44.85} \\
        \rowcolor{cyan!20}
        w/ SpatialAgent-PE~(Ours) & 48.93 & \underline{56.15} & \textbf{54.98} & 45.83 & \underline{36.55} & \textbf{54.70} & 62.41 & 36.68 & \underline{60.48} & 41.90 & 38.24 \\
        \rowcolor{cyan!20}
        w/ SpatialAgent-ReAct~(Ours) & \underline{50.30}
        & 46.53 & \underline{53.46} & \underline{51.75} & \textbf{39.01} & 50.51 & 58.11 & \underline{39.20} & 53.25 & \textbf{59.00} & \underline{42.28} \\
        \midrule
        \multicolumn{12}c{\textbf{Qwen3-VL-8B}} \\
        \midrule
        Blind~(text only) & 30.73 & 21.70 & 17.21 & 28.78 & 30.49 & 42.30 & 45.77 & 28.25 & 20.72 & 31.75 & 20.59 \\
        Zero-shot & 45.48 & 38.26 & 50.35 & 47.12 & 37.67 & 52.12 & 61.12 & 37.22 & 55.90 & 34.19 & 41.54 \\
        One-shot & 46.26 & 41.16 & 51.71 & 45.83 & 39.46 & 52.64 & 61.26 & 38.74 & 56.39 & 37.53 & 34.19  \\
        Two-shot & 47.61 & 41.39 & 54.01 & 48.97 & 40.81 & 53.81 & 60.11 & 38.54 & \underline{61.93} & 38.17 & 38.60 \\
        Four-shot & 49.00 & 47.87 & \underline{54.35} & 49.64 & 39.01 & 56.76 & \textbf{64.13} & 39.56 & 61.45 & 36.76 & 41.18  \\
        \rowcolor{cyan!10}
        w/ SpatialCorpus~(Ours) & \textbf{54.71} & \textbf{57.27} & 52.67 & \textbf{64.12} & 36.77 & \underline{56.81} & 60.26 & \textbf{49.50} & \textbf{64.34} & \underline{53.73} & \underline{44.85} \\
        \rowcolor{cyan!20}
        w/ SpatialAgent-PE~(Ours) & 52.75 & \underline{50.11} & \textbf{54.48} & \underline{54.44} & \underline{43.50} & \textbf{58.72} & \textbf{64.13} & 43.57 & 61.45 & 48.71 & 43.38 \\
        \rowcolor{cyan!20}
        w/ SpatialAgent-ReAct~(Ours) & \underline{53.81} & 44.30 & 53.49 & 54.21 & \textbf{45.74} & 56.52 & \underline{62.84} & \underline{43.67} & 60.96 & \textbf{58.10} & \textbf{51.84}
        \\
    \bottomrule[1.5pt]
    \end{tabular}
    \vspace{-4pt}
}
\end{table*}

%% file: sections/5_relatedwork.tex
\section{Related Work}
\label{sec:related_work}
\noindent \textbf{Multimodal Benchmarks.}
Recent advances in MLLMs~\cite{Qwen2.5-VL, Qwen3-VL, liu2023llava, li2025llavaonevision, lin2024vila, InternVL3}, exemplified by Gemini~\cite{team2023gemini, comanici2025gemini, Gemini3}, GPT-5~\cite{openai_gpt5_systemcard}, and Claude~\cite{claude35, claude45}, have spurred demand for comprehensive evaluation benchmarks.
Prior works like MMMU~\cite{yue2024mmmu}, Seed-Bench~\cite{li2024seed-bench}, and MMBench~\cite{liu2024mmbench}, have established broad assessments on visual-language understanding, while MMIU~\cite{MMIU} and BLINK~\cite{fu2024blink} focus on multi-image analysis, and VideoMME~\cite{fu2024video-mme} targets video comprehension.
Additional benchmarks specialize in domain-specific tasks, such as math~\cite{lu2024mathvista, wang2024Math-Vision, zhang2024mathverse, zou2025dynamath}, physics~\cite{chow2025physbench}, and sports~\cite{rao2025socceragent, xia2024sportu}.

\vspace{2pt}
\noindent \textbf{Spatial Intelligence.}
Prior work~\cite{OmniSpatial, MMSI-Bench, SPAR, Multi-SpatialMLLM, RoboSpatial, SpatialEval, yin2025spatial, cai2025SenseNova-SI, wang2025svqa, shen2025satori, batra2025spatialthinker} has explored diverse aspects of spatial understanding, including position relationship~\cite{xai2023realworldQA, kamath2023whatsup, VSR, MMVP, yang2019spatialsense}, size/orientation/distance estimation~\cite{cai2024spatialbot, xai2023realworldQA, ma20243dsrbench, tong2024cambrian}, and metric-based question answering~\cite{chen2024spatialvlm, daxberger2025mm, liao2024QSpatialBench}.
Recent efforts~\cite{brown2025SIMS-V, yang2025VisualSpatialTuning}, such as Open3DVQA~\cite{zhan2025open3dvqa} and Spatial457~\cite{wang2025spatial457}, leverage simulators for scalable and controllable data construction, enabling targeted fine-tuning of MLLMs, while SpatialRGPT~\cite{cheng2025spatialrgpt} introduces extra mask inputs for region-based analysis.
Several works~\cite{VSI-Bench, STI-Bench, VLM4D, yang2025cambrian-s, yan2026omnistream} further extend research focus to videos, facilitating dynamic scene understanding.
However, current data still suffer from limitations such as restricted task complexity, narrow evaluation scopes, and fragmented protocols.
To this end, we consolidate repurposed 3D data with samples from 23 existing datasets, establishing \textbf{SpatialScore}, the most comprehensive multi-modal spatial intelligence benchmark to date, which aims to promote progress in future spatial intelligence research.
 
\vspace{2pt}
\noindent \textbf{Multi-Agent Systems.}
As a powerful paradigm for tackling complex tasks, multi-agent systems~(MAS)~\cite{li2023camel, li2024optimus, ma2024taco} have found broad applications across software development~\cite{qian2024chatdev}, robotics~\cite{guo2024embodied, tan2020multi}, collaborative workflows~\cite{chen2024agentverse, yao2023react}, and scientific problem solving~\cite{ghafarollahi2024sciagents, li2024agenthospital}.
Recent frameworks like ReAct~\cite{yao2023react} and Reflexion~\cite{shinn2023reflexion} enable iterative reasoning through dynamic tool use, while platforms such as AutoGen~\cite{wu2023autogen} and MetaGPT~\cite{MetaGpt} facilitate multi-agent collaboration.
Given that spatial understanding inherently requires such reasoning abilities, we propose \textbf{SpatialAgent}, a multi-agent framework integrating 12 spatial perception tools, improving MLLMs in spatial intelligence in a training-free manner.

%% file: sections/6_conclusion.tex
\section{Conclusion}
\label{sec:conclusion}
This paper aims to systematically investigate the spatial intelligence of current MLLMs.
Concretely, we introduce \textbf{SpatialScore}, the most comprehensive and diverse spatial understanding benchmark to date, comprising around 5K carefully curated samples across 30 distinct tasks. 
Beyond evaluation, we propose two targeted solutions to bridge the observed performance gap:
(i) we construct \textbf{SpatialCorpus}, a large-scale training resource with 331K multimodal QA pairs for supervised fine-tuning;
and 
(ii) we develop \textbf{SpatialAgent}, a novel multi-agent framework with 12 specialized spatial perception tools, supporting both {\em Plan-Execute} and {\em ReAct} reasoning paradigms, which improves spatial reasoning abilities of existing MLLMs in a training-free manner.
Extensive evaluations on 49 representative MLLMs not only demonstrate the efficacy of our data-driven and agent-based approaches, but also reveal persistent challenges in spatial understanding.
We envision that these contributions will provide a rigorous foundation for advancing spatial intelligence in the future.

\section{Limitations \& Future Works}
\label{sec:limitations_and_future_works}
\subsection{Limitations}
While SpatialScore offers a comprehensive evaluation framework for spatial intelligence, SpatialCorpus provides large-scale, high-quality training samples, and SpatialAgent demonstrates promising improvements with a multi-agent system, our work is not without its limitations.
Although SpatialScore covers evaluation across single images, multi-frame sequences, and videos, it primarily relies on RGB frames, lacking samples that take point clouds, depth maps, or surface normals as input. 
Likewise, despite the effectiveness of SpatialCorpus, its diversity is still limited and cannot fully cover the breadth of spatial understanding tasks, leading to biased performance gains in fine-tuned models. 
Moreover, while SpatialAgent effectively boosts spatial understanding capabilities of MLLMs in a training-free manner by leveraging specialized tools, its current toolbox remains relatively rudimentary, and fundamental advances in spatial perception abilities of MLLMs are still required. 
These gaps are left for future work.

\subsection{Future Works}
To tackle the potential limitations, we outline several promising directions for advancing spatial intelligence:
(i) beyond RGB images/videos from existing datasets, incorporating diverse in-the-wild data and direct 3D inputs~({\em e.g.}, point clouds and depth maps) will further enhance the evaluation of spatial understanding capabilities and drive progress in related research areas;
(ii) expanding training samples to cover a broader range of task categories~({\em e.g.}, counting, orientation) may enable more stable and comprehensive improvements in spatial reasoning tasks through supervised fine-tuning;
(iii) enriching the toolbox with more robust expert models and facilitating better multi-agent collaboration will yield more reliable spatial reasoning systems;
and
(iv) substantial and holistic improvements in the spatial intelligence of multimodal large language models~(MLLMs) still require deeper, foundational advances, such as equipping models with essential intrinsic 3D representational understanding.

%% file: sections/7_acknowledgement.tex
\section*{Acknowledgments}
Weidi would like to acknowledge the funding from Scientific Research Innovation Capability Support Project for Young Faculty~(ZY-GXQNJSKYCXNLZCXM-I22).

%% file: sections/X_suppl.tex
\onecolumn
{
    \centering
    \Large
    \textbf{\fontsize{13.6}{10}\selectfont SpatialScore: Towards Comprehensive Evaluation for Spatial Intelligence} \\
    % \vspace{0.5em} Supplementary Material \\
    \vspace{0.5em} Appendix \\
    \vspace{1.0em}
}

\appendix
% {
%   \hypersetup{linkcolor=black}
%   \tableofcontents
% }

% \clearpage

This Appendix provides comprehensive implementation details, visualizations, experimental results, and in-depth analysis to support the main paper.
Specifically:
\begin{itemize}
    \item Sec.~\ref{sec:additional_data_details} provides further details on the construction, data statistics, and representative samples of our SpatialScore benchmark and SpatialCorpus dataset.
    \item Sec.~\ref{sec:more_implementation_details} elaborates on our implementation details, including parameter settings during evaluation and the construction details of SpatialAgent.
    \item Sec.~\ref{sec:more_experiment_results} supplements the paper with additional experimental results and analysis.
\end{itemize}

\section{Additional Data Details}
\label{sec:additional_data_details}
In this section, we provide additional details about our established \textbf{SpatialScore} benchmark and the \textbf{SpatialCorpus} training resources.
Specifically, in Sec.~\ref{subsec:annotation_quality_issues_in_existing_datasets}, we present several examples of inaccurate annotations found in existing datasets, highlighting the motivation and necessity for our thorough manual verification;
Next, we elaborate on how we repurpose existing 3D annotations into spatial-intelligence QA pairs in Sec~\ref{subsec:data_construction_details};
Then, Sec~\ref{subsec:additional_data_statistics} presents more detailed statistics and analyses for both SpatialScore and SpatialCorpus;
Finally, we include additional representative visualization examples in Sec~\ref{subsec:additional_representative_visualizations}.

\subsection{Annotation Quality Issues in Existing Datasets}
\label{subsec:annotation_quality_issues_in_existing_datasets}
\begin{figure}[!htb]
  \centering
  \includegraphics[width=\textwidth]{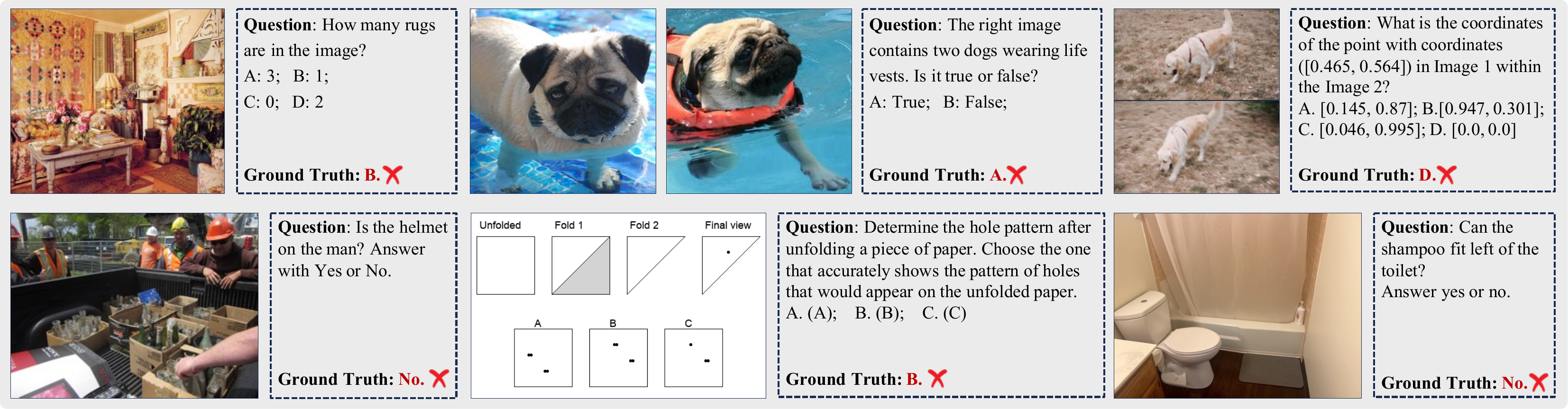} 
  % \vspace{-3pt}
  \caption{
  \textbf{Annotation Issues in Existing Benchmarks.}
  We observe that existing datasets contain various annotation errors or ambiguities, including those found in CV-Bench~\cite{tong2024cambrian}, SpatialSense~\cite{yang2019spatialsense}, MMIU~\cite{MMIU}, SRBench~\cite{ma20243dsrbench}, SITE-Bench~\cite{SITE}, and RoboSpatial-Home~\cite{RoboSpatial}.
  }
 \label{fig:bad_case}
 \vspace{-6pt}
\end{figure}

As depicted in Fig.~\ref {fig:bad_case}, we observe that existing spatial intelligence benchmarks contain numerous inaccurately annotated samples, which can undermine a fair evaluation of model performance.
This finding motivates two critical steps in the construction of SpatialScore: 
(i) creating new evaluation samples derived from high‑quality 3D annotations, 
and 
(ii) integrating existing benchmarks while manually verifying and filtering their samples to ensure high quality and sufficient difficulty. 
Consequently, we present \textbf{SpatialScore}, the most comprehensive and diverse spatial understanding benchmark to date.
It comprises \textbf{5,025} manually validated samples spanning \textbf{30} tasks across \textbf{10} categories, providing a comprehensive foundation for evaluating the spatial intelligence of current MLLMs.

\subsection{Data Construction Details}
\label{subsec:data_construction_details}
\noindent \textbf{Repurpose 3D Data for SpatialScore.}
Considering that ScanNet~\cite{dai2017scannet} has been widely used in prior work~\cite{fu2024blink, liao2024QSpatialBench, MMIU, SPAR, VSI-Bench} and may already be included in the training data of existing models, we intentionally avoid using such heavily reused datasets as metadata for constructing our data. 
Instead, we first sample 500 scenes from ScanNet++~\cite{yeshwanth2023scannet++}, Omni3D~\cite{brazil2023omni3d}, PointOdyssey~\cite{zheng2023pointodyssey}, WildRGB‑D~\cite{xia2024WildRGB-D}, and CA‑1M~\cite{lazarow2024cubify}, and convert their original annotated metadata into question–answer pairs. 
After extracting essential key information from each scene, we generate corresponding questions and contexts using predefined templates for each task, as presented below.

\vspace{3pt} 
\noindent \textbf{Task 1: Object Existence.}
Using the data from Omni3D~\cite{brazil2023omni3d}, we construct samples for the {\em object existence} task, {\em i.e.}, determining whether a specific object category appears in an image, based on the following question templates:
{
    \begin{tcolorbox}[breakable]
        IS\_THERE: ["Is there any \{{\em category}\} present in the image?"]

        DOES\_CONTAIN: ["Does the image contain any \{{\em category}\}?"]

        IS\_VISIBLE: ["Is any \{{\em category}\} visible in this image?"]

        CAN\_YOU\_SEE: ["Can you see any \{{\em category}\} in this image?"]
    \end{tcolorbox}

}

\vspace{3pt} 
\noindent \textbf{Task 2: 3D Object Detection.}
We adopt the 3D bounding‑box annotations from Omni3D~\cite{brazil2023omni3d} and CA‑1M~\cite{lazarow2024cubify}~(represented by the coordinates of eight 3D corner points) to construct samples for {\em 3D object detection} task, and generate distractors by adding small random noise to each coordinate of the ground‑truth box.

{
    \begin{tcolorbox}[breakable]
        DETECT\_3D\_BBOX: [
        
            \quad\quad"Detect the 3D bounding box of the \{{\em object\_name}\} in the image.",
            
            \quad\quad"Provide the 3D bounding box for the \{{\em object\_name}\}.",
            
            \quad\quad"What is the 3D bounding box of the \{{\em object\_name}\}?",
            
            \quad\quad"Locate the \{{\em object\_name}\} and output its 3D bounding box.",
            
            \quad\quad"Output the 3D bounding box coordinates for the \{{\em object\_name}\}."
        
        ]
        
        PROVIDE\_3D\_BBOX: [
        
            \quad\quad"Can you provide the 3D bounding box of the \{{\em object\_name}\}?",
            
            \quad\quad"Please detect and output the 3D bounding box for the \{{\em object\_name}\}.",
            
            \quad\quad"Identify the \{{\em object\_name}\} and provide its 3D bounding box coordinates.",
            
            \quad\quad"What are the 3D coordinates of the \{{\em object\_name}\}'s bounding box?"
        
        ]
        
        WHAT\_IS\_3D\_BBOX: [
        
            \quad\quad"What is the 3D bounding box of the \{{\em object\_name}\}?",
            
            \quad\quad"Where is the \{{\em object\_name}\} located in 3D space? Provide its bounding box.",
            
            \quad\quad"Determine the 3D bounding box coordinates for the \{{\em object\_name}\}."
        
        ]
    \end{tcolorbox}
}

\vspace{3pt} 
\noindent \textbf{Task 3-5: Absolute Depth \& Absolute Distance \& Absolute Size.}
We construct these metric‑based task samples~(typically using units such as meters, feet, or centimeters) from the 3D bounding‑box data in the Omni3D~\cite{brazil2023omni3d} and CA‑1M~\cite{lazarow2024cubify} datasets. 
For distractors, we first generate one value very close to the correct answer~(usually within 85\%–95\% or 105\%–115\% of the ground truth). 
We then create additional distractors within the broader ranges of 50\%–90\% and 110\%–180\% of the correct value. 
If the above strategy does not yield enough distractors, we generate additional ones by simply adding or subtracting fixed values~(e.g., 0.5 or 1.0) from the correct answer.

{
    \begin{tcolorbox}[breakable]
        \#\#\# Absolute Depth:

        HOW\_FAR: [
            
            \quad\quad "How far is the \{{\em object\_name}\} from the camera?",
            
            \quad\quad "What is the distance of the \{{\em object\_name}\} from the camera?",
            
            \quad\quad "How far away is the \{{\em object\_name}\}?"
        
        ],
        
        DISTANCE\_FROM\_CAMERA: [
            
            \quad\quad "What is the approximate distance from the camera to the \{{\em object\_name}\}?",
            
            \quad\quad "How many \{{\em unit}\} away is the \{{\em object\_name}\} from the camera?",
            
            \quad\quad "At what distance is the \{{\em object\_name}\} located from the camera?"
        
        ],
        
        APPROXIMATE\_DISTANCE: [
            
            \quad\quad "Approximately how far is the \{{\em object\_name}\}?",
            
            \quad\quad "What is the rough distance to the \{{\em object\_name}\}?",
            
            \quad\quad "How far would you estimate the \{{\em object\_name}\} to be?"
            
        ]\\

        \#\#\# Absolute Distance:
        
        [
    
            \quad\quad"What is the distance between the \{{\em object1}\} and the \{{\em object2}\}?",
            
            \quad\quad"How far apart are the \{{\em object1}\} and the \{{\em object2}\}?",
            
            \quad\quad"What is the approximate distance from the \{{\em object1}\} to the \{{\em object2}\}?",
            
            \quad\quad"How much distance separates the \{{\em object1}\} and the \{{\em object2}\}?",
            
            \quad\quad"What is the spatial separation between the \{{\em object1}\} and the \{{\em object2}\}?"

        ]\\
        
        \#\#\# Absolute Size:

        WHAT\_IS\_DIMENSION: [
        
            \quad\quad"What is the \{{\em dimension}\} of the \{{\em object\_name}\}?",
            
            \quad\quad"What is the \{{\em dimension}\} \{{\em dimension\_type}\} of the \{{\em object\_name}\}?",
            
            \quad\quad"How much is the \{{\em dimension}\} of the \{{\em object\_name}\}?"
    
        ],
    
        HOW\_DIMENSION: [
            
            \quad\quad"How \{{\em dimension}\} is the \{{\em object\_name}\}?",
            
            \quad\quad"How \{{\em dimension\_adj}\} is the \{{\em object\_name}\}?",
        
        ],
        
        DIMENSION\_OF\_OBJECT: [
            
            \quad\quad"What is the \{{\em object\_name}\}'s \{{\em dimension}\}?",
            
            \quad\quad"How would you measure the \{{\em dimension}\} of the \{{\em object\_name}\}?",
            
            \quad\quad"What dimension represents the \{{\em dimension}\} of the \{{\em object\_name}\}?"
        
        ]
    \end{tcolorbox}
}

\vspace{3pt} 
\noindent \textbf{Task 6-8: Relative Depth \& Relative Distance \& Relative Size.}
Samples for these tasks involving relative comparisons can likewise be constructed from the metadata of Omni3D~\cite{brazil2023omni3d} and CA‑1M~\cite{lazarow2024cubify}, and the distractors can be easily generated by simply selecting metadata corresponding to other objects or points.

{
    \begin{tcolorbox}[breakable]
        \#\#\# Relative Depth:
        
        [
        
            \quad\quad"Which object is closest to the camera?",
            
            \quad\quad"Among the following objects, which one is nearest to the camera?", 
            
            \quad\quad"Which of these objects has the shortest distance from the camera?",
            
            \quad\quad"Select the object that is closest to the camera:",
            
            \quad\quad"Which object appears closest in the image?"
            
        ]\\

        \#\#\# Relative Distance:

        [
        
            \quad\quad"Among the following objects, which one is closest to the \{{\em reference}\}?",
            
            \quad\quad"Which object has the shortest distance to the \{{\em reference}\}?",
            
            \quad\quad"Select the object that is nearest to the \{{\em reference}\}:",
            
            \quad\quad"Which of these objects is closest to the \{{\em reference}\}?",
            
            \quad\quad"What object is positioned closest to the \{{\em reference}\}?"

        ]\\
        
        \#\#\# Relative Size:

        (ComparisonDimension.HEIGHT, ComparisonType.LARGER): [
        
        \quad\quad"Which object is taller, the \{{\em object1}\} or the \{{\em object2}\}?",
        
        \quad\quad"Between the \{{\em object1}\} and the \{{\em object2}\}, which one is higher?",
        
        \quad\quad"Which is taller: the \{{\em object1}\} or the \{{\em object2}\}?",
        
        \quad\quad"Compare the height of the \{{\em object1}\} and the \{{\em object2}\}. Which one is taller?"
    
    ],
    
    (ComparisonDimension.WIDTH, ComparisonType.LARGER): [
    
        \quad\quad"Which object is wider, the \{{\em object1}\} or the \{{\em object2}\}?",
        
        \quad\quad"Between the \{{\em object1}\} and the \{{\em object2}\}, which one is wider?",
        
        \quad\quad"Which is wider: the \{{\em object1}\} or the \{{\em object2}\}?",
        
        \quad\quad"Compare the width of the \{{\em object1}\} and the \{{\em object2}\}. Which one is wider?"
    
    ],
    
    (ComparisonDimension.LENGTH, ComparisonType.LARGER): [
        
        \quad\quad"Which object is longer, the \{{\em object1}\} or the \{{\em object2}\}?",
        
        \quad\quad"Between the \{{\em object1}\} and the \{{\em object2}\}, which one is longer?",
        
        \quad\quad"Which is longer: the \{{\em object1}\} or the \{{\em object2}\}?",
        
        \quad\quad"Compare the length of the \{{\em object1}\} and the \{{\em object2}\}. Which one is longer?"
        
    ],
    
    (ComparisonDimension.HEIGHT, ComparisonType.SMALLER): [
        
        \quad\quad"Which object is shorter, the \{{\em object1}\} or the \{{\em object2}\}?",
        
        \quad\quad"Between the \{{\em object1}\} and the \{{\em object2}\}, which one is lower?",
        
        \quad\quad"Which is shorter: the \{{\em object1}\} or the \{{\em object2}\}?",
        
        \quad\quad"Compare the height of the \{{\em object1}\} and the \{{\em object2}\}. Which one is shorter?"
        
    ],
    
    (ComparisonDimension.WIDTH, ComparisonType.SMALLER): [
        
        \quad\quad"Which object is narrower, the \{{\em object1}\} or the \{{\em object2}\}?",
        
        \quad\quad"Between the \{{\em object1}\} and the \{{\em object2}\}, which one is narrower?",
        
        \quad\quad"Which is narrower: the \{{\em object1}\} or the \{{\em object2}\}?",
        
        \quad\quad"Compare the width of the \{{\em object1}\} and the \{{\em object2}\}. Which one is narrower?"
    
    ],
    
    (ComparisonDimension.LENGTH, ComparisonType.SMALLER): [
        
        \quad\quad"Which object is shorter in length, the \{{\em object1}\} or the \{{\em object2}\}?",
        
        \quad\quad"Between the \{{\em object1}\} and the \{{\em object2}\}, which one is shorter?",
        
        \quad\quad"Which is shorter: the \{{\em object1}\} or the \{{\em object2}\}?",

        \quad\quad"Compare the length of the \{{\em object1}\} and the \{{\em object2}\}. Which one is shorter?"
        
    ]
    \end{tcolorbox}
}

\vspace{3pt} 
\noindent \textbf{Task 9: Camera Intrinsics.}
Our considered available 3D annotations all provide {\em camera intrinsics} information. 
For distractors, we generate random values within predefined proportional ranges: for focal lengths~(fx and fy), principal‑point coordinates~(cx and cy), and the skew parameter~(s), we randomly sample distractors using variance ratios of 0.25, 0.20, and 0.10, respectively.

{
    \begin{tcolorbox}[breakable]
        FOCAL\_LENGTH: [
            
            \quad\quad"What is the camera's focal length in pixels?",
            
            \quad\quad"What is the focal length of the camera?",
            
            \quad\quad"Can you determine the camera's focal length?"
        
        ]
        
        PRINCIPAL\_POINT: [
            
            \quad\quad"What are the image center coordinates in the camera intrinsics?",
            
            \quad\quad"What is the principal point of the camera?",
            
            \quad\quad"What are the coordinates of the image center?"
        
        ]
        
        FOCAL\_LENGTH\_X: [
            
            \quad\quad"What is the horizontal focal length (fx) in pixels?",
            
            \quad\quad"What is the camera's focal length in the x-direction?",
            
            \quad\quad"Can you determine the horizontal focal length?"
        
        ]
        
        FOCAL\_LENGTH\_Y: [
            
            \quad\quad"What is the vertical focal length (fy) in pixels?",
            
            \quad\quad"What is the camera's focal length in the y-direction?",
            
            \quad\quad"Can you determine the vertical focal length?"
        ]
        
        ASPECT\_RATIO: [
            
            \quad\quad"What is the aspect ratio of the camera's focal lengths (fx/fy)?",
            
            \quad\quad"What is the ratio between horizontal and vertical focal lengths?",
            
            \quad\quad"Can you calculate the aspect ratio from the camera intrinsics?"
        
        ]
    \end{tcolorbox}

}

\vspace{3pt} 
\noindent \textbf{Task 10: Camera Extrinsics.}
Constructing samples for {\em camera extrinsics} estimation requires the annotations from CA‑1M~\cite{lazarow2024cubify}, ScanNet++~\cite{yeshwanth2023scannet++}, and WildRGB‑D~\cite{xia2024WildRGB-D}. 
For distractors, we randomly choose from three strategies:
(i) Axis swapping or sign flipping: randomly swap axes of the rotation matrix or flip the sign of an axis while keeping the matrix orthogonal;
(ii) Translation vector perturbation: keep the rotation unchanged and add random noise to the translation vector;
and
(iii) Small rotational perturbation: apply a slight rotational disturbance to the rotation matrix to produce a new, approximately orthogonal matrix.

{
    \begin{tcolorbox}[breakable]
        [
        
        \quad\quad"What is the transformation matrix from the first camera coordinate system to the second camera coordinate system in OpenCV convention?",
    
        \quad\quad"Can you provide the relative transformation matrix between the two camera poses in OpenCV convention?",
        
        \quad\quad"What is the 4x4 transformation matrix that transforms coordinates from the first camera frame to the second camera frame in OpenCV convention?",
    
        \quad\quad"Please calculate the extrinsic transformation matrix from camera 1 to camera 2 in OpenCV convention.",

]
    \end{tcolorbox}
}

\vspace{3pt} 
\noindent \textbf{Task 11: Camera Motion.}
The {\em camera motion} task is derived from {camera extrinsics}.
We first identify the axes with the most significant rotational or translational changes. If the dominant motion exceeds a high threshold~({\em e.g.}, rotation greater than 10°), it is considered a salient motion to be described.
If the motion is below a low threshold~({\em e.g.}, rotation less than 5°), the camera is treated as stationary along that axis. 
Motions falling in between are ignored in the correct answer but may be used when generating distractors. 
Each degree of freedom~(roll, pitch, yaw, x, y, z) is labeled with a state~(changed, ignored, stationary) and a direction (e.g., left, up, forward).
A standardized, human‑readable description is then produced as the correct answer. 
For example, if roll changes significantly to the left and translation moves significantly backward, the output becomes: {\em ``The camera rolled left and moved backward.''}

Distractors are generated using the following strategies:
(i) Opposite motion: For true motions~(state = changed), there is a 70\% chance of describing them with the opposite direction.
(ii) Omission: For true motions, there is a 30\% chance of omitting them entirely and treating the camera as stationary.
and
(iii) Fabrication: For ignored minor motions~(state = ignored), there is a 30\% chance of fabricating a new motion.
These incorrect motion fragments~({\em e.g.}, ``moved backward,'' ``pitched up'') are combined into complete sentences. 
If all true motions are omitted, a fallback option {\em ``The camera remained stationary.''} is generated.
Finally, distractors that duplicate each other or the correct answer are removed. 
If the remaining number is insufficient, additional distractors are sampled from a preset list of generic motions ({\em e.g.}, ``The camera moved forward.'') to ensure adequate coverage.

{
    \begin{tcolorbox}[breakable]
        multi\_choice: [
    
        \quad\quad"Which best describes the camera motion between these two images?",
    
        \quad\quad"How did the camera primarily move?",
        
        \quad\quad"What type of camera movement occurred?",
    
        \quad\quad"Which motion pattern best matches the camera transformation?",

]

        open\_ended: [
        
            \quad\quad"What kind of camera motion occurred between the two images?",
            
            \quad\quad"Describe the relative motion of the camera from the first image to the second image.",
            
            \quad\quad"How did the camera move between these two frames?",
            
            \quad\quad"Can you describe the camera movement between the two views?",
            
            \quad\quad"What is the camera's motion from the first view to the second view?",
        
        ]
    \end{tcolorbox}
}

\vspace{3pt} 
\noindent \textbf{Task 12: Point Tracking.}
We construct {\em point tracking} samples by leveraging the metadata from CA‑1M~\cite{lazarow2024cubify}, PointOdyssey~\cite{zheng2023pointodyssey}, and WildRGB‑D~\cite{xia2024WildRGB-D} to establish correspondences across multiple images. 
For distractors, we randomly select other non‑corresponding points from the images.

{
    \begin{tcolorbox}[breakable]
        [
    
        \quad\quad"In the first image, there is a point at coordinates (\{x1\}, \{y1\}). Which point in the second image corresponds to this tracked point?",
    
        \quad\quad"Given a point at position (\{x1\}, \{y1\}) in image 1, which of the following coordinates in image 2 represents the same tracked point?",
    
        \quad\quad"A point is tracked from image 1 at (\{x1\}, \{y1\}). Where does this point appear in image 2?",
    
        \quad\quad"Tracking point from image 1: (\{x1\}, \{y1\}). Select its corresponding location in image 2:"
        
]
    \end{tcolorbox}
}

\vspace{3pt} 
\noindent \textbf{Task 13: Homography Matrix.}
Data for the {\em homography matrix} task can be easily constructed from all available metadata, and multiple additional homography matrices can be randomly generated as distractors. 
The corresponding question templates are presented as follows.

{
    \begin{tcolorbox}[breakable]
        [
        
        \quad\quad"What is the homography matrix that transforms the original image to the given transformed image?",
        
        \quad\quad"Please provide the 3x3 homography transformation matrix between the original and transformed images.",
    
        \quad\quad"Calculate the homography matrix that maps the original image to the transformed version.",
    
        \quad\quad"What is the perspective transformation matrix from the original image to the transformed image?"
        
]
    \end{tcolorbox}
}

\vspace{3pt}
Furthermore, to boost the linguistic diversity of QA pairs while preserving semantic integrity, we employ an off-the-shelf LLM~({\em i.e.}, DeepSeek-V3~\cite{liu2024deepseekv3}) to systematically rephrase questions, generate distractors, and convert question types, using the following prompts:

{

    \begin{tcolorbox}[breakable]
    \#\#\# For rephrasing open-ended questions:

    \quad Please rephrase the following question while maintaining its original meaning. Requirements:
    
    \quad 1. Keep the core meaning of the question unchanged
    
    \quad 2. Use natural and fluent language
    
    \quad 3. Return only the rephrased question, nothing else
    
    \quad Original question: \{{\em question}\}
    
    \quad Rephrased question:\\

    \#\#\# For rephrasing multi-choice questions:
    
    \quad Please rephrase the following multiple-choice question while maintaining its original meaning. Requirements:
    
    \quad 1. Keep the core meaning of the question unchanged
    
    \quad 2. If there is an instruction phrase like "Select from the following choices", keep it
    
    \quad 3. Use natural and fluent language
    
    \quad 4. Return only the rephrased question, nothing else
    
    \quad Original question: \{{\em question}\}
    
    \quad Rephrased question:\\

    \#\#\# For generating distractor options:

    \quad Based on the following question and correct answer, generate {num\_options} options (including the correct answer). Requirements:
    
    \quad 1. Options should be reasonable and have distraction value
    
    \quad 2. The correct answer is: \{{\em correct\_answer}\}
    
    \quad 3. Other options should be incorrect but plausible answers
    
    \quad 4. Return in JSON format: \{"options": ["option1", "option2", ...]\}
    
    \quad 5. Return only JSON, nothing else
    
    \quad Question: \{{\em question}\}
    
    \quad Correct answer: \{{\em correct\_answer}\}
    
    \quad Required answer: \{{\em required\_ans}\}
    
    \quad Generated options:\\

    \#\#\# For converting questions into judgment questions:
    
    \quad Convert the following question to a yes/no question format. Requirements:
    
    \quad 1. Keep the core meaning unchanged.
    
    \quad 2. The question should be answerable with yes or no.
    
    \quad 3. The converted question should be as specific as possible, directly incorporating relevant details and data points (e.g., specific values, coordinates, identifiers) from the original question or answer. 
    Avoid asking general questions about detection, presence, or existence if more specific information can be queried.
    
    \quad 4. Based on the original answer "\{{\em correct\_answer}\}", determine if the yes/no answer should be "yes" or "no".
    
    \quad 5. Return in JSON format: \{"question": "yes/no question", "answer": "yes or no"\}.
    
    \quad 6. Return only JSON, nothing else.
    
    \quad Original question: \{{\em question}\}
    
    \quad Original answer: \{{\em correct\_answer}\}
    
    \quad Required answer: \{{\em required\_ans}\}
    
    \quad Converted question:
    \end{tcolorbox}
}

\vspace{4pt}

\vspace{2pt}
\noindent \textbf{Scaling Data for SpatialCorpus.}
Building on the automated pipeline that repurposes 3D annotations into spatial understanding question–answer pairs, we employ the metadata of the training sets of ScanNet++~\cite{yeshwanth2023scannet++}, Omni3D~\cite{brazil2023omni3d}, WildRGB‑D~\cite{xia2024WildRGB-D}, and PointOdyssey~\cite{zheng2023pointodyssey} to create additional training samples. 
Note that CA‑1M~\cite{lazarow2024cubify} is excluded at this stage since it only contains class‑agnostic object annotations. 
We enhance data diversity by designing richer question templates, thereby avoiding the heavy cost of large‑scale LLM‑based rephrasing.

Additionally, to further improve model performance on tasks related to {\em mental animation}, we utilize simulators to generate scalable data for {\em spatial map}, {\em multi-view projection}, and {\em 2D/3D rotations}.
For these {\em mental animation} tasks, distractors are constructed as follows:
(i) Spatial Map: We randomly place several non‑overlapping locations on a map and compute directional relations based on their coordinates. 
From this, we generate four types of multi‑choice questions: determining directional relations, finding an object in a specified direction, counting objects in a given direction, and identifying the nearest object.
Each question contains exactly one correct answer and spans diverse spatial‑relation types.
(ii) 2D Rotation: We generate a colored grid reference image with no rotational or mirror symmetry, create a single correct option by rotating it~(90°/180°/270°), and add three distractors~({\em e.g.}, flipped or color‑shifted versions). 
All distractors are guaranteed not to match the reference under any rotation, ensuring a strictly single‑answer setting.
and
(iii) 3D Rotation: We construct a colored voxel‑based 3D shape without self‑symmetry. 
The correct option is obtained via one of the 24 valid rotation matrices, while distractors include shapes with different voxel counts, mismatched colors, or altered spatial layouts with the same voxel count. 
Only the correct option remains equivalent to the reference under valid 3D rotations.

The corresponding question templates are provided below:

{
    \begin{tcolorbox}[breakable]
        \#\#\# Spatial Map:

        \quad 1. direction\_relation: "question": "In which direction is \{{\em q1\_p1}\} relative to \{{\em q1\_p2}\}? \{{\em DIRECTION\_RULE}\}"
        
        \quad 2. find\_object: "question": "Which object is in the \{{\em target\_dir}\} of \{{\em q2\_p1}\}? \{{\em DIRECTION\_RULE}\}"
        
        \quad 3. count\_objects: "question": "How many objects are in the \{{\em q3\_target\_dir}\} of \{{\em q3\_p1}\}? \{{\em DIRECTION\_RULE}\}"
        
        \quad 4. closest\_object: "question": "Which object is closest to \{{\em q4\_p1}\}?"
        \\
        
        \#\#\# Multi-view Projection:
        
        \quad 1. view\_identification: "The first image shows a 3D view of the scene, while the second shows one of the three orthographic views of this 3D scene. 
        What type of view is displayed in the second image?
        The front view is observed from the positive direction of the Y-axis toward the negative direction, the left view is observed from the positive direction of the X-axis toward the negative direction, and the top view is observed from the positive direction of the Z-axis toward the negative direction."

        \quad 2. view\_matching: "Which option shows the \{{\em target\_view}\} view of the 3D scene?"
                The front view is observed from the positive direction of the Y-axis toward the negative direction, the left view is observed from the positive direction of the X-axis toward the negative direction, and the top view is observed from the positive direction of the Z-axis toward the negative direction."
        \\
        
        \#\#\# 2D\_Rotation: 
        
        \quad "Which option is the rotated version of the reference shape?"
        \\
        
        \#\#\# 3D\_Rotation:
        
        \quad "Which option is the rotated version of the reference 3D shape?"
    \end{tcolorbox}
}

\vspace{4pt}

\subsection{Data Statistics}
\label{subsec:additional_data_statistics}
\begin{figure}[ht!]
  \centering
  \includegraphics[width=\textwidth]{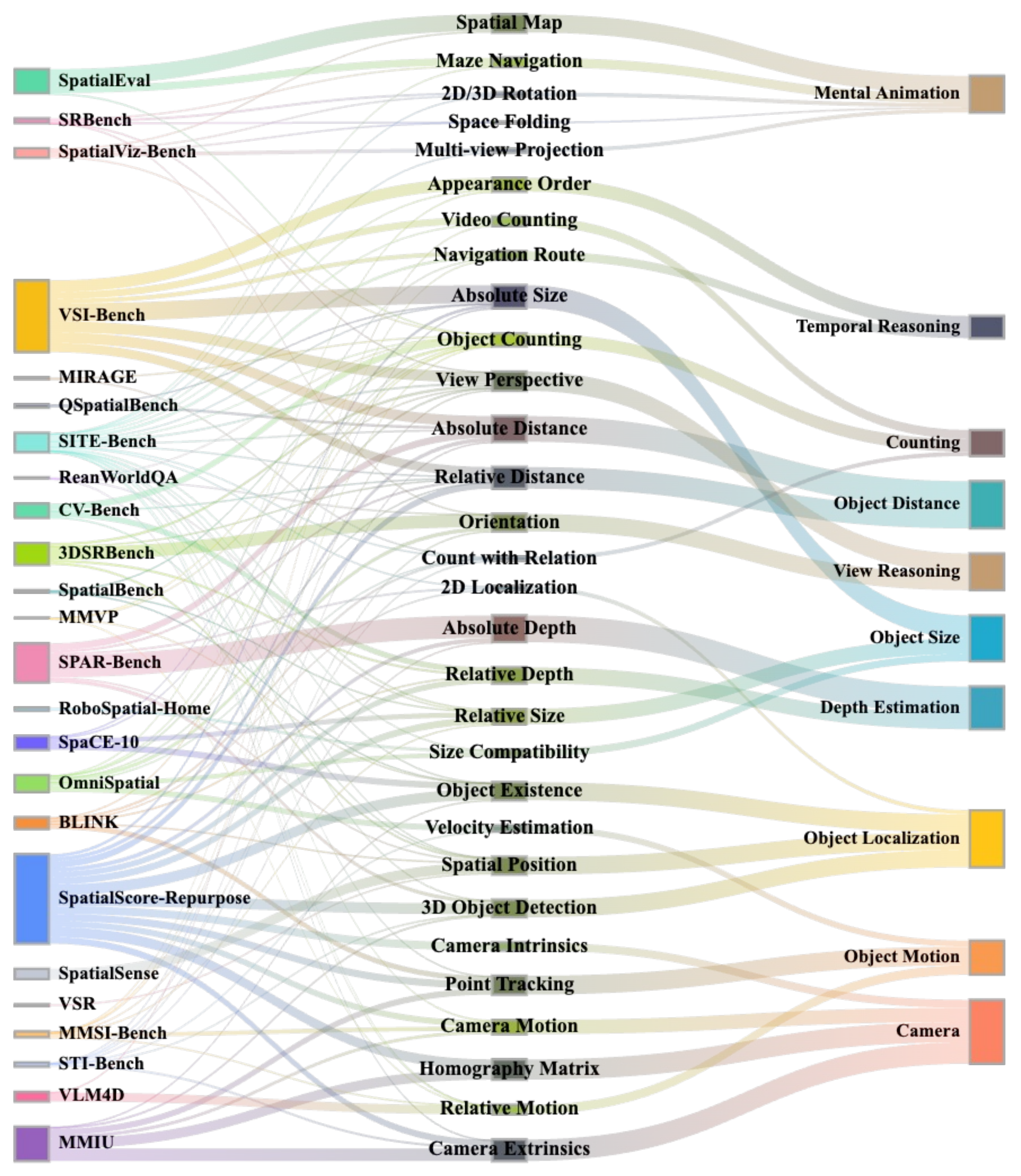} 
  \caption{
  \textbf{Data Sources and Task Category Statistics Visualization of SpatialScore.}
  Here, we denote the newly constructed samples repurposed from 3D annotations as {\em SpatialScore‑Repurpose}.
  }
 \label{fig:spatialscore_distribution}
 % \vspace{-12pt}
\end{figure}

We provide the details on the distributions of question formats, input modalities, data sources, and the samples across categories and tasks for \textbf{SpatialCorpus} and \textbf{SpatialScore} in Tab.~\ref{tab:statistics_spatialcorpus} and Tab.~\ref{tab:statistics_spatialscore}, respectively.
Then we further visualize the distributions of data sources, as well as the distributions across categories and tasks within SpatialScore, in Fig.~\ref{fig:spatialscore_distribution}.
Here, we denote the newly constructed samples repurposed from 3D annotations as {\em SpatialScore‑Repurpose}.
 
By integrating these data and conducting manual verification, SpatialScore encompasses a wide range of spatial intelligence tasks with diverse question–answering formats~(judgment, multiple-choice, and open-ended) and input modalities~(single image, multi-frame sequence, and video), establishing the most comprehensive and heterogeneous benchmark for spatial understanding to date.
This makes it an effective testbed for evaluating the spatial reasoning capabilities of current MLLMs. 
We believe this advancement will further drive research progress in the field of spatial intelligence.

\input{tables/spatialcorpus_statistics}

\input{tables/spatialscore_statistics}

\subsection{Additional Representative Visualizations}
\label{subsec:additional_representative_visualizations}
\begin{figure}[t]
  \centering
  \includegraphics[width=\textwidth]{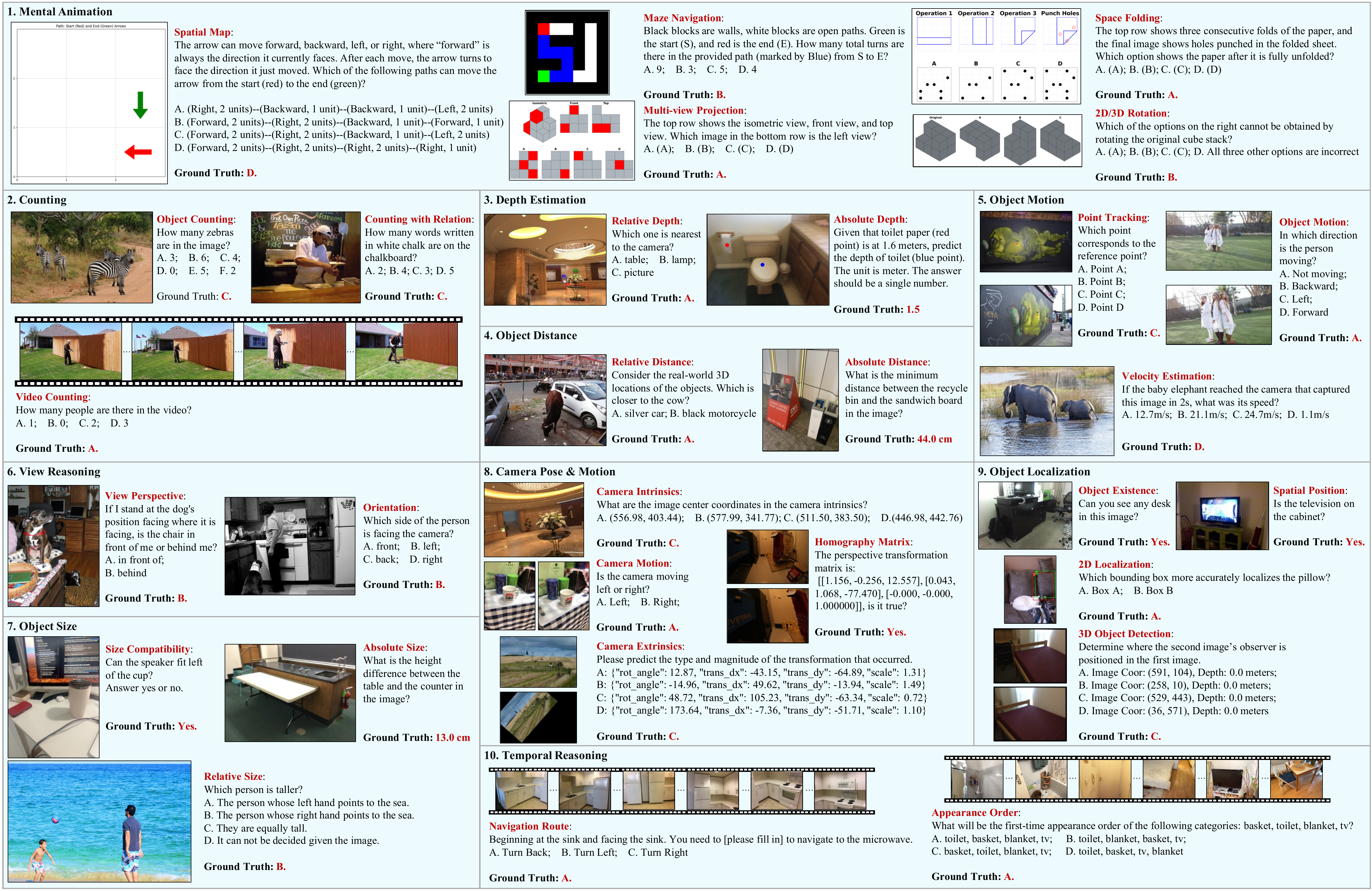} 
  \caption{
    \textbf{More Representative Examples in SpatialScore.} 
    Here, some questions have been slightly rewritten for clarity of presentation.
  }
 \label{fig:visualization_spatialscore}
 \vspace{-6pt}
\end{figure}

\begin{figure}[t]
  \centering
  \includegraphics[width=\textwidth]{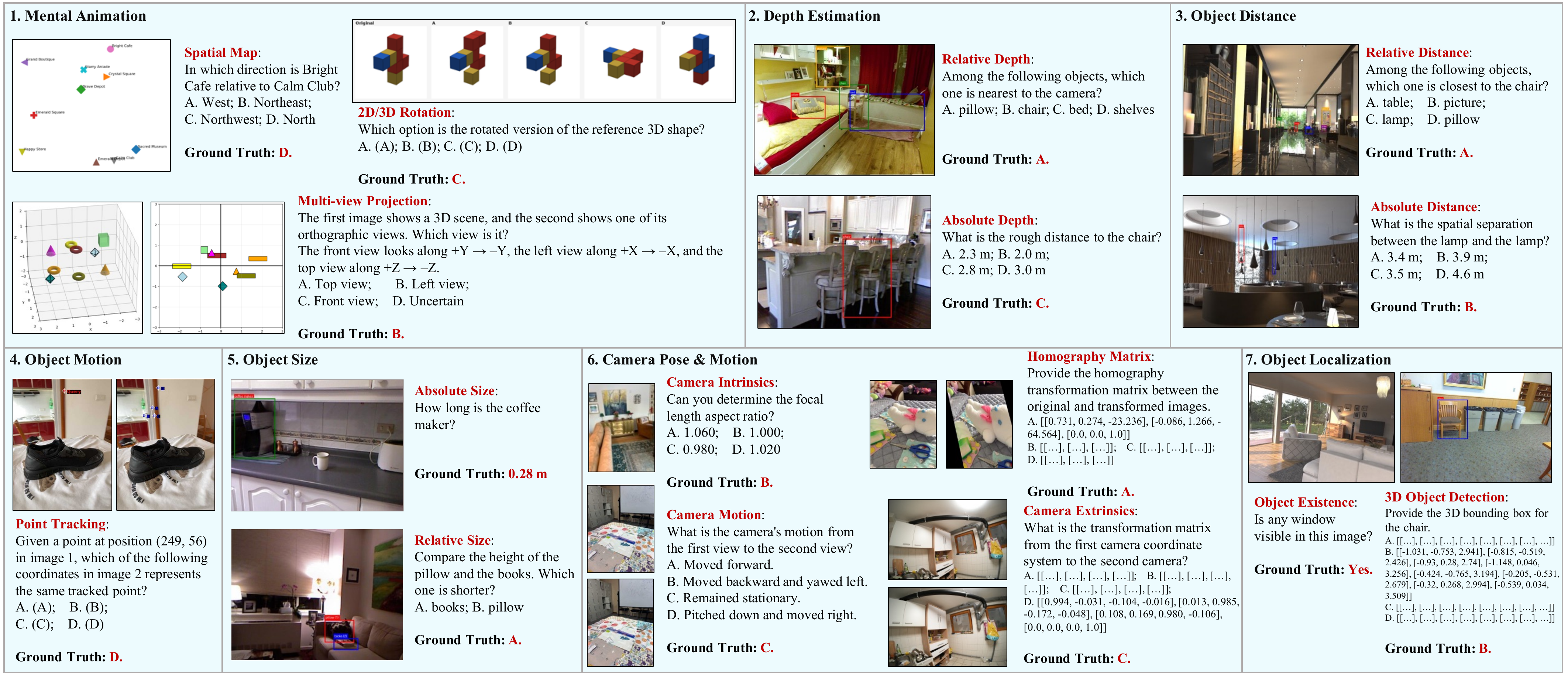} 
  \caption{
    \textbf{More Representative Examples in SpatialCorpus.}
    Here, some questions have been slightly rewritten for clarity of presentation.
  }
 \label{fig:visualization_spatialcorpus}
 \vspace{-6pt}
\end{figure}

We further present representative examples from each task in SpatialScore and SpatialCorpus to demonstrate their comprehensiveness and diversity. 
As depicted in Fig.~\ref{fig:visualization_spatialscore}, SpatialScore contains \textbf{5,025} samples covering \textbf{30} tasks across \textbf{10} categories, while Fig.~\ref{fig:visualization_spatialcorpus} illustrates that SpatialCorpus includes over \textbf{331K} samples spanning \textbf{16} tasks across \textbf{7} categories. 
This diversity ensures that SpatialScore serves as the most comprehensive benchmark to date for spatial intelligence, and that SpatialCorpus provides the high‑quality data necessary for supervised fine‑tuning~(SFT) on spatial reasoning tasks.

\section{More Implementation Details}
\label{sec:more_implementation_details}
In this section, we provide additional technical details of our work.
Concretely, we first describe the evaluation setup used for the SpatialScore benchmark in Sec~\ref{subsec:spatialscore_evaluation};
Next, we outline the procedures for supervised fine-tuning with SpatialCorpus in Sec~\ref{subsec:supervised_fine-tuning_with_spatialcorpus};
Then, we present the development details of SpatialAgent in Sec~\ref{subsec:spatialagent_development};
Finally, Sec~\ref{subsec:toolbox_specifications} introduces the design of the spatial perception expert models included in the toolbox, particularly highlighting instruction prompts.

\subsection{SpatialScore Evaluation}
\label{subsec:spatialscore_evaluation}
\noindent \textbf{Hyper-parameters.} 
To ensure reproducibility, we standardize the following configurations: 
all models adopt deterministic sampling~(TEMPERATURE=0.0, DO\_SAMPLE=False) and a maximum output length of 512 tokens, except for reasoning-oriented models such as KimiVL-16B-A3B-Thinking~\cite{team2025kimivl} and LLaMA-3.2V-11B-CoT~\cite{xu2024llama-cot}, which are allocated 2048 tokens.
For our \textbf{SpatialAgent}, we set the maximum attempt limit to 3 iterations under the {\em Plan-Execute} paradigm and permit 10 dialogue turns for {\em ReAct} interactions. 
To accommodate the extended reasoning requirements in multi-agent collaboration, the token limit is correspondingly increased to 4096 for these cases.

\vspace{2pt}
\noindent \textbf{Baselines.}
Our chance-level~(Random) baseline is implemented as follows: 
For judgment and multi-choice questions, we randomly sample an answer based on the number of available options. 
For open‑ended questions, to ensure that the baseline yields a reasonably meaningful result, we uniformly sample a value within a range of 0.25 to 4 times the ground‑truth value as the final answer.
Additionally, we employ a powerful model~({\em i.e.}, GPT-5~\cite{openai_gpt5_systemcard}) with text-only input to serve as a blind baseline. 
For the human-level evaluation, we invite three PhD students with extensive experience in 3D vision research to answer the questions using basic computational tools~({\em e.g.}, OpenCV), and report their average score.

\vspace{2pt}
\noindent \textbf{Prompts.}
Since SpatialScore encompasses samples from diverse data sources, covering judgment, multi-choice, and open-ended questions, we carefully design tailored system prompts for each format to ensure models can properly follow instructions and produce correctly formatted answers, as detailed below:

\vspace{4pt}

For judgment~({\em i.e.}, Yes/No) questions, we employ the following prompt to guide models to provide their determined correct answers:

{
    \begin{tcolorbox}[breakable]
    **Please answer with yes or no based on the image.**
    
    **Respond ONLY with 'yes' or 'no'.**
    
    Question: \{{\em question}\}
    \end{tcolorbox}
}

\vspace{4pt}

For multi-choice questions, we expect models to concisely output their selected option, with the following prompt:

{
    \begin{tcolorbox}[breakable]  
    **Please select the most appropriate answer from the given options.**
    
    **Respond ONLY with the capital letter and its parentheses.**
    
    Question: \{{\em question}\}
    \end{tcolorbox}
}

\vspace{4pt}

For open-ended questions, we first address those that require estimating metric‑based distances or sizes. 
In these cases, we adopt the following prompt to guide the model to provide both a numerical value and its corresponding unit of measurement~({\em e.g.}, meter).

{
    \begin{tcolorbox}[breakable]
    Please answer the question by measuring the precise distance in 3D space through 2D images or videos.
    
    Respond ONLY with a numeric answer consisting of a scalar and a distance unit in the format of **scalar \{{\em scalar}\} distance\_unit \{{\em distance unit}\}**.

    Question: \{{\em question}\}
    \end{tcolorbox}
}

\vspace{4pt}

For other types of open-ended questions, such as counting, we utilize the following prompt:

{
    \begin{tcolorbox}[breakable]
    Please answer the question based on the given image or video.
    
    Respond ONLY with a concise and accurate scalar or a scalar with corresponding unit.**

    Question: \{{\em question}\}
    \end{tcolorbox}
}

\vspace{4pt}

To ensure a fair evaluation, we employ identical prompts for all baseline models~(also including thinking and proprietary models).
Notably, we have carefully designed answer-parsing functions for all models to ensure accurate extraction of final answers. 
However, some responses still fail to comply with instruction prompts or result in refusal to answer~(primarily in certain fine-tuned models and smaller-scale architectures).
Therefore, when computing overall accuracy, we adopt the following strategy:
For judgment and multi-choice questions, we apply the parsing functions to extract the models' answers and directly compare them with the ground truth.
For open-ended questions, we report the average Mean Relative Accuracy~(MRA)~\cite{VSI-Bench} obtained by two methods:
(i) extracting the models' answers from responses with the meticulously crafted parsing function to calculate the score,
and
(ii) use an off-the-shelf LLM~(GPT-OSS-20B~\cite{agarwal2025gpt-oss}) to score the response with the following prompt:

{
    \begin{tcolorbox}[breakable]
    You are an evaluator. 
    
    Your ONLY job is to compute a score using the following algorithm. 
    
    Do NOT answer or solve the question.
    \\
    \\
    TASK TYPE:
    
    - If Type == "counting": treat both GT and PRED as plain scalar numbers (no unit conversion).
    
    - If Type == "distance": parse numeric value + unit; if PRED unit is missing, borrow GT unit; if both are missing and both look like plain numbers, treat as scalar.
    
    - If a numeric RANGE like "10-15" appears, use the MAX value (e.g., 15).
    \\
    \\
    ALGORITHM (VSI-Bench MRA):
    
    1) Compute abs\_dist\_norm:
       
       - For scalar/counting: abs\_dist\_norm = abs(pred - gt) / gt   (if gt == 0, set abs\_dist\_norm = +Infinity)
       
       - For distance: convert both to centimeters using:
           meter~(m): 100 cm; 
           centimeter~(cm): 1 cm; 
           millimeter~(mm): 0.1 cm; 
           inch~(in): 2.54 cm; 
           foot~(ft): 30.48 cm.
         
         Then abs\_dist\_norm = abs(pred\_cm - gt\_cm) / gt\_cm  (if gt\_cm == 0, set +Infinity).
    \\
    \\
    2) For thresholds C = linspace(start, end, steps) with steps = int((end-start)/interval+2):
        
        \quad\quad accuracy(C) = 1 if abs\_dist\_norm <= (1 - C) else 0
       
        \quad\quad mean\_relative\_accuracy = average of accuracy(C) over all thresholds.
    \\
    \\
    3) The final score is this mean\_relative\_accuracy, a float in [0,1].
    \\
    \\
    IMPORTANT OUTPUT RULE:
    
    - After you finish the calculation, OUTPUT EXACTLY ONE LINE at the end in the form: output: <float> 
    For example: output: 0.83
    \\
    \\
    Config:
    
    - start=\{{\em start}\}
    
    - end=\{{\em end}\}
    
    - interval=\{{\em interval}\}
    \\
    \\
    Inputs:
    
    - Type: \{{\em open\_type}\}   \# "counting" or "distance"
    
    - gt\_answer: \{{\em gt\_answer}\}
    
    - pred\_answer: \{{\em pred\_answer}\}
    \end{tcolorbox}
}

\vspace{4pt}

\subsection{Supervised Fine-tuning with SpatialCorpus}
\label{subsec:supervised_fine-tuning_with_spatialcorpus}
To efficiently and effectively enhance MLLMs’ performance on spatial understanding tasks, we construct \textbf{SpatialCorpus}, a large-scale training resource consisting of 331K multimodal QA pairs spanning 16 distinct tasks across 7 categories. 
It includes both single‑frame and multi‑frame inputs and covers judgment, multi‑choice, and open‑ended QA formats, serving as a large-scale dataset for supervised fine‑tuning~(SFT).
To be specific, we fine‑tune Qwen3‑VL‑4B and Qwen3‑VL‑8B for one epoch on SpatialCorpus using $8\times$ Nvidia A100 GPUs, with bfloat16 precision, a batch size of 512, a peak learning rate of $1\times10^{-5}$, and a warm-up ratio of 3\%. 
The visual encoder is kept fixed, while the MLP that projects visual features to language space and the LLM parameters are jointly optimized.

\subsection{SpatialAgent Development}
\label{subsec:spatialagent_development}
To build \textbf{SpatialAgent}, a multi-agent system tailored for spatial intelligence and capable of using open-source MLLMs~({\em e.g.}, Qwen3-VL~\cite{Qwen3-VL}) as the agent core, we have meticulously designed a series of instruction prompts that guide the agent core~($\Phi$) to serve as different components within the system. 
These prompts enable SpatialAgent to invoke proper spatial perception tools and think step-by-step through two distinct paradigms: {\em Plan-Execute} and {\em ReAct}, thereby improving the spatial understanding capabilities of MLLMs in a training-free manner. 
Details are presented below.

\vspace{4pt}

\vspace{2pt}
\noindent \textbf{Plan-Execute Paradigm.}
For the {\em Plan-Execute} paradigm, SpatialAgent primarily consists of three components: {\em planner}~($\Phi_{\mathrm{plan}}$), {\em executor}~($\Phi_{\mathrm{exe}}$), and {\em summarizer}~($\Phi_{\mathrm{sum}}$). 
First, we use the following prompt to guide the {\em planner}~($\Phi_{\mathrm{plan}}$) in formulating a detailed tool invocation plan based on the descriptions in the toolbox:

{
    \begin{tcolorbox}[breakable]
    [BEGIN OF GOAL]
    
    Generate a JSON-formatted tool-calling plan to solve spatial understanding questions about given images or videos.
    
    [END OF GOAL]
    \\
    
    [BEGIN OF TOOLBOX]

    \{{\em action\_details}\}

    [END OF TOOLBOX]
    \\

    [BEGIN OF TASK INSTRUCTIONS]

    Generate a step-by-step plan to answer the given spatial understanding question about given images or videos.
    
    ***Use ONLY the tools listed in the TOOLBOX section (e.g., GetObjectOrientation, EstimateObjectGeometryProperties, LocalizeObjects, EstimateObjectDepth)***
    
    ***Follow their argument specifications EXACTLY as defined in the toolbox, and try to give detailed and comprehensive instructions in queries.***
    
    Do NOT invent new tools or modify the existing tool interfaces.
    
    The plan should strictly follow what these tools can and cannot do.

    [END OF TASK INSTRUCTIONS]
    \\
    
    [BEGIN OF FORMAT INSTRUCTIONS]

    You are a helpful assistant tasked with solving spatial reasoning questions. Think step by step.
    
    ***
    
    Return a JSON list of tool calls inside ```json``` tags, where each call is a dictionary with 'name' and 'arguments'.
    
    The 'name' MUST match exactly one of the tool names provided in the toolbox.
    
    The 'arguments' MUST include ALL required parameters for that specific tool with EXACT parameter names.
    
    The 'images' or 'image' argument must be specified as 'image-0', 'image-1', and 'image-2', to refer to the provided images. 
    
    Do not answer the question directly, and do not use absolute paths for the 'images' or 'image' argument.
    
    ***
    
    [END OF FORMAT INSTRUCTIONS]
    \\

    [BEGIN OF EXAMPLES]

        Example for 'Which is closer to the camera, the dog or the cat?':
    
        ```json
        \quad [ 
        
        \quad\quad \{"name": "LocalizeObjects", "arguments": \{"image": "image-0", "objects": ["dog", "cat"]\}\},
        
        \quad\quad \{"name": "EstimateObjectDepth", "arguments": \{"image": "image-0", "objects": ["dog", "cat"], "indoor\_or\_outdoor": "outdoor"\}\},
        
        \quad ]
        
        ```
        
    [END OF EXAMPLES]
    \\

    ***
    
    Do not answer the question directly. Instead, think step-by-step, and output the tool-calling plan inside ```json``` tags.
    
    ***

    \end{tcolorbox}
}

\vspace{4pt}

Subsequently, the {\em executor}~($\Phi_{\mathrm{exe}}$) follows the prompt below to sequentially execute tool invocations according to the plan and obtain the tool execution results.

{  
    \begin{tcolorbox}[breakable]

    [BEGIN OF GOAL]
    
    Generate a Chain of Thought (CoT) reasoning process using the provided tool execution results.
        
    [END OF GOAL]
    \\
    
    [BEGIN OF TASK INSTRUCTIONS]

    You are a helpful assistant tasked with solving spatial reasoning questions. 
    Analyze the given question and tool execution results. Think step by step. 
    
    Generate a step-by-step reasoning process that shows how the tools contribute to solving the question. 
    
    Use ONLY the tools and results provided, following their specifications STRICTLY.
    
    The results of tool calls can sometimes be incomplete or incorrect, so please be critical and decide how to make use of them. 
    
    If a tool failed, note the failure and proceed with your prior knowledge and reasoning.
    
    Repeat for each tool result in order. 
    
    [END OF TASK INSTRUCTIONS]
    \\
    
    [BEGIN OF FORMAT INSTRUCTIONS]
    
    ***
    
    Output a CoT with:
    
    \quad - <thinking> Explain why this tool was used and how its result contributes to the answer. </thinking>
    
    \quad - <tool> The tool call in JSON format, e.g., \{\{"name": "LocalizeObjects", "arguments": \{\{"image": "image-0", "objects": ["dog", "cat"]\}\}\}\}. </tool>
    
    \quad - <observation>: The tool result as a string. </observation>
    
    Repeat for each tool result in order.
    
    ***
    
    [END OF FORMAT INSTRUCTIONS]
    \\
    
    [BEGIN OF EXAMPLES]
    
    Example for 'In image-0, which is closer to the camera, the dog or the cat?':
    
    \quad <thinking> To determine which object is closer to the camera, I need first localize the dog and cat in the image. </thinking>
    
    \quad <tool> \{\{"name": "LocalizeObjects", "arguments": \{\{"image": "image-0", "objects": ["dog", "cat"]\}\}\}\} </tool>

    \quad <observation> \{\{"results": [\{\{"label": "dog", "region": [0.5, 0.6, 0.6, 0.8], "confidence": 0.95\}\}, \{\{"label": "cat", "region": [0.4, 0.5, 0.45, 0.7], "confidence": 0.87\}\}]\}\} </observation>\\
    
    \quad <thinking> The bounding box for the dog is [0.5, 0.6, 0.6, 0.8], and for the cat is [0.4, 0.5, 0.45, 0.7]. Then, I need estimate the depth of them to reflect their distances to the camera. </thinking>
    
    \quad <tool> \{\{"name": "EstimateObjectDepth", "arguments": \{\{"image": "image-0", "objects": ["dog", "cat"], "indoor\_or\_outdoor": "outdoor"\}\}\}\} </tool>
    
    \quad <observation> \{\{"results": [\{\{"object": "dog", "depth": 1.0, "error": null\}\}, \{\{"object": "cat", "depth": 1.2, "error": null\}\}]\}\} </observation>
    
    [END OF EXAMPLES]
    \\
    
    Tool Plan: \{{\em tool\_plan}\}
    
    Tool Results: \{{\em tool\_results}\}\\
    
    ***
    
    **Notably, you should AVOID outputting terms like <final\_thining>, <answer>, or <final\_answer> here.** 
    
    **Now, output your reasoning between <thinking> and </thinking>, the tool call in JSON format between <tool> and </tool>, and the observation between <observation> and </observation>.**
    
    ***
    
    \end{tcolorbox}
}

\vspace{4pt}

Finally, the {\em summarizer}~($\Phi_{\mathrm{sum}}$) consolidates the tool execution results and produces the final reasoning and answer, guided by the following instruction prompt:

{
    \begin{tcolorbox}[breakable]
    [BEGIN OF GOAL]
    
    Generate a final REASONING and ANSWER for spatial understanding questions about given images or videos, based on tool results and prior Chain of Thought (CoT) steps.
    
    [END OF GOAL]
    \\
    
    [BEGIN OF TASK INSTRUCTIONS]
    
    You are a helpful assistant tasked with solving spatial reasoning questions. 
    
    Given the question, tool execution results, and CoT steps, synthesize the information to provide a final REASONING and ANSWER. 
    
    **The results of tool calls can sometimes be incomplete or incorrect, so please be critical and decide how to make use of them.** 
    
    If tool results are unclear or contradictory, use your prior knowledge to think the problem step-by-step.

    For multi-choice questions, select the most appropriate answer from options based on reasoning. 
    Respond ONLY with the capital letter and its parentheses.
    
    For judgment questions, answer with yes or no based on reasoning. 
    Respond ONLY with 'yes' or 'no'.
    
    For open-ended measurement questions, answer the question by measuring the precise distance in 3D space through a 2D images or videos. 
    DO NOT use generic and unclear units like 'units' or 'pixels'
    
    Respond ONLY with a numeric answer consisting of a scalar and a distance unit in the format of **scalar distance\_unit**.
    
    For other questions, answer the question based on the given image or video. 
    Respond ONLY with a concise and accurate scalar or a scalar with corresponding unit.
    
    **CRITICAL: You MUST always provide a reasonable answer. Never respond with 'cannot be determined', 'none of the above', or similar phrases.**
        
    [END OF TASK INSTRUCTIONS]
    \\
    
    [BEGIN OF FORMAT INSTRUCTIONS]
    
    ***
    
    Output:
    
    \quad - <thinking> A complete analysis synthesizing all tool results and CoT steps to derive the answer. </thinking>
    
    \quad - <answer> **The final answer** </answer>
    
    ***
    
    [END OF FORMAT INSTRUCTIONS]
    \\
    
    CoT Steps: 
    \{{\em cot\_steps}\}\\
    
    ***
    
    CRITICAL: You MUST always provide a reasonable answer. Never respond with 'cannot be determined', 'none of the above', or similar phrases.**
    
    Now, output **your thinking** between <thinking> and </thinking>, and **your answer** between <answer> and </answer>.
    
    ***
    
    \end{tcolorbox}
}

\vspace{4pt}

Moreover, in the {\em Plan-Execute} paradigm, scenarios may arise where either (i) the {\em planner}~($\Phi_{\mathrm{plan}}$) fails to generate a correct plan, or (ii) the {\em executor}~($\Phi_{\mathrm{exe}}$) encounters tool invocation failures.
To address this, we set a maximum attempt threshold~(default to 3).
When the system exceeds this limit without completing the {\em Plan-Execute} reasoning process, SpatialAgent will bypass the workflow and directly answer the question using the following prompt:

{
    \begin{tcolorbox}[breakable]
    [BEGIN OF GOAL]
    
    Provide a direct ANSWER to a spatial understanding question about given 2D images or videos without external tools.
    
    [END OF GOAL]
    \\
    
    [BEGIN OF TASK INSTRUCTIONS]
    
    You are a helpful assistant tasked with solving spatial reasoning questions. Think step by step. 
    
    Answer the spatial understanding question by reasoning about the provided images or videos. 
    
    Provide a direct answer by reasoning logically based on typical spatial relationships and visual cues in the images or videos.

    **CRITICAL: You MUST always provide a reasonable answer. Never respond with 'cannot be determined', 'none of the above', or similar phrases.**

    ***
    
    For multi-choice questions, select the most appropriate answer from options based on reasoning. Respond ONLY with the capital letter and its parentheses.
    
    For judgment questions, answer with yes or no based on reasoning. Respond ONLY with 'yes' or 'no'.
    
    For open-ended measurement questions, answer the question by measuring the precise distance in 3D space through 2D images or videos. DO NOT use generic and unclear units like 'units' or 'pixels'.
    
    Respond ONLY with a numeric answer consisting of a scalar and a distance unit in the format of **scalar distance\_unit**.
    
    For other questions, answer the question based on the given image or video. Respond ONLY with a concise and accurate scalar or a scalar with corresponding unit.
    
    ***

    [END OF TASK INSTRUCTIONS]
    \\
    
    [BEGIN OF FORMAT INSTRUCTIONS]

    ***
    
    Output your response in the format:
    
    \quad <thinking> [Your reasoning here] </thinking>
    
    \quad <answer> [Your final answer] </answer>

    ***
    
    [END OF FORMAT INSTRUCTIONS]
    \\
    
    ***
    
    **CRITICAL: You MUST always provide a reasonable answer. Never respond with 'cannot be determined', 'none of the above', or similar phrases.**
    
    Now, output **your thinking** between <thinking> and </thinking>, and **your answer** between <answer> and </answer>.
    
    ***
    
    \end{tcolorbox}
}

\vspace{4pt}

\vspace{2pt}
\noindent \textbf{ReAct Paradigm.}
For the {\em ReAct} paradigm, SpatialAgent comprises three components: {\em observer}~($\Phi_{\mathrm{obs}}$), {\em executor}~($\Phi_{\mathrm{exe}}$), and {\em summarizer}~($\Phi_{\mathrm{sum}}$).
By default, SpatialAgent can perform up to 10 rounds of dialogue iterations in this paradigm.
In each iteration, we use the following prompt to guide the {\em observer}~($\Phi_{\mathrm{obs}}$) to decide the next action based on the current context:

{
    
    \begin{tcolorbox}[breakable]
    USER REQUEST: \{{\em USER REQUEST}\}
    
    [BEGIN OF GOAL]

    You are a helpful assistant, and your goal is to solve the \# USER REQUEST \#. 
    
    You can either rely on your own capabilities or perform actions with external tools to help you. 
    
    A list of all available actions is provided to you below.
    
    [END OF GOAL]
    \\

    [BEGIN OF ACTIONS]
    
    \{for each action in actions\}
    
    [END OF ACTIONS]\\
    
    [BEGIN OF TASK INSTRUCTIONS]
    
    1. You must only select actions from \# ACTIONS \#.
    
    2. You can only call one action at a time.
    
    3. If no action is needed, please make actions an empty list (i.e., “actions”: []).
    
    4. You must always call **Terminate** with your final answer at the end.

    5. Please note that the priority of the SelfThinking tool is relatively low. Please give priority to using other tools, and only consider using this tool if the problem cannot be solved otherwise.

    [END OF TASK INSTRUCTIONS]\\
    
    [BEGIN OF TOOL USAGE INSTRUCTIONS]

    1. **Construct the correct image path** for the tool to use, ensuring the path can be accessed and read properly.
    
    2. For object distance and object size(Length, width, height,tall, short, slim, or heavy) problems, first observe the image. 
    If the scene is outdoors, **FIRST** use `LocalizeObjects` to obtain the 2D bounding boxes, then determine the pair of points (one from each object) that are closest to each other, and use these points as the 'point' inputs for `Get3DDistance` to get the distance between the two objects. 
    
    Do **NOT** simply use the center points of the boxes as the closest points between two objects.
    
    3. For counting-related problems, **USE** `CountObjects`; the number of returned points equals the number of objects.

    4. For camera-related problems, you may need to **USE** `GetCameraParametersVGGT` to obtain the camera parameters.
    
    ======================= CRITICAL WARNING =======================
    
    **DO NOT** invent or mention any tool that is **NOT explicitly defined** in \#ACTIONS\#.
    
    **DO NOT** fabricate tool usage results if you have NOT actually called the tool.

    You MUST only describe tool results that are actually obtained during execution.

    Violation of this rule is considered a **SERIOUS ERROR**.

    ================================================================

    ====================== RELIABILITY WARNING =====================

    If a tool result contains **ambiguous references** - for example, `LocalizeObjects` returns multiple bounding boxes for the same object - **the result is unreliable**.

    In such cases, you SHOULD rely on **reasoning** instead of depending on the tool output.

    Treat this as a high-risk situation and avoid making decisions solely based on such tool results.

    ================================================================

    =================== TOOL CHAIN LENGTH WARNING ==================

    If the tool invocation chain becomes **too long**, you MUST **STOP** calling further tools to avoid reaching the maximum number of allowed calls.

    In such cases, immediately switch to using **SelfThinking** to answer, **INCLUDING all input images** required for reasoning.

    Failure to follow this rule may result in task termination without producing a valid answer.

    ================================================================

    [END OF TOOL USAGE INSTRUCTIONS]\\

    [BEGIN OF FORMAT INSTRUCTIONS]

    Your output should be in a strict JSON format as follows:

    \{"thought": "the thought process, or an empty string", "actions": [\{"name": "action1", "arguments": \{"argument1": "value1", "argument2": "value2"\}\}]\}
    
    [END OF FORMAT INSTRUCTIONS]\\
    
    [BEGIN OF EXAMPLES]
    
    \{for each demo in {\em demo\_examples}\}
    
    [END OF EXAMPLES]\\

    \end{tcolorbox}
}

\vspace{4pt}

Next, we employ the following prompt to instruct the {\em executor}~($\Phi_{\mathrm{exe}}$) to perform the selected action and produce new intermediate results:

{
    
    \begin{tcolorbox}[breakable]
    OBSERVATION: \{{\em OBSERVATION}\}
    
    The OBSERVATION can be incomplete or incorrect, so please be critical and decide how to make use of it. 
    
    If you've gathered sufficient information to answer the question, call **Terminate** with the final answer. 
    
    Now, please generate the response for the next step.
    
    \end{tcolorbox}
}

\vspace{4pt}

Finally, the {\em summarizer}~($\Phi_{\mathrm{sum}}$) adopts a prompt similar to that of the {\em executor}~($\Phi_{\mathrm{exe}}$) to summarize all intermediate results and provide the final conclusion:

{
    
    \begin{tcolorbox}[breakable]
    ALL\_OBSERVATION: \{{\em ALL\_OBSERVATION}\}
    
    The ALL\_OBSERVATION can be incomplete or incorrect, so please be critical and decide how to make use of it. 
    
    Call **Terminate** with the final answer. 
    
    Now, please generate the response for the next step.
    
    \end{tcolorbox}
}

\vspace{4pt}

As mentioned above, when using the {\em observer}~($\Phi_{\mathrm{obs}}$) to select actions, we first provide a set of examples, as shown below. 
It is important to note that, since MLLMs may hallucinate when handling long contexts, for instance, pretending to call tools and producing fabricated results, we restrict these in‑context examples to be concise, focused, and unambiguous.

{
    \begin{tcolorbox}[breakable]
    [
        
        \quad\quad\{
        
            \quad\quad\quad\quad "user\_request": """Between image-0 and image-1, what is the primary direction of the camera's movement? Please answer with one of the following options: 
            A. The camera moved to the right
            B. The camera moved to the left
            C. The camera moved downward
            D. The camera moved upward""",
            
            \quad\quad\quad\quad "steps": [
                
                \quad\quad\quad\quad\quad\quad\{
                
                    \quad\quad\quad\quad\quad\quad\quad\quad"id": 1,
                    
                    \quad\quad\quad\quad\quad\quad\quad\quad"thought": "To determine the camera's movement direction, I need to compute the average optical flow between the two images using RAFT. The average optical flow indicates pixel offsets, where positive mean\_flow\_x suggests camera movement to the right, negative to the left, positive mean\_flow\_y downward, and negative upward.",
                    
                    \quad\quad\quad\quad\quad\quad\quad\quad"actions": [\{"name": "EstimateOpticalFlow", "arguments": \{"images": ["image-0", "image-1"]\}\}],
                    
                    \quad\quad\quad\quad\quad\quad\quad\quad"observation": \{"output": \{"mean\_flow\_x": 2.5, "mean\_flow\_y": -0.3\}\}
                
                \quad\quad\quad\quad\quad\quad\},
                
                \quad\quad\quad\quad\quad\quad\{
                    
                    \quad\quad\quad\quad\quad\quad\quad\quad"id": 2,
                    
                    \quad\quad\quad\quad\quad\quad\quad\quad"thought": "The optical flow results show mean\_flow\_x = 2.5 (positive, indicating camera moved right) and mean\_flow\_y = -0.3 (negative, indicating camera moved up). Since the absolute value of mean\_flow\_x (2.5) is much larger than mean\_flow\_y (0.3), the primary camera movement is to the right.",
                    
                    \quad\quad\quad\quad\quad\quad\quad\quad"actions": [\{"name": "Terminate", "arguments": \{"answer": "(A)"\}\}],
                    
                    \quad\quad\quad\quad\quad\quad\quad\quad"observation": \{\}
                
                \quad\quad\quad\quad\quad\quad\}
            
            \quad\quad\quad\quad]
        
        \quad\quad\},
        
        \quad\quad\{   
        
            \quad\quad\quad\quad"user\_request": """In image-0, which direction is the person facing?
            A. Facing the viewer while slightly to the right
            B. Facing the viewer while slightly to the left
            C. Facing away from the viewer while slightly to the right
            D. Facing away from the viewer while slightly to the left""",
            
            \quad\quad\quad\quad"steps": [
                
                \quad\quad\quad\quad\quad\quad\{
                
                    \quad\quad\quad\quad\quad\quad\quad\quad"thought": "To determine the precise orientation of the person, I need to use GetObjectOrientation to analyze their position.",
                    
                    \quad\quad\quad\quad\quad\quad\quad\quad"actions": [\{"name": "GetObjectOrientation", "arguments": \{"image": "image-0", "objects": "person"\}\}],
                    
                    \quad\quad\quad\quad\quad\quad\quad\quad"observation": \{"results": [\{\"object": "person", "angle\_data": \{"azimuth": 315.0, "polar": 90.0, "rotation": 0.0, "confidence": 0.89\}, "error": 'null'\}]\}
                
                \quad\quad\quad\quad\quad\quad\},
                
                \quad\quad\quad\quad\quad\quad\{
                    
                    \quad\quad\quad\quad\quad\quad\quad\quad"thought": "The person's azimuth angle is 315 degrees, which falls in the range of 292.5° < $\phi$ < 337.5°. According to the orientation guidelines, this means the person is facing the viewer and to the right of the viewer.",
                    
                    \quad\quad\quad\quad\quad\quad\quad\quad"actions": [\{"name": "Terminate", "arguments": \{"answer": "(A)"\}\}],
                    
                    \quad\quad\quad\quad\quad\quad\quad\quad"observation": \{\}
                
                \quad\quad\quad\quad\quad\quad\}
            
            \quad\quad\quad\quad]
        
        \quad\quad\},
        
        \quad\quad\{
        
        \quad\quad\quad\quad"user\_request": """In image-0, what is the distance between the mug and the laptop?
        A. 15 centimeters
        B. 29 centimeters
        C. 45 centimeters
        D. 60 centimeters""",
        
        \quad\quad\quad\quad"steps": [
            
            \quad\quad\quad\quad\quad\quad\{
            
                \quad\quad\quad\quad\quad\quad\quad\quad"thought": "First, I need to determine whether the scene in image-0 is indoors or outdoors. Observing the image, I see a desk, a mug, and a laptop in an office setting, indicating an indoor scene. Since it is indoors, according to the instruction, I do not need to call any tools and can directly estimate the distance.",
                
                \quad\quad\quad\quad\quad\quad\quad\quad"actions": [],
                
                \quad\quad\quad\quad\quad\quad\quad\quad"observation": \{\}
                
            \quad\quad\quad\quad\quad\quad\},
            
            \quad\quad\quad\quad\quad\quad\{
            
                \quad\quad\quad\quad\quad\quad\quad\quad"thought": "Based on the visual cues in the image, the distance between the mug and the laptop is approximately 29 centimeters.",
                
                \quad\quad\quad\quad\quad\quad\quad\quad"actions": [\{"name": "Terminate", "arguments": \{"answer": "(B)"\}\}],
                
                \quad\quad\quad\quad\quad\quad\quad\quad"observation": \{\}
            
            \quad\quad\quad\quad\quad\quad\}
        
        \quad\quad\quad\quad]
    
    \quad\quad\},
    
    \quad\quad\{
        
        \quad\quad\quad\quad"user\_request": """In image-0, which object on the desk is larger in size?
        A. The mug
        B. It can not be decided given the image only
        C. The laptop
        D. They seem to be of almost the same size""",
        
        \quad\quad\quad\quad"steps": [
            
            \quad\quad\quad\quad\quad\quad\{
                
                \quad\quad\quad\quad\quad\quad\quad\quad"thought": "First, I need to determine whether the scene in image-0 is indoors or outdoors. Observing the image, I see a desk with a mug and a laptop in an office setting, indicating an indoor scene(For object size questions (including length, width, height, tall, short, slim, or heavy), do NOT call any tools. You should answer directly based on the visual information from the image.). 
                Since it is indoors, according to the instruction, I do not need to call any tools and can directly estimate the relative size of the objects.",
                
                \quad\quad\quad\quad\quad\quad\quad\quad"actions": [],
                
                \quad\quad\quad\quad\quad\quad\quad\quad"observation": \{\}
            
            \quad\quad\quad\quad\quad\quad\},
            
            \quad\quad\quad\quad\quad\quad\{
                
                \quad\quad\quad\quad\quad\quad\quad\quad"thought": "Based on the visual cues in the image, the laptop is clearly larger in size compared to the mug.",
                
                \quad\quad\quad\quad\quad\quad\quad\quad"actions": [\{"name": "Terminate", "arguments": \{"answer": "(C)"\}\}],
                
                \quad\quad\quad\quad\quad\quad\quad\quad"observation": \{\}
            
            \quad\quad\quad\quad\quad\quad\}
        
        \quad\quad\quad\quad]
    
    \quad\quad\}    
    
    ]
    \end{tcolorbox}
}

\vspace{4pt}

\vspace{2pt}
\noindent \textbf{Exception Handling.}
Despite our carefully crafted prompts for SpatialAgent, it may occasionally produce malformed outputs or fall into infinite loops.
To mitigate this, beyond the prompt‑based error‑correction mechanisms that inspect abnormal intermediate results, we also verify whether the final extracted answer is an empty string or \texttt{null}.
If such failures occur, the agent system is downgraded to a single agent core to complete the task.
Concretely, when evaluated on the 5,025 samples in our SpatialScore benchmark, Qwen3‑VL‑4B‑SpatialAgent‑PE and Qwen3‑VL‑8B‑SpatialAgent‑PE exhibit 113 and 414 reasoning failures, corresponding to failure rates of 2.25\% and 8.24\%, respectively. 
In contrast, Qwen3‑VL‑4B‑SpatialAgent‑ReAct and Qwen3‑VL‑8B‑SpatialAgent‑ReAct each fail on only one sample, yielding a failure rate of just 0.02\%. 
This trend aligns with the characteristics of the {\em Plan‑Execute} and {\em ReAct} paradigms: the former is more efficient but lacks strong error‑correction capabilities, whereas the latter, though requiring more complex iterative reasoning, ensures higher stability and success rates.

\vspace{2pt}
\noindent \textbf{Computational Efficiency.}
When evaluated on our SpatialScore benchmark, Qwen3-VL-8B~\cite{Qwen3-VL} takes an average of 0.9 seconds per sample, whereas the {\em Plan-Execute} and {\em ReAct} reasoning paradigms require around 5.4 and 9.3 seconds, respectively, depending on the input length, interaction turns, and number of tool invocations.
As presented in Tab.~\ref{tab:tool_computation}, we report the GPU memory, latency, and utilization rate for each tool used by SpatialAgent, with Qwen3-VL-8B serving as its agent core, under the {\em Plan-Execute} paradigm.
All tools utilize GPU acceleration, excluding {\em terminate}, {\em point matching}, and {\em homography estimation}.
Our tools are lightweight and can be offloaded to the CPU when idle, allowing the entire reasoning process to execute on a single Nvidia A100 GPU.

\input{tables/tool_computation}

\subsection{Toolbox Specifications}
\label{subsec:toolbox_specifications}
To facilitate our proposed multi-agent system, \textbf{SpatialAgent}, to effectively perform visual reasoning for spatial understanding questions via tool invocation, we have designed detailed input-output descriptions for each tool, accompanied by concrete examples. 
These specifications serve as contextual information for the agent core to select proper expert tools, with the details elaborated as follows.

\vspace{4pt}

For general perception tools, we first implement the {\em LocalizeObjects} action using Rex-Omni~\cite{Rex-Omni}, which localizes objects based on given text prompts.  
The detailed tool specification is presented below:

{
    \begin{tcolorbox}[breakable]
        description = """
        
            \quad\quad Localize specific objects in an image. 
            
            \quad\quad Returns bounding boxes for target categories, optionally visualizing them.
        
        """
        
        args\_spec = \{
        
            \quad\quad "image": "The image to analyze.",

            \quad\quad "objects": "A list of object categories to detect."

        \}
        
        rets\_spec = \{"regions": "List of detected regions with label, bbox"\}
        
        examples = [\{
        
            \quad\quad "name": "LocalizeObjects", 
            
            \quad\quad "arguments": \{"image": "image-0", "objects": ["dog", "cat"]\}
        
        \}]
    \end{tcolorbox}
}

\vspace{4pt}

Next, we adopt Rex-Omni~\cite{Rex-Omni} to create {\em CountObjects} function, which can count specific objects based on given text prompts, with the following functionality explanation:

{
    \begin{tcolorbox}[breakable]
        description = """
        
            \quad\quad Count target objects in an image. Returns the coordinates of each detected target as points.
            
        """
        
        args\_spec = \{
            
            \quad\quad "image": "The image to analyze.",
            
            \quad\quad "objects": "List of object categories to count."
        
        \}
        
        rets\_spec = \{"points": "Dictionary \{category: [points...]\}, points in normalized coordinates.\}
        
        examples = [\{"name": "CountObjects", "arguments": \{"image": "image-0", "objects": ["bed"]\}]
    \end{tcolorbox}
}

\vspace{4pt}

Then, we integrate Rex-Omni~\cite{Rex-Omni} for advanced object localization and SAM2~\cite{SAM2} for precise segmentation, creating {\em GetObjectMask} function, with the following tool description:

{
    \begin{tcolorbox}[breakable]
        description = """
            
            \quad\quad Generate pixel-level segmentation masks for specified objects.
        
            \quad\quad Returns mask area ratios and bounding boxes for each detected object.
            
            \quad\quad Suitable for analyzing object shapes, sizes, and coverage.
        
        """
        
        args\_spec = \{
            
            \quad\quad "image": "Image file to process.",
            
            \quad\quad "objects": "List of object descriptions to localize and segment."
        
        \}
        
        rets\_spec = \{
        
            \quad\quad "results": "List of dicts with mask area ratio, bounding box, and optional error: [\{'object': str, 'mask\_area': float, 'bbox': [left, top, right, bottom], 'error': str or None\}]"
        
        \}
        
        examples = [
        
            \quad\quad \{"name": "GetObjectMask", "arguments": \{"image": "image-0", "objects": ["coffee mug", "microwave"]\}\}
            
        ]
    \end{tcolorbox}
}

\vspace{4pt}

Additionally, we employ DetAny3D~\cite{DetAny3D} as the tool for 3D object detection and build the {\em Detect3DObjects} module, with the detailed tool specifications provided below:

{
    \begin{tcolorbox}[breakable]
    description = """
            
        \quad\quad Detect specific objects in an image and estimate their 3D bounding boxes.
            
        \quad\quad Returns 3D bounding box parameters in the following format:

        \quad\quad x, y, z -> object center in camera coordinates (meters);
    
        \quad\quad width, height, length -> physical size (width, height, length) in meters;
        
        \quad\quad yaw -> heading angle around vertical axis (radians).
        
        """
        
        args\_spec = \{
        
            \quad\quad "image": "Path to the input image."

            \quad\quad "objects": "List of object categories to detect (or a single string)."
            
        \}
        
        rets\_spec = \{
        
            \quad\quad "objects": "List of dicts with \{label: str, bbox\_3d: \{x:float, y:float, z:float, width:float, height:float, length:float, yaw:float\}\}"
        
        \}
        
        examples = [
            
            \quad\quad \{"name": "Detect3DObjects", "arguments": \{"image": ["image-1"], "objects": ["dog", "rabbit"]\}
            
        ]
    \end{tcolorbox}
}

\vspace{4pt}

To further equip SpatialAgent with motion understanding and image transformation analysis, we have developed specialized expert tools such as {\em EstimateOpticalFlow} action implemented with RAFT~\cite{teed2020raft}:

{
    \begin{tcolorbox}[breakable]
    description = """
            
        \quad\quad Estimate optical flow between two images to measure motion in pixels.

        \quad\quad Returns average displacement in horizontal (x) and vertical (y) directions.
        
        \quad\quad First image is earlier in time; second is later. 

        \quad\quad - mean\_flow\_x > 0: objects move left / camera moves right.
        
        \quad\quad - mean\_flow\_x < 0: objects move right / camera moves left.
            
        \quad\quad - mean\_flow\_y > 0: objects move up / camera moves down.
        
        \quad\quad - mean\_flow\_y < 0: objects move down / camera moves up.

        \quad\quad Useful for analyzing camera motion, object movement, and 3D spatial reasoning.
        
        """
        
        args\_spec = \{
        
            \quad\quad "image": "A list of exactly two image paths to compute optical flow between. First image is earlier in time."
            
        \}
        
        rets\_spec = \{
        
            \quad\quad "output": "Dictionary containing 'mean\_flow\_x' (average horizontal pixel displacement) and 'mean\_flow\_y' (average vertical pixel displacement)."
        
        \}
        
        examples = [\{"name": "EstimateOpticalFlow", "arguments": \{"image": ["image-1", "image-3"]\}]
    \end{tcolorbox}
}

\vspace{4pt}

The {\em MatchImagesSIFT} functionality performs keypoint extraction and feature matching between images using SIFT~\cite{lowe2004SIFT} implemented via OpenCV.
The detailed specifications are as follows:

{
    \begin{tcolorbox}[breakable]
        description = """
            
            \quad\quad Match keypoints between two images using SIFT. 
            
            \quad\quad Detects distinctive features and returns matched coordinate pairs for tasks like alignment or recognition.
            
        """
        
        args\_spec = \{
        
            \quad\quad "image": "List of two image paths.",
            
            \quad\quad "num\_keypoints": "Max keypoints per image (default: 1200).",
            
            \quad\quad "ratio\_th": "Ratio test threshold for matching (default: 0.75)."
        
        \}
        
        rets\_spec = \{
        
            \quad\quad "matches": "List of matched coordinate pairs: [[x1, y1], [x2, y2]].",
            
            \quad\quad "num\_matches": "Total number of matches found."
        
        \}

        examples = [

            \quad\quad \{"name": "MatchImagesSIFT", "arguments": \{"image": ["image-0", "image-1"], "num\_keypoints": 1200, "ratio\_th": 0.75\}\}

        ]
    \end{tcolorbox}
}

\vspace{4pt}

The {\em EstimateHomographyMatrix} tool, also implemented via OpenCV, calculates the homography transformation matrix between two images based on extracted keypoints, with the following specification:

{
    \begin{tcolorbox}[breakable]
    description = """
            
            \quad\quad Compute a 3*3 homography matrix between two images using SIFT features and RANSAC.
            
            \quad\quad Useful for alignment, perspective correction, and planar transformations.

        """
        
        args\_spec = \{
            
            \quad\quad "image": "List of two image paths.",
            
            \quad\quad "num\_keypoints": "Max keypoints per image (default: 1200).",
            
            \quad\quad "ratio\_th": "Ratio test threshold (default: 0.75).",
            
            \quad\quad "ransac\_reproj\_threshold": "Max reprojection error in RANSAC (default: 5.0)."
        
        \}
        
        rets\_spec = \{
            
            \quad\quad "homography\_matrix": "3*3 matrix mapping points from first image to second.",
            
            \quad\quad "inliers\_count": "Number of inlier matches used.",
            
            \quad\quad "total\_matches": "Total matches found.",

            \quad\quad "status": "Success or failure."
        
        \}

        examples = [
            
            \quad\quad \{"name": "EstimateHomographyMatrix", "arguments": \{"image": ["image-0", "image-1"], "num\_keypoints": 1200, "ratio\_th": 0.75, "ransac\_reproj\_threshold": 5.0\}\}

        ]
    \end{tcolorbox}
}

\vspace{4pt}

Moreover, VGGT~\cite{wang2025vggt} can predict the camera intrinsics and extrinsics for each frame within a frame sequence.
We implement the {\em GetCameraParametersVGGT} module using the following tool description:

{
    \begin{tcolorbox}[breakable]
    description = """
            
            \quad\quad Extract camera extrinsic (3*4, relative to first image) and intrinsic (3*3) parameters from images using VGGT.
            
            \quad\quad Useful for 3D reconstruction, novel view synthesis, and geometric analysis.

        """
        
        args\_spec = \{"image": "List of image paths (at least one)."\}
        
        rets\_spec = \{
            
            \quad\quad "output": "List of dicts with image\_index (int), extrinsic (3*4 matrix), and intrinsic (3*3 matrix)."
        
        \}

        examples = [
            
            \quad\quad\{"name": "GetCameraParametersVGGT", "arguments": \{"image": ["image-0", "image-1"]\}\}

        ]
    \end{tcolorbox}
}

\vspace{4pt}

Additionally, to empower SpatialAgent with 3D spatial reasoning capabilities, we have integrated several specialized visual geometry models.
The {\em EstimateObjectGeometryProperties} function is implemented via the integration of SAM2~\cite{SAM2}, Depth-Anything-V2~\cite{DepthAnythingv2}, and VGGT~\cite{wang2025vggt} to obtain detailed spatial and geometry properties of objects in the given image, with the following tool specification:

{
    \begin{tcolorbox}[breakable]
        description = """
        
            \quad\quad Analyze objects in an image to obtain bounding boxes, mask areas, depth (m), and camera parameters.
            
            \quad\quad Camera parameters include intrinsic (3*3) and extrinsic (3*4) matrices for 3D geometry tasks.
        
        """
        
        args\_spec = \{
            
            \quad\quad "image": "Image file path to analyze.",
            
            \quad\quad "object\_descs": "List of object descriptions (e.g., ['dog', 'cat'])."
        
        \}
        
        rets\_spec = \{
            
            \quad\quad "results": "List of dicts with object, bbox, mask\_area, depth (m), and optional error.",
        
            \quad\quad "camera\_parameters": "Dict with intrinsic (3*3) and extrinsic (3*4) matrices."
        
        \}
        
        examples = [
            
            \quad\quad \{"name": "EstimateObjectGeometryProperties", "arguments": \{"image": "image-0", "object\_descs": ["coffee cup", "keyboard"]\}\}

        ]
    \end{tcolorbox}
}

\vspace{4pt}

We have also implemented the {\em EstimateRegionDepth} action with Depth-Anything-V2~\cite{DepthAnythingv2} for scene depth estimation and region-specific average depth calculation based on given 2D bounding boxes, with the following tool description:

{
    \begin{tcolorbox}[breakable]
        description = """
            
            \quad\quad Estimate metric depth (in meters) of specified regions in an image.
            
            \quad\quad Supports indoor (0-20m) and outdoor (0-80m) scenes.
            
            \quad\quad Works with single or multiple bounding boxes in pixel coordinates.
            
            \quad\quad Depth is distance from camera to object, not between objects or object size.
            
        """
        
        args\_spec = \{
            
            \quad\quad "image": "Image to analyze.",
            
            \quad\quad "bboxes": "Bounding box or list of boxes in pixel coordinates: [left, top, right, bottom] or [[...], ...].",
            
            \quad\quad "indoor\_or\_outdoor": "Scene type ('indoor' or 'outdoor').",
            
            \quad\quad "mode": "Depth calculation: 'mean' (average) or 'center' (center point). Default: 'mean'."
        
        \}
        
        rets\_spec = \{
            
            \quad\quad "depths": "List of dicts with bbox, depth (m), and optional error: [{'bbox': list, 'depth': float, 'error': str or None}]",
            
            \quad\quad "unit": "Always 'meters'."
        
        \}
        
        examples = [
            
            \quad\quad \{"name": "EstimateRegionDepth", "arguments": \{"image": "image-0", "bboxes": [100, 50, 200, 150], [150, 100, 250, 200] "indoor\_or\_outdoor": "indoor"\}\}
        
        ]
    \end{tcolorbox}
}

\vspace{4pt}

By combining Rex-Omni~\cite{Rex-Omni} and Depth-Anything-V2~\cite{DepthAnythingv2}, we also facilitate {\em EstimateObjectDepth}, with the corresponding specification as follows:

{
    \begin{tcolorbox}[breakable]
        description = """

            \quad\quad Estimate object depth (in meters) from an image.
            
            \quad\quad Supports indoor (0-20m) and outdoor (0-80m) scenes.
            
            \quad\quad Depth indicates distance from camera to object, not between objects or object size.
            
        """
        
        args\_spec = \{
            
            \quad\quad "image": "Image to analyze.",
            
            \quad\quad "objects": "List of object descriptions to measure distance to (e.g., ['dog', 'cat']).",
            
            \quad\quad "indoor\_or\_outdoor": "Scene type ('indoor' or 'outdoor')."

        \}
        
        rets\_spec = \{
        
            \quad\quad "results": "List of dicts with object description, depth (m), and optional error: [{'object': str, 'depth': float, 'error': str or None}]"
        
        \}
        
        examples = [
            
            \quad\quad \{"name": "EstimateObjectDepth", "arguments": \{"image": "image-0", "objects": ["the red car", "dog"], "indoor\_or\_outdoor": "outdoor"\}\}

        ]
    \end{tcolorbox}
}

\vspace{4pt}

OrientAnything~\cite{orient_anything} model is employed for the {\em GetObjectOrientation} functionality:

{
    \begin{tcolorbox}[breakable]
        description = """
            
            \quad\quad Estimate 3D orientation of objects in an image using Orient-Anything.

            \quad\quad Measures:
            
            \quad\quad - Azimuth: Horizontal rotation (0-360° clockwise)
            
            \quad\quad - Polar: Vertical inclination (0-180°)
            
            \quad\quad - Rotation: In-plane rotation (-180° to +180°)
            
            \quad\quad - Confidence: Reliability score

            \quad\quad Useful for 3D understanding, pose estimation, and spatial reasoning.
        
        """
        
        args\_spec = \{
            
            \quad\quad "image": "Image to analyze.", 
            
            \quad\quad "objects": "Object description(s) to analyze; string or list."
        
        \}
        
        rets\_spec = \{
            
            \quad\quad "results": "List of dicts with object orientation data: [\{'object': str, 'angle\_data': \{'azimuth': float, 'polar': float, 'rotation': float, 'confidence': float\}, 'error': str or None\}]"
        
        \}
        
        examples = [
            
            \quad\quad \{"name": "GetObjectOrientation", "arguments": \{"image": "image-0", "objects": "a red car"\}\}
            
        ]
    \end{tcolorbox}
}

\vspace{4pt}

To estimate metric-based distances between points in 3D space, we employ MapAnything~\cite{MapAnything} to reconstruct the 3D scene and compute the real-world distances, following the specification below:

{
    \begin{tcolorbox}[breakable]
        description = """
            
            \quad\quad Calculates the absolute 3D spatial distance (in meters) between two pixel points (x, y) in an image. 
            
            \quad\quad Note: this tool should be used in outdoor scenes.
            
            \quad\quad Returns the calculated distance (in meters).
        
        """
        
        args\_spec = \{
            
            \quad\quad "image": "Path to the input image.", 
            
            \quad\quad "point\_1": "List of [x, y] pixel coordinates for the first point."

            \quad\quad "point\_2": "List of [x, y] pixel coordinates for the second point."
        
        \}
        
        rets\_spec = \{
            
            \quad\quad "distance\_meters": "The calculated 3D distance (float, in meters)."
        
        \}
        
        examples = [
            
            \quad\quad \{"name": "Get3DDistance", "arguments": \{"image": "image-0", "point\_1": [100, 100], "point\_2": [1000, 1000]\}\}
            
        ]
    \end{tcolorbox}
}

\vspace{4pt}

Moreover, we have developed a suite of general-purpose tools to assist in tool invocation and reasoning processes.
We design a dedicated {\em Terminate} action to formally conclude the reasoning process and give structured final answers.
The tool function description is as follows:

{
    \begin{tcolorbox}[breakable]
    
    description = """
            
            \quad\quad Use this function ONLY when you are completely confident in your final answer.
            
            \quad\quad For multiple-choice questions: Specify the letter of the correct option.
            
            \quad\quad For numerical answers: Include both the specific value and appropriate unit of measurement (e.g., meter or centimeter).
            
            \quad\quad For yes/no questions: Clearly state 'Yes' or 'No'.
            
            \quad\quad DO NOT call this function if you are uncertain or need to perform additional analysis.
            
            \quad\quad Double-check your answer before terminating!
        
        """
        
        args\_spec = \{
        
            \quad\quad "answer": "The final answer with proper formatting. For multiple choice: include letter (e.g., 'A. explanation' or '(B)'). For numerical answers: include units (e.g., '3.25 meters')."
            
        \}
        
        rets\_spec = \{
            
            \quad\quad "answer": "The final answer that will be submitted."
            
        \}
        
        examples = [
        
            \quad\quad \{"name": "Terminate", "arguments": \{"answer": "A. Yes."\}\},
            
            \quad\quad \{"name": "Terminate", "arguments": \{"answer": "(B)."\}\},
            
            \quad\quad \{"name": "Terminate", "arguments": \{"answer": "B. 3.25 meters."\}\},
            
            \quad\quad \{"name": "Terminate", "arguments": \{"answer": "(A) 2 inches."\}\},
            
            \quad\quad \{"name": "Terminate", "arguments": \{"answer": "47.3 centimeters."\}\},
            
            \quad\quad \{"name": "Terminate", "arguments": \{"answer": "38.2 degrees."\}\}
                
        ]
    \end{tcolorbox}
}

\vspace{4pt}

A {\em SelfThinking} module is designed to guide the agent core~({\em e.g.}, Qwen3-VL~\cite{Qwen3-VL}) in self-reflecting on questions through meticulous prompt engineering, thereby fully leveraging their inherent potential to better tackle spatial understanding tasks, with the specification detailed as below.

{
    \begin{tcolorbox}[breakable]
    description = """
        
        \quad\quad Modes:
        
        \quad\quad 1. Text-only: Provide 'query' for pure language tasks.
            
        \quad\quad 2. Vision+Language: Provide 'images' + 'query' for visual analysis.
        
        \quad\quad Suitable for: Scene understanding, OCR, object/color recognition, classification, and concept-level Q\&A.
        
        """
        
        args\_spec = \{

            \quad\quad "query": "Text question or instruction (REQUIRED).",
            
            \quad\quad "image": "List of image paths. If omitted, the model performs text-only reasoning.",
            
        \}
        
        rets\_spec = \{"response": "Model's response string."\}
        
        examples = [
            
            \quad\quad \{"name": "SelfThinking", "arguments": \{"query": "Summarize the image content.", "image": "image-0"\}\}
        
        ]
    \end{tcolorbox}
}

\vspace{4pt}

\section{More Experiment Results}
\label{sec:more_experiment_results}
In this section, we present additional quantitative and qualitative results in Sec~\ref{subsec:additional_quantitative_results} and Sec~\ref{subsec:additional_qualitative_results}, followed by comprehensive and in-depth analyses and discussions.

\input{tables/quantitative_results_open_source}

\input{tables/quantitative_results_ours}

\subsection{Additional Quantitative Results}
\label{subsec:additional_quantitative_results}
Here, we further conduct an in-depth analysis of the quantitative results on our proposed \textbf{SpatialScore} benchmark. 
Concretely, we divide the 5,025 samples in SpatialScore into two subsets: 
(i) the newly constructed subset repurposed from 3D annotations, denoted as \textbf{SpatialScore‑Repurpose}~(1,091 samples), and 
(ii) the remaining 3,934 samples collected from existing datasets and manually curated, which form the \textbf{SpatialScore‑OpenSource} subset. 

As presented in Tab.~\ref{tab:quantitative_results_open_source} and Tab.~\ref{tab:quantitative_results_ours}, we compare our data‑driven and agent‑based approaches with several representative baselines on these subsets, leading to the following key observations:
(i) On both the SpatialScore‑OpenSource and SpatialScore‑Repurpose subsets, the Qwen3‑VL~\cite{Qwen3-VL} models fine‑tuned on SpatialCorpus, as well as our SpatialAgent, achieve substantial performance gains over their respective base models. 
This confirms the feasibility of both routes for enhancing spatial reasoning capabilities; 
(ii) Although supervised fine‑tuning brings performance gains, it may introduce potential biases. 
For example, Qwen3‑VL‑8B fine‑tuned on SpatialCorpus improves its overall accuracy on SpatialScore‑OpenSource from 42.97 to 48.72, while on SpatialScore‑Repurpose it increases from 54.53 to 76.29. 
This discrepancy largely arises because the latter subset is more closely aligned with the distribution of training data in SpatialCorpus;
and
(iii) In contrast, SpatialAgent shows a more moderate improvement on SpatialScore‑Repurpose~(from 54.53 to 67.51), but it also delivers consistent gains on the SpatialScore‑OpenSource subset~(from 48.72 to 50.01). 
As a training‑free approach, it further preserves the model’s general capabilities and avoids introducing distributional biases across different data sources.

\subsection{Additional Qualitative Results}
\label{subsec:additional_qualitative_results}
\begin{figure}[ht!]
  \centering
  \includegraphics[width=\textwidth]{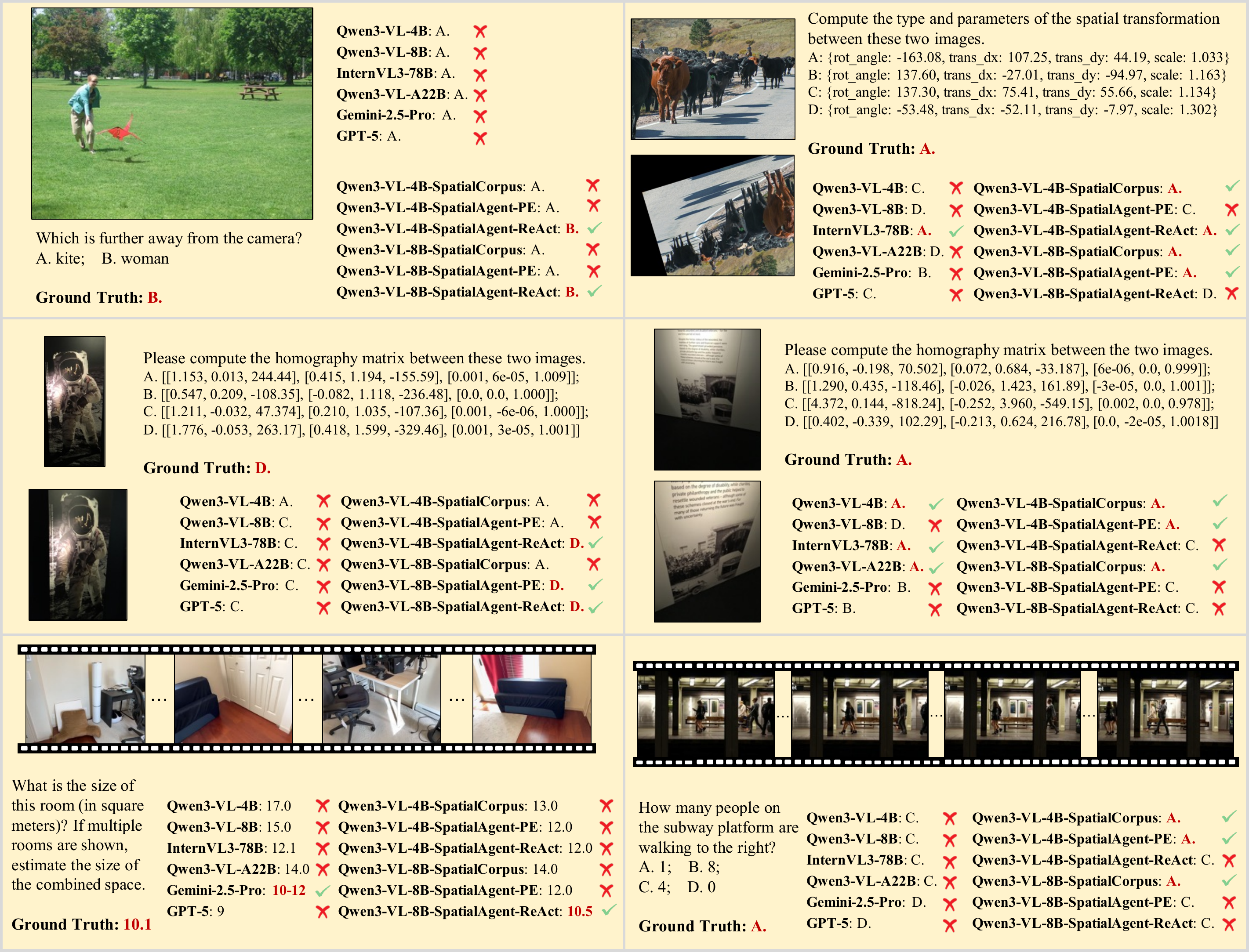} 
  % \vspace{-3pt}
  \caption{
  \centering
    \textbf{Additional Qualitative Results.}
  }
 \label{fig:more_qualitative_results_1}
 % \vspace{-6pt}
\end{figure}

\begin{figure}[ht!]
  \centering
  \includegraphics[width=\textwidth]{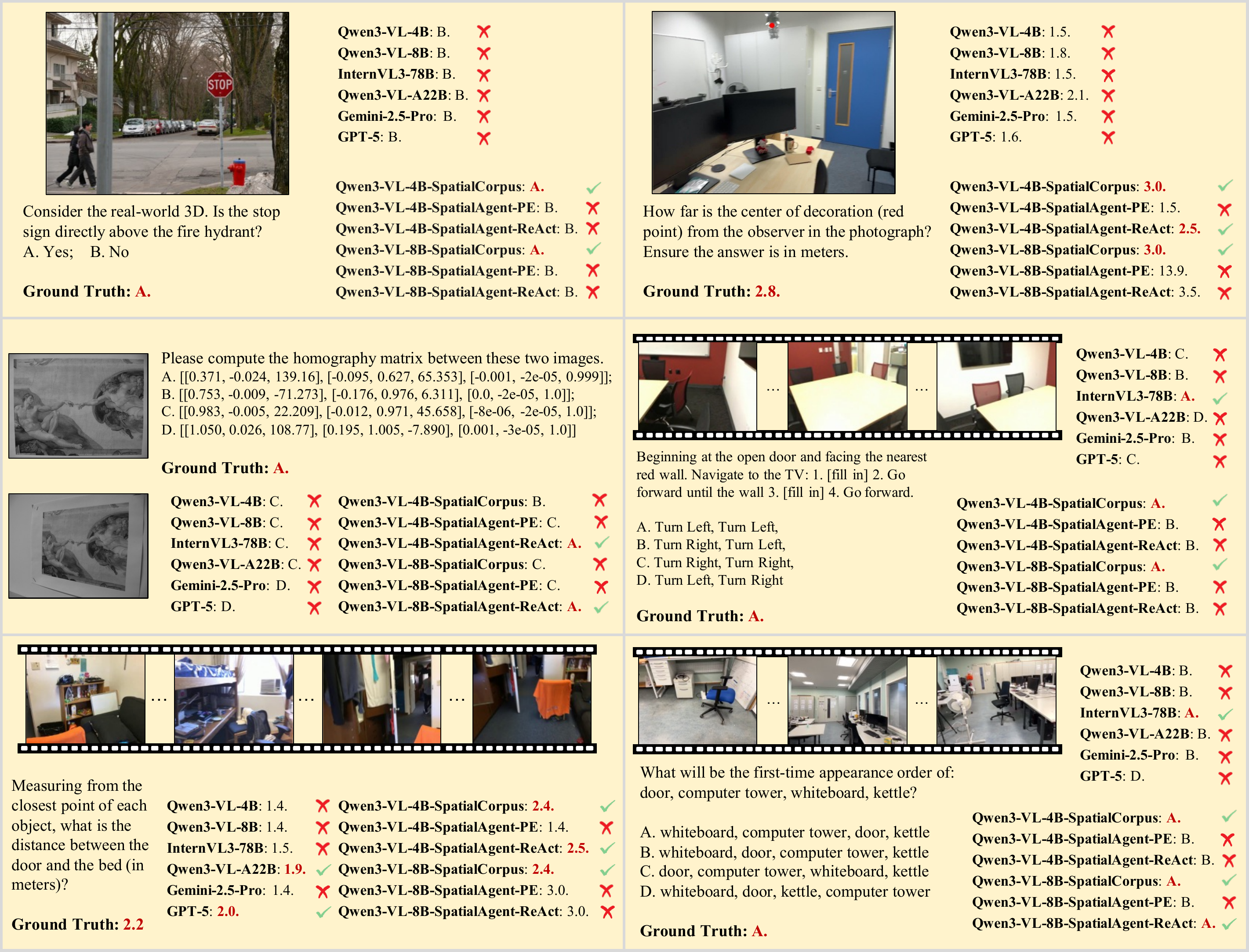} 
  % \vspace{-3pt}
  \caption{
  \centering
    \textbf{Additional Qualitative Results.}
  }
 \label{fig:more_qualitative_results_2}
 % \vspace{-6pt}
\end{figure}

We further include more qualitative results on SpatialScore from the models fine-tuned on our SpatialCorpus, as well as from our constructed SpatialAgent, as presented in Fig.~\ref{fig:more_qualitative_results_1} and Fig.~\ref{fig:more_qualitative_results_2}.
These results demonstrate that both our data‑driven and agent‑based approaches achieve superior performance on spatial intelligence tasks, even surpassing larger-scale models and proprietary systems.

%% file: tables/spatialcorpus_statistics.tex
\begin{table}[thbp]
    \centering
    \caption{
    \centering
        \textbf{Data Statistics of SpatialCorpus.}
    }
    \label{tab:statistics_spatialcorpus}
    % \vspace{-6pt}
    \renewcommand{\arraystretch}{1.2}
    
    \setlength{\tabcolsep}{5pt} 
    \resizebox{\textwidth}{!}{
    \begin{tabular}{@{} c | c | c @{}} 
        \toprule[1.2pt]
        
        \begin{tabular}[t]{@{}lr@{}}
            \multicolumn{2}{c}{\textbf{Question Type}} \\
            \cmidrule(lr){1-2}
            Multi-choice & 262,601 \\
            Judgment & 9,776 \\
            Open-ended & 58,425 \\
            
            \cmidrule(lr){1-2} 
            \multicolumn{2}{c}{\textbf{Input Modalities}} \\ 
            \cmidrule(lr){1-2} 
            
            Single-image & 270,812 \\
            Multi-image & 59,990 \\
        \end{tabular}
        &
        
        \begin{tabular}[t]{@{}lr@{}}
            \multicolumn{2}{c}{\textbf{Categories}} \\
            \cmidrule(lr){1-2}
            Mental Animation & 57,997 \\
            Depth Estimation & 57,843 \\
            Object Distance & 51,596 \\
            Object Motion & 20,000 \\
            Object Size & 21,614 \\
            Camera & 81,976 \\
            Object Localization & 39,776 \\
        \end{tabular}
        &
        
        \begin{tabular}[t]{@{}lr lr@{}}
            
            \multicolumn{4}{c}{\textbf{Tasks}} \\
            \cmidrule(lr){1-4}
            Spatial Map & 40,000 & Relative Size & 11,841 \\
            Multi-view Projection & 7,998 & Absolute Size & 9,773 \\
            2D/3D Rotation & 9,999 & Camera Intrinsics & 49,984 \\
            Absolute Depth & 32,594 & Camera Extrinsics & 4,997 \\
            Relative Depth & 25,249 & Camera Motion & 4,997 \\
            Absolute Distance & 25,712 & Homography Matrix & 21,998 \\
            Relative Distance & 25,884 & Object Existence & 9,776 \\
            Point Tracking & 20,000 & 3D Object Detection & 30,000 \\
        \end{tabular} \\
        
        \midrule[0.8pt] 
        \multicolumn{3}{c}{\textbf{Total: 330,802}} \\
        \bottomrule[1.2pt]
    \end{tabular}
    }
\end{table}

%% file: tables/spatialscore_statistics.tex
\begin{table}[htbp]
    \centering
    \caption{
        \textbf{Data Statistics of SpatialScore.}
        Here, {\em SpatialScore‑Repurpose} denotes the newly constructed samples based on 3D annotations.
    }
    \label{tab:statistics_spatialscore}
    % \vspace{-6pt}
    \renewcommand{\arraystretch}{1.2} % 保持舒适的行间距
    \setlength{\tabcolsep}{5pt} % 调整列间距
    \resizebox{\textwidth}{!}{
    \begin{tabular}{@{} c | c @{}} 
        \toprule[1.2pt]
        
        \begin{tabular}[t]{@{}lr@{}}
            \multicolumn{2}{c}{\textbf{Question Type}} \\
            \cmidrule(lr){1-2}
            Multi-choice & 3,686 \\
            Judgment & 463 \\
            Open-ended & 876 \\
            
            \addlinespace[10pt] 
            
            \multicolumn{2}{c}{\textbf{Input Modalities}} \\
            \cmidrule(lr){1-2}
            Single-image & 2,493 \\
            Multi-image & 1,339 \\
            Video & 1,193 \\
        \end{tabular}
        &
        
        \begin{tabular}[t]{@{}lr lr lr@{}}
            \multicolumn{6}{c}{\textbf{Data Sources}} \\
            \cmidrule(lr){1-6}
            SpatialScore‑Repurpose & 1,091 & CV-Bench & 171 & QSpatialBench & 39 \\
            VSI-Bench & 876 & Space-10 & 169 & RoboSpatial & 35 \\
            MMIU & 419 & BLINK & 143 & SpatialBench & 27 \\
            SPAR-Bench & 475 & SpatialSense & 124 & VSR & 21 \\
            SpatialEval & 293 & VLM4D & 116 & MIRAGE & 20 \\
            3DSRBench & 266 & SpatialViz & 114 & RealWorldQA & 11 \\
            SITE-Bench & 233 & MMSI-Bench & 69 & MMVP & 5 \\
            OmniSpatial & 205 & SRBench & 54 & STI-Bench & 49 \\
        \end{tabular} \\
        
        \midrule[0.8pt]
        
        \begin{tabular}[t]{@{}lr@{}}
            \multicolumn{2}{c}{\textbf{Categories}} \\
            \cmidrule(lr){1-2}
            Mental Animation & 447 \\
            Counting & 315 \\
            Depth Estimation & 520 \\
            Object Distance & 576 \\
            Object Motion & 415 \\
            View Reasoning & 446 \\
            Object Size & 559 \\
            Camera & 778 \\
            Temporal Reasoning & 272 \\
            Object Localization & 697 \\
        \end{tabular}
        &
        
        \begin{tabular}[t]{@{}lr lr lr@{}}
            \multicolumn{6}{c}{\textbf{Tasks}} \\
            \cmidrule(lr){1-6}
            Spatial Map & 220 & Absolute Distance & 313 & Camera Intrinsics & 112 \\
            Maze Navigation & 115 & Relative Distance & 263 & Camera Extrinsics & 246 \\
            Multi-view Projection & 46 & Point Tracking & 229 & Camera Motion & 174 \\
            Space Folding & 20 & Object Motion & 113 & Homography Matrix & 246 \\
            2D/3D Rotation & 46 & Velocity Estimation & 73 & Navigation Route & 105 \\
            Object Counting & 154 & View Perspective & 235 & Appearance Order & 167 \\
            Video Counting & 111 & Orientation & 211 & Object Existence & 220 \\
            Count with Relation & 50 & Relative Size & 189 & 2D Localization & 50 \\
            Absolute Depth & 325 & Absolute Size & 281 & 3D Object Detection & 212 \\
            Relative Depth & 195 & Size Compatibility & 89 & Spatial Position & 215 \\
        \end{tabular} \\
        
        \midrule[0.8pt]
        \multicolumn{2}{c}{\textbf{Total: 5,025}} \\
        \bottomrule[1.2pt]
    \end{tabular}
    }
\end{table}

%% file: tables/tool_computation.tex
\begin{table}[htp]
    \centering
    \caption{
        \textbf{Analysis of Computational Overhead for the SpatialAgent Toolbox.}
        Here, we evaluate SpatialAgent~(with Qwen3-VL-8B serving as the agent core and executing the {\em Plan-Execute} reasoning paradigm) on SpatialScore, reporting the average GPU memory usage, single-invocation latency, and invocation frequency of each tool in the toolbox.
    }
    \label{tab:tool_computation}
    \setlength{\tabcolsep}{4pt} 
    \renewcommand{\arraystretch}{1.2} 
    \resizebox{0.99\linewidth}{!}{
        \begin{tabular}{lr|lr|lr}
            \toprule[1.2pt]
            Depth & 2.5G/0.45s/29.5\% & Localization & 7.1G/0.60s/39.1\% & OpticalFlow & 0.3G/0.04s/3.76\%  \\
            Counting & 7.1G/0.54s/3.61\% & Orientation & 8.2G/0.62s/6.68\% & Segmentation & 8.6G/1.78s/2.16\% \\
            Intrinsics & 7.3G/0.51s/4.14\% & 3D Distance & 2.1G/0.37s/18.5\% & Homography & ---/3.79s/9.29\%  \\
            Extrinsics & 7.3G/0.51s/4.14\% & 3D Detection & 11.6G/0.52s/10.4\% & PointMatching & ---/1.78s/2.41\% \\
            \bottomrule[1.2pt]
        \end{tabular}
        }
    \vspace{-6pt}
\end{table}

%% file: tables/quantitative_results_open_source.tex
\begin{table*}[ht!]
    \caption{
        \textbf{Quantitative Comparisons on SpatialScore-OpenSource Subset.}
        Qwen3-VL is adopted in two ways: (i) supervised fine-tuned on our SpatialCorpus; and (ii) as the agent core to conduct reasoning using the Plan-Execute~(PE) and ReAct paradigms in SpatialAgent.
    }
    \label{tab:quantitative_results_open_source}
    \centering
    \setlength{\tabcolsep}{0.1cm}
    \renewcommand{\arraystretch}{1.1}
    \resizebox{\textwidth}{!}{
    \begin{tabular}{r|c|*{10}{c}}
    \toprule[1.5pt]
        \textbf{Methods} & \textbf{Overall} & \textbf{Mental.} & \textbf{Count.} & \textbf{Depth.} & \textbf{View-Rea.} & \textbf{Obj-Size.} &  \textbf{Obj-Loc.} & \textbf{Obj-Dist.} & \textbf{Obj-Mo.} & \textbf{Camera.} & \textbf{Temp-Rea.}   \\
        \midrule[0.8pt]
        \multicolumn{12}c{\textbf{\textit{Baselines}}} \\
        \midrule
        Chance-level~(Random) & 26.12 & 23.71 & 22.80 & 20.19 & 31.84 & 24.76 & 34.94 & 21.83 & 25.30 & 26.60 & 28.68 \\
        Human-level & \textbf{87.66} & \textbf{96.87} & \textbf{89.72} & \textbf{81.49} & \textbf{92.15} & \textbf{84.05} & \textbf{90.34} & \textbf{75.58} & \textbf{92.99} & \textbf{89.41} & \textbf{84.19} \\
        \midrule
        \multicolumn{12}c{\textbf{\textit{Representative Models}}} \\
        \midrule
        Qwen3-VL-30B-A3B~\cite{Qwen2.5-VL} & 48.60 & 46.31 & 58.86 & 45.06 & 47.98 & 53.87 & 61.38 & 33.65 & 56.10 & 40.39 & 45.59 \\
        Qwen3-VL-32B~\cite{Qwen2.5-VL} & 51.86 & 43.40 & 61.16 & 51.18 & 50.22 & \textbf{57.81} & 67.13 & \textbf{46.47} & 61.28 & 34.73 & 47.79 \\
        Qwen2.5-VL-72B~\cite{Qwen2.5-VL} & 45.44 & 53.69 & 49.49 & \underline{55.88} & 36.10 & 47.21 & 60.92 & 32.13 & 53.96 & 37.93 & 22.43 \\
        InternVL3-78B~\cite{InternVL3} & 48.23 & 50.34 & 59.19 & 44.46 & 45.74 & 44.78 & 65.98 & 35.74 & 57.32 & 37.93 & 43.75 \\
        Qwen3-VL-235B-A22B~\cite{Qwen2.5-VL} & 54.82 & 57.27 & \textbf{65.19} & 52.84 & \underline{52.47} & 56.85 & 66.90 & \underline{44.28} & \textbf{67.38} & 38.42 & 49.63 \\
        Claude-4.5-Sonnet~\cite{claude45} & 44.64 & 51.01 & 49.67 & 48.32 & 39.91 & 50.02 & 50.57 & 36.31 & 53.35 & 27.59 & 41.18 \\
        Gemini-2.5-Pro~\cite{comanici2025gemini} & \underline{56.70} & \underline{73.38} & \underline{64.06} & 51.06 & 46.41 & 54.70 & \underline{69.89} & 43.83 & 64.63 & \underline{45.57} & \underline{56.25} \\
        GPT-5~\cite{openai_gpt5_systemcard} & \textbf{59.09} & \textbf{78.08} & 57.59 & \textbf{56.33} & \textbf{54.04} & \underline{56.85} & \textbf{70.97} & 42.22 & \underline{67.07} & \textbf{47.29} & \textbf{62.13} \\
        \midrule
        \multicolumn{12}c{\textbf{\textit{Qwen3-VL-4B}}} \\
        \midrule
        Blind~(text only) & 24.96 & 27.29 & 11.20 & 25.63 & 25.34 & 37.29 & 27.82 & 22.74 & 23.17 & 22.66 & 20.22 \\
        Zero-shot & 40.39 & 37.81 & 48.22 & 36.81 & 34.75 & 43.70 & \underline{54.71} & 26.92 & 50.61 & 35.96 & 38.24 \\
        \rowcolor{cyan!10}
        w/ SpatialCorpus~(Ours) & \textbf{46.74} & \textbf{65.55} & 52.24 & \textbf{53.87} & 33.86 & 34.52 & 52.18 & \textbf{37.27} & \underline{55.49} & \underline{40.89} & \textbf{44.85} \\
        \rowcolor{cyan!20}
        w/ SpatialAgent-PE~(Ours) & 45.28 & \underline{56.15} & \textbf{54.98} & 42.40 & \underline{36.55} & \textbf{50.05} & \textbf{58.85} & 28.92 & \textbf{57.32} & 31.53 & 38.24 \\
        \rowcolor{cyan!20}
        w/ SpatialAgent-ReAct~(Ours) & \underline{46.49} & 46.53 & \underline{53.46} & \underline{47.36} & \textbf{39.01} & \underline{45.15} & 54.48 & \underline{31.86} & 53.05 & \textbf{54.93} & \underline{42.28} \\
        \midrule
        \multicolumn{12}c{\textbf{\textit{Qwen3-VL-8B}}} \\
        \midrule
        Blind~(text only) & 27.81 & 21.70 & 17.21 & 27.71 & 30.49 & 39.38 & 41.61 & 26.33 & 19.21 & 26.11 & 20.59 \\
        Zero-shot & 42.97 & 38.26 & 50.35 & 43.58 & 37.67 & 48.61 & 55.86 & 31.45 & 51.52 & 34.48 & 41.54 \\
        \rowcolor{cyan!10}
        w/ SpatialCorpus~(Ours) & 48.72 & \textbf{57.27} & 52.67 & \textbf{58.59} & 36.77 & 49.84 & 55.40 & \textbf{40.51} & 54.88 & 37.93 & \underline{44.85} \\
        \rowcolor{cyan!20}
        w/ SpatialAgent-PE~(Ours) & \underline{49.58} & \underline{50.11} & \textbf{54.48} & 50.99 & \underline{43.50} & \textbf{54.84} & \textbf{62.30} & \underline{38.43} & \textbf{58.23} & \underline{40.64} & 43.38 \\
        \rowcolor{cyan!20}
        w/ SpatialAgent-ReAct~(Ours) & \textbf{50.01} & 44.30 & \underline{53.49} & \underline{51.55} & \textbf{45.74} & \underline{51.39} & \underline{58.62} & 37.92 & \underline{57.62} & \textbf{51.97} & \textbf{51.84} \\
    \bottomrule[1.5pt]
    \end{tabular}
    % \vspace{-6pt}
}
\end{table*}

%% file: tables/quantitative_results_ours.tex
\begin{table*}[ht!]
    \caption{
        \textbf{Quantitative Comparisons on SpatialScore-Repurpose Subset.}
        Qwen3-VL is adopted in two ways: (i) supervised fine-tuned on our SpatialCorpus; and (ii) as the agent core to conduct reasoning using the Plan-Execute~(PE) and ReAct paradigms in SpatialAgent.
    }
    \label{tab:quantitative_results_ours}
    \centering
    \setlength{\tabcolsep}{0.1cm}
    \renewcommand{\arraystretch}{1.1}
    % \footnotesize
    \resizebox{.9\textwidth}{!}{
    \begin{tabular}{r|c|*{6}{c}}
    \toprule[1.5pt]
        \textbf{Methods} & \textbf{Overall} & \textbf{Depth.} & \textbf{Obj-Size.} &  \textbf{Obj-Loc.} & \textbf{Obj-Dist.} & \textbf{Obj-Mo.} & \textbf{Camera.}   \\
        \midrule[0.8pt]
        \multicolumn{8}c{\textbf{\textit{Baselines}}} \\
        \midrule
        Chance-level~(Random) & 36.13 & 31.86 & 50.33 & 44.66 & 31.78 & 24.14 & 31.18 \\
        Human-level & \textbf{82.79} & \textbf{85.12} & \textbf{89.29} & \textbf{67.18} & \textbf{90.70} & \textbf{100.00} & \textbf{84.14} \\
        \midrule
        \multicolumn{8}c{\textbf{\textit{Representative Models}}} \\
        \midrule
        Qwen3-VL-30B-A3B~\cite{Qwen2.5-VL} & 58.30 & \underline{59.92} & 66.12 & \underline{74.81} & 65.26 & 72.41 & 37.90 \\
        Qwen3-VL-32B~\cite{Qwen2.5-VL} & \underline{62.22} & 53.85 & 65.75 & 73.66 & 66.41 & \textbf{87.36} & 48.39 \\
        Qwen2.5-VL-72B~\cite{Qwen2.5-VL} & 59.15 & 57.38 & 64.31 & 65.27 & 60.21 & \underline{80.46} & 48.39 \\
        InternVL3-78B~\cite{InternVL3} & 59.47 & \textbf{63.01} & 70.37 & 65.27 & \underline{65.89} & 72.41 & 45.43 \\
        Qwen3-VL-235B-A22B~\cite{Qwen2.5-VL} & \textbf{63.14} & 58.01 & \textbf{70.92} & \textbf{75.19} & \textbf{71.61} & 77.01 & 47.58 \\
        Claude-4.5-Sonnet~\cite{claude45}  & 49.45 & 44.05 & 68.75 & 58.78 & 46.88 & 22.99 & 45.43 \\
        Gemini-2.5-Pro~\cite{comanici2025gemini} & 55.18 & 51.39 & \underline{70.59} & 60.69 & 58.14 & 30.12 & \underline{52.15} \\
        GPT-5~\cite{openai_gpt5_systemcard} & 54.63 & 51.15 & 67.77 & 61.45 & 54.69 & 19.54 & \textbf{54.86} \\
        \midrule
        \multicolumn{8}c{\textbf{\textit{Qwen3-VL-4B}}} \\
        \midrule
        Blind~(text only) & 39.45 & 33.65 & 63.64 & 40.84 & 36.43 & 16.09 & 38.98 \\
        Zero-shot & 50.21 & 40.60 & 64.48 & 67.18 & 55.87 & 62.07 & 31.99 \\
        \rowcolor{cyan!10}
        w/ SpatialCorpus~(Ours) & \textbf{75.53} & \textbf{73.89} & \textbf{77.93} & \textbf{70.23} & \textbf{81.47} & \textbf{96.55} & \textbf{72.04} \\
        \rowcolor{cyan!20}
        w/ SpatialAgent-PE~(Ours) & 62.07 & 57.26 & \underline{71.51} & \underline{68.32} & 63.57 & \underline{72.41} & 53.23 \\
        \rowcolor{cyan!20}
        w/ SpatialAgent-ReAct~(Ours) & \underline{64.04} & \underline{66.38} & 69.94 & 64.12 & \underline{64.64} & 54.02 & \underline{63.44} \\
        \midrule
        \multicolumn{8}c{\textbf{\textit{Qwen3-VL-8B}}} \\
        \midrule
        Blind~(text only) & 41.23 & 32.34 & 52.89 & 52.67 & 34.88 & 26.44 & 37.90 \\
        Zero-shot & 54.53 & 58.94 & 64.86 & \textbf{69.85} & 57.18 & 72.41 & 33.87 \\
        \rowcolor{cyan!10}
        w/ SpatialCorpus~(Ours) & \textbf{76.29} & \textbf{82.54} & \textbf{82.06} & \underline{68.32} & \textbf{80.62} & \textbf{100.00} & \textbf{70.97} \\
        \rowcolor{cyan!20}
        w/ SpatialAgent-PE~(Ours) & 64.19 & \underline{65.91} & 72.77 & 67.18 & 61.35 & \underline{73.56} & 57.53 \\
        \rowcolor{cyan!20}
        w/ SpatialAgent-ReAct~(Ours) & \underline{67.51} & 63.09 & \underline{75.09} & \textbf{69.85} & \underline{63.57} & \underline{73.56} & \underline{64.78} \\
    \bottomrule[1.5pt]
    \end{tabular}
    \vspace{-6pt}
}
\end{table*}